\documentclass[runningheads]{llncs}
\usepackage{graphicx}
\usepackage{comment}
\usepackage{amsmath,amssymb} 
\usepackage{color}
\usepackage{siunitx}

\newcommand{\boldstart}[1]{\noindent\textbf{#1}}

\usepackage[width=122mm,left=12mm,paperwidth=146mm,height=193mm,top=12mm,paperheight=217mm]{geometry}
\usepackage{adjustbox}

\usepackage{subfig}
\begin{document}
\pagestyle{headings}
\mainmatter
\def\ECCVSubNumber{1152}  

\title{TexMesh: Reconstructing Detailed Human Texture and Geometry from RGB-D Video}

\titlerunning{TexMesh}
%
\author{Tiancheng Zhi\inst{1}\thanks{Work was done during TZ internship at Facebook Reality Labs, Sausalito, CA, USA.} \and
Christoph Lassner\inst{2} \and
Tony Tung\inst{2} \and Carsten Stoll\inst{2} \and \\ Srinivasa G. Narasimhan \inst{1} \and Minh Vo\inst{2}}
\authorrunning{T. Zhi et al.}
%
\institute{Carnegie Mellon University, \email{\{tzhi,srinivas\}@cs.cmu.edu} \and Facebook Reality Labs, \email{\{classner,tony.tung,carsten.stoll,minh.vo\}@fb.com}
}
\maketitle

\begin{abstract}
We present TexMesh, a novel approach to reconstruct detailed human meshes with high-resolution full-body texture from RGB-D video. TexMesh enables high quality free-viewpoint rendering of humans. Given the RGB frames, the captured environment map, and the coarse per-frame human mesh from RGB-D tracking, our method reconstructs spatiotemporally consistent and detailed per-frame meshes along with a high-resolution albedo texture. By using the incident illumination we are able to accurately estimate local surface geometry and albedo, which allows us to further use photometric constraints to adapt a synthetically trained model to real-world sequences in a \textit{self-supervised} manner for detailed surface geometry and high-resolution texture estimation. In practice, we train our models on a short example sequence for self-adaptation and the model runs at interactive framerate afterwards. We validate TexMesh on synthetic and real-world data, and show it outperforms the state of art quantitatively and qualitatively.
\end{abstract}

\section{Introduction}
An essential component of VR communication, modern game and movie production is the ability to reconstruct accurate and detailed human geometry with high-fidelity texture from real world data. This allows us to re-render the captured character from novel viewpoints. This is challenging even when using complex multi-camera setups~\cite{kanade1997virtualized,vlasic2009dynamic,collet2015high,matsuyama2002generation}. Recent works such as Tex2Shape~\cite{alldieck2019tex2shape} and Textured Neural Avatars~\cite{shysheya2019textured} (TNA) have shown how to reconstruct geometry and texture respectively using nothing but a single RGB image/video as input.

\begin{figure}
\centering
\subfloat[(a) RGB]{
\includegraphics[width=42pt]{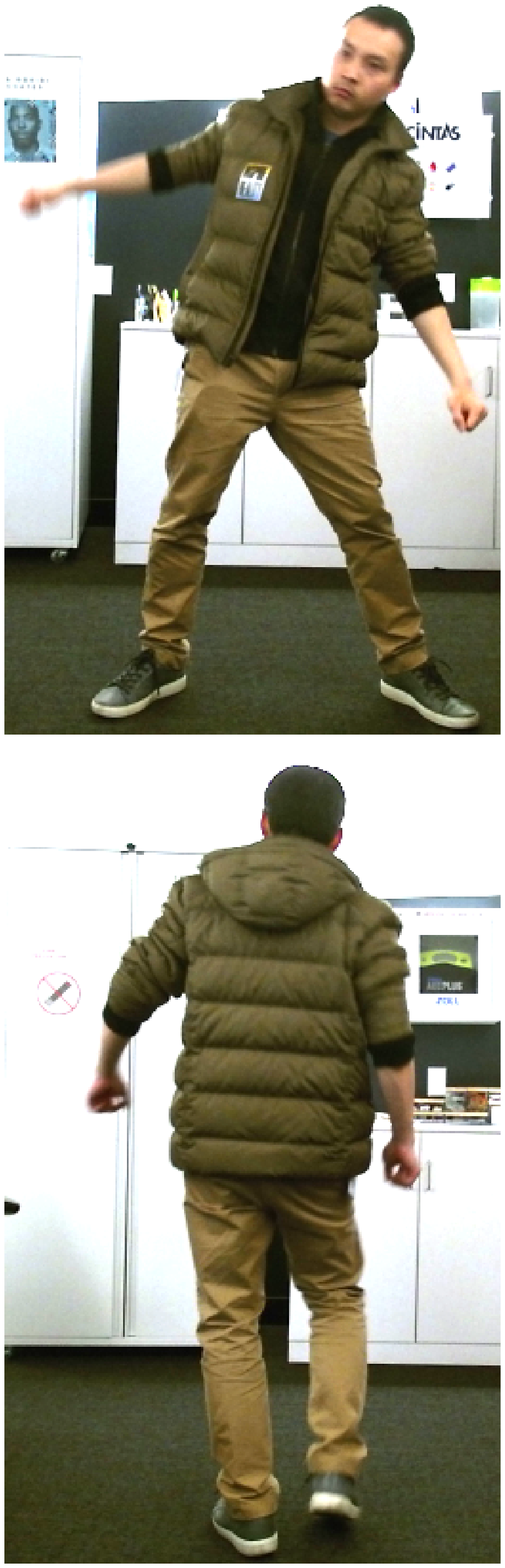}}
\quad
\subfloat[(b) Tex2Shape\cite{alldieck2019tex2shape}+TNA\cite{shysheya2019textured}]{
\label{fig:teaser-tex2shape+tna}
\includegraphics[width=126pt]{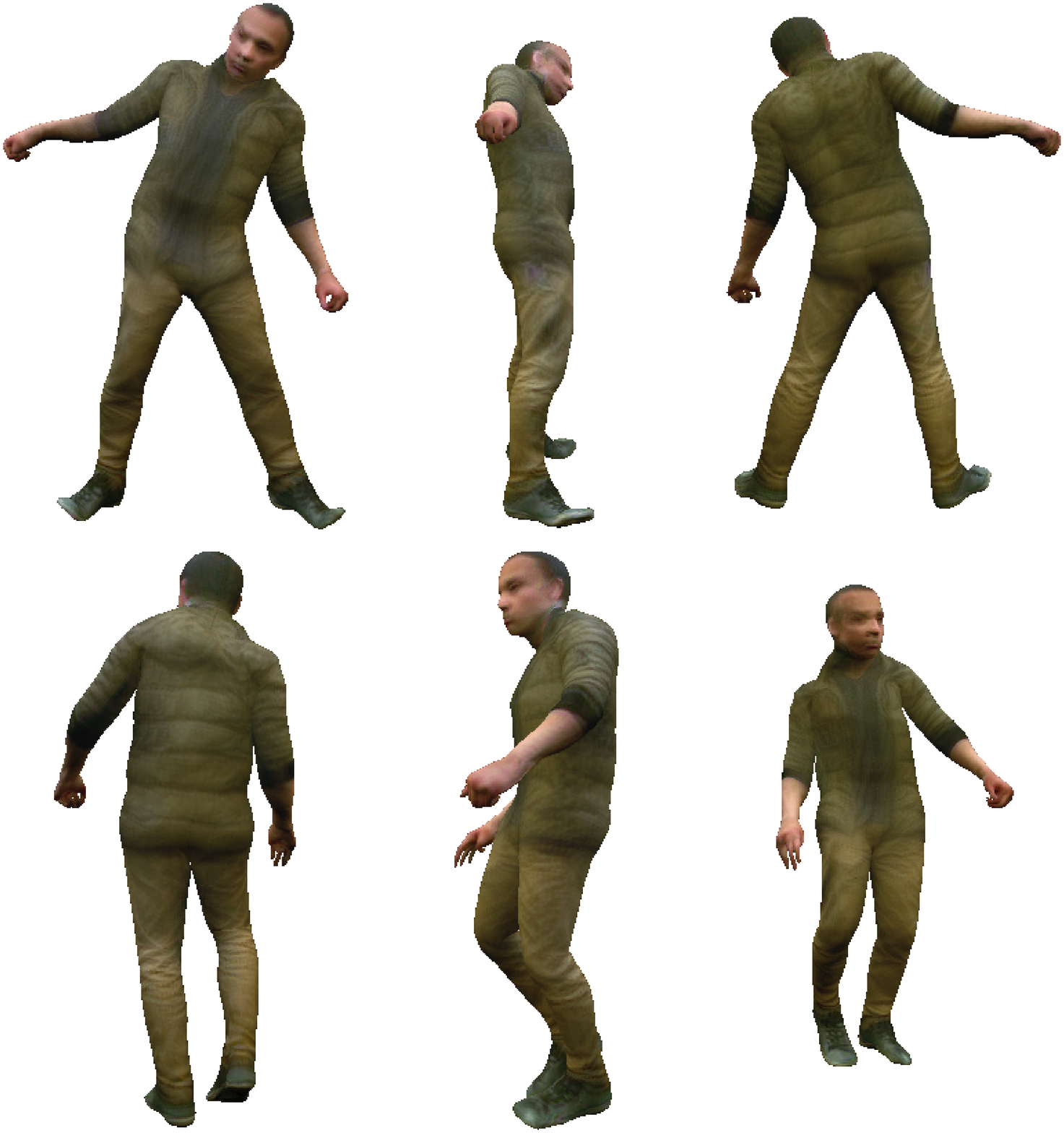}}
\quad
\subfloat[(c) Ours]{
\includegraphics[width=126pt]{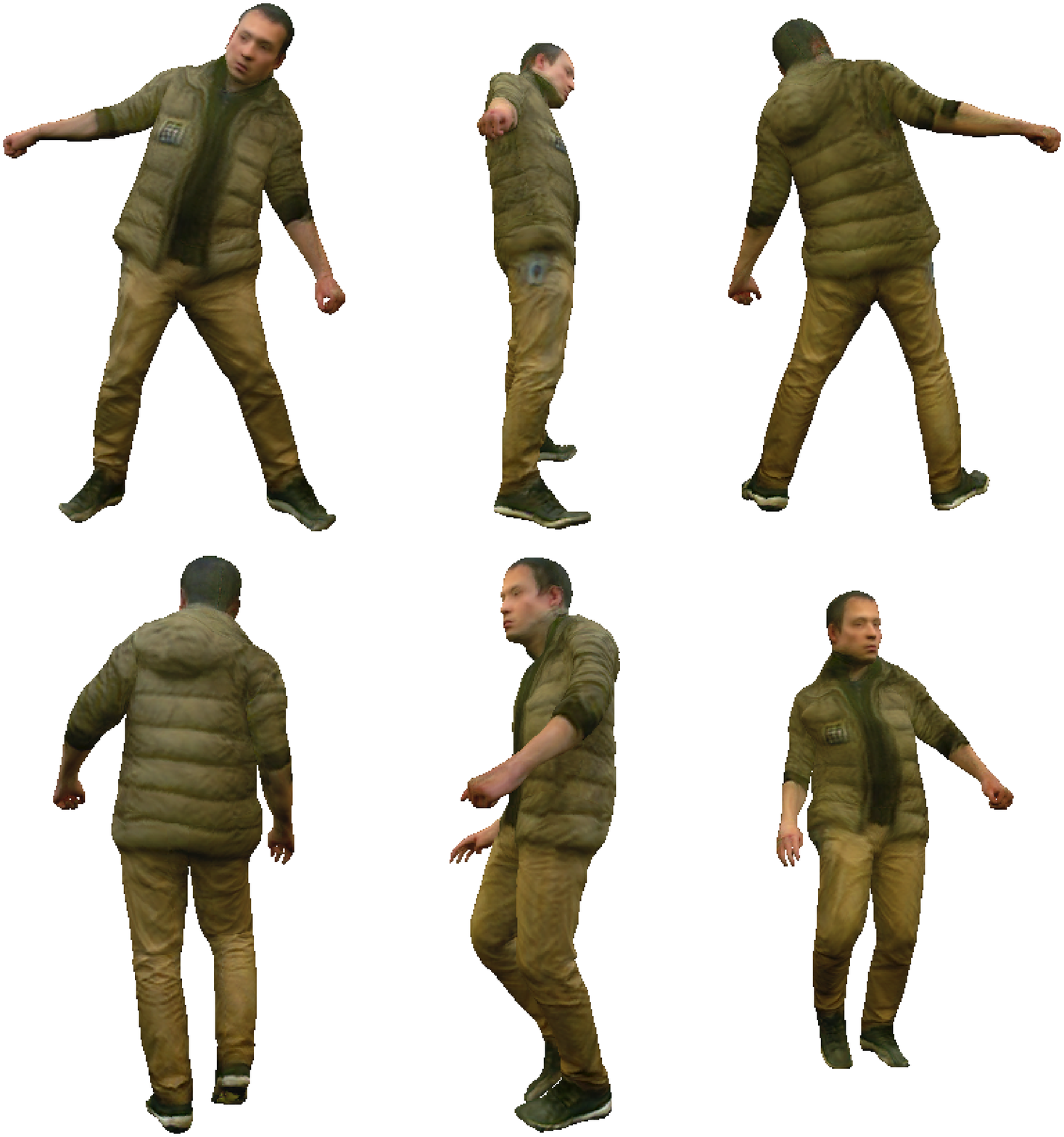}}
\caption{Tex2Shape~\cite{alldieck2019tex2shape} + TNA~\cite{shysheya2019textured} vs. Our Result. The mesh with texture is rendered from different viewpoints. Our approach reconstructs more detailed geometry, such as the moving jacket, as well as more accurate texture.}
\label{fig:teaser}
\end{figure}

Fig.~\ref{fig:teaser} shows examples of novel viewpoint synthesis using Tex2Shape for geometry reconstruction and TNA for texture estimation. Tex2Shape is a single view method trained only on synthetic images without adaptation to real data. Hence, it generates the rough shape of the actor but misses some of the finer geometric details that appear in the real data (Fig.~\ref{fig:teaser} (b)), and often hallucinates incorrect deformation memorized from its training data especially in occluded parts. As TNA does not consider the input lighting, the estimated texture contains the baked-in lighting of the original input sequence. Besides, due to small geometric misalignments, the estimated texture is blurry.

To address these issues, we introduce TexMesh, a novel framework to reconstruct both significantly higher quality mesh and texture from a real world video (see Fig.~\ref{fig:teaser} (c)). Our model takes an RGB video, a corresponding environment map, and a per-frame coarse mesh as inputs, and produces a per-frame fine mesh and a high-resolution texture shared across the whole video that can be used for free-viewpoint rendering. The coarse mesh is a parametric human model obtained by 3D tracking from an RGB-D camera~\cite{walsman2017dynamic}. We use a short real video clip for self-adaptation after which TexMesh runs at 18 fps (not including human segmentation). 
In the pipeline, depth is used only for obtaining the coarse mesh.

Concretely, for texture generation, we parameterize the texture using a CNN and optimize it on real data by comparing the rasterized images with a limited number of selected key albedo images. Our design offers three benefits: no shading and texture mixing, less geometric misalignment leading to less blur, and built-in CNN structure prior for noise and artifact removal~\cite{ulyanov2018deep}. 
For mesh reconstruction, we propose to first pre-train a displacement map prediction model on synthetic images with supervision, and later optimize it on a real sequence in a self-supervised manner using photometric perceptual loss and spatiotemporal deformation priors to obtain detailed clothing wrinkles even for occluded parts.

Experiments show that the proposed method provides clear texture with high perceptual quality, and detailed dynamic mesh deformation in both the visible and occluded parts. The resulting mesh and texture can produce realistic free-viewpoint rendering (Fig.~\ref{fig:render}) and relighting results (Fig.~\ref{fig:relight}).

\boldstart{Contributions:} (1) We present a self-supervised framework to adapt the training on synthetic data to real data for high quality texture and mesh reconstruction. This framework is based on (2) a texture generation method including albedo estimation, frame selection, and CNN refinement and (3) a mesh reconstruction method that faithfully reconstructs clothing details even in invisible regions using shading and spatiotemporal deformation priors. Our method enables state of art free-viewpoint rendering of humans on challenging real videos.

\section{Related Work}

\noindent\textbf{Human Shape Reconstruction. }The key to human shape reconstruction is to incorporate human priors to limit the solution space. Template-based methods~\cite{habermann2019livecap,xu2018monoperfcap} obtain human geometry by deforming the pre-scanning model. Model-based methods~\cite{zhou2010parametric,jain2010moviereshape,bogo2016keep,lassner2017unite,rhodin2016general,huang2017towards,kanazawa2018end,omran2018neural,xu2019denserac} fit a parametric naked-body model~\cite{loper2015smpl} to 2D poses or silhouette. While these methods estimate the coarse shape well, the recovered surface geometry is usually limited to tight clothing only~\cite{bogo2015detailed}. To tackle this problem,~\cite{yu2018doublefusion,yu2017bodyfusion} combine depth fusion~\cite{newcombe2011kinectfusion,newcombe2015dynamicfusion} and human priors~\cite{loper2015smpl} and show highly accurate reconstruction in visible parts but not occluded regions. With multiple images, \cite{alldieck2018detailed,alldieck2019learning,alldieck2018video} model clothing by deforming a parametric model~\cite{loper2015smpl} to obtain an animatable avatar, which enables powerful VR applications. However, the clothing details are inconsistent across frames, making the re-targeting result not faithful to the observation. Some methods treat clothing as separate meshes, providing strong possibilities for simulation, but are limited to a single clothing~\cite{lahner2018deepwrinkles}, pre-defined categories~\cite{bhatnagar2019multi}, or mechanical properties~\cite{yu2019simulcap}. Recently, single image methods utilizes deep learning for recovering detailed shapes, including UV space methods~\cite{lahner2018deepwrinkles,alldieck2019tex2shape}, volumetric methods~\cite{zheng2019deephuman}, implicit surface~\cite{saito2019pifu,huang2020arch}, and method combining learning and shading~\cite{zhu2019detailed}. 
They provide excellent details in visible regions, but hallucinate invisible parts rather than using temporal information for faithful reconstruction. In contrast to the above methods, we exploit photometric and spatiotemporal deformation cues to obtain detailed mesh, even in occluded regions.\\

\noindent\textbf{Human Texture Generation.} The key of texture generation is to fuse information from multiple images. Sampling based methods~\cite{alldieck2018detailed,alldieck2018video} sample colors from video frames and merge them together. TNA~\cite{shysheya2019textured} uses photometric supervision from rendering. These methods work well for videos with limited deformation but fail when the misalignment caused by large clothing deformation is significant.
Single view methods~\cite{grigorev2019coordinate,oechsle2019texture} avoid the problem of fusing multi-view information by hallucinating the occluded part. Yet, the hallucinated texture may not match the real person. Different from these methods, our method handles large deformation and provides high quality albedo texture.\\

\noindent\textbf{Face Reconstruction.} Face modeling is closely related to body modeling but with limited self-occlusions. Methods using photometric cues reconstruct detailed geometry and albedo via self-supervision~\cite{tewari2019fml,tewari2018self}. Deep learning also provides the opportunity for learning face geometry from synthetic data \cite{sela2017unrestricted} or both synthetic data and real data~\cite{sengupta2018sfsnet}. These methods achieve high quality results but cannot be trivially extended to full body, especially for occluded parts.

\begin{figure}
\centering
\includegraphics[width=0.85\linewidth]{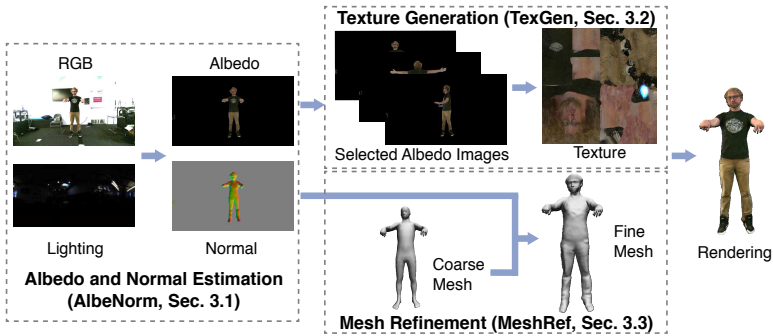}
\caption{Framework Overview. Our method consists of three modules: Albedo and Normal Estimation (AlbeNorm) pre-precosses RGB images to estimate albedo and normal; Texture Generation (TexGen) selects key albedo frames and recovers a texture map; Mesh Refinement (MeshRef) takes a coarse mesh and a normal image as input and outputs a fine mesh. We pre-train AlbeNorm and MeshRef on synthetic data. Then given a short clip of the real video, we optimize TexGen to obtain texture and finetune MeshRef via self-supervision. Finally, we run AlbeNorm and MeshRef on the whole video for fine meshes.}
\label{fig:framework}
\end{figure}

\section{Method}
As in Fig.~\ref{fig:framework}, our framework consists of three modules: Albedo and Normal Estimation (AlbeNorm), Texture Generation (TexGen), and Mesh Refinement (MeshRef). AlbeNorm takes an RGB image and the lighting, represented using Spherical Harmonics~\cite{ramamoorthi2001efficient}, and estimates texture and geometry information in the form of albedo and normal images. This is used consecutively to refine the texture and geometry estimates: TexGen selects albedo key frames and generates a high-resolution texture map from them. MeshRef takes a coarse mesh from RGB-D tracking and a normal image and estimates a refined mesh. Ground truth data for these tasks is naturally scarce. However, we observe that (1) synthetic data can be used to train the AlbeNorm and MeshRef. In synthetic settings we can use datasets with detailed person models to obtain highly detailed geometry estimates; (2) TexGen and MeshRef can be finetuned on a short sequence using perceptual photometric loss and the spatiotemporal deformation priors in a \emph{self-supervised} manner. This makes training on large annotated video datasets obsolete.
While we train AlbeNorm using only synthetic data, the model empirically generalizes well to real data. The final results are a single high-resolution full-body texture for the whole sequence and fine body geometry predicted at every frame. We describe our method in details in the following sections.

\subsection{Albedo and Normal Estimation}
\label{ssec:albenorm}
The cornerstone for our method is a good albedo and normal estimation: the normal is key to recover detailed geometry in MeshRef and the albedo is key to estimate clear texture in TexGen. To extract albedo and normals, under the usual assumptions of Lambertian materials, distant light sources, and no cast shadows, we can fully represent the geometry and color of an image using a normal image and an albedo image. The normal encodes the local geometry information, and together with the incident illumination, it can be used to generate a shading image. The albedo encodes the local color and texture. The decomposition into shading and albedo is typically not unique, as we can potentially explain texture changes through normal changes. This is where the AlbeNorm module comes into play: to prevent shading from `leaking' into texture and the albedo gradients from being explained as geometry change, we use the module to decouple the two components. Unlike \cite{alldieck2018detailed}, we resolve the scale ambiguity with the known lighting. 

The AlbeNorm module uses a CNN to predict albedo and normal images. The inputs to this module are a segmented human image~\cite{chen2017rethinking,piccardi2004background} and the incident illumination represented as Spherical Harmonics~\cite{ramamoorthi2001efficient}. Knowing the lighting information, we omit the scene background and process the masked human region. Concretely, let $A_{p}$ and $A_{g}$ be the predicted and ground truth albedo images, $N_{p}$ and $N_{g}$ be the predicted and ground truth normal images, and $M$ the human mask, respectively. Then, our supervised loss $L_{AN}$ with weights $\lambda^{an}_{a}$ and $\lambda^{an}_{n}$ is:
\begin{equation}
    L_{AN} = \lambda^{an}_{a} ||(A_{p} - A_{g}) \cdot M||_{1} + \lambda^{an}_{n} ||(N_{p} - N_{g}) \cdot M||_{1},
\end{equation}
\begin{figure}[t]
\centering
\subfloat[(a)]{
\includegraphics[height=0.23\linewidth]{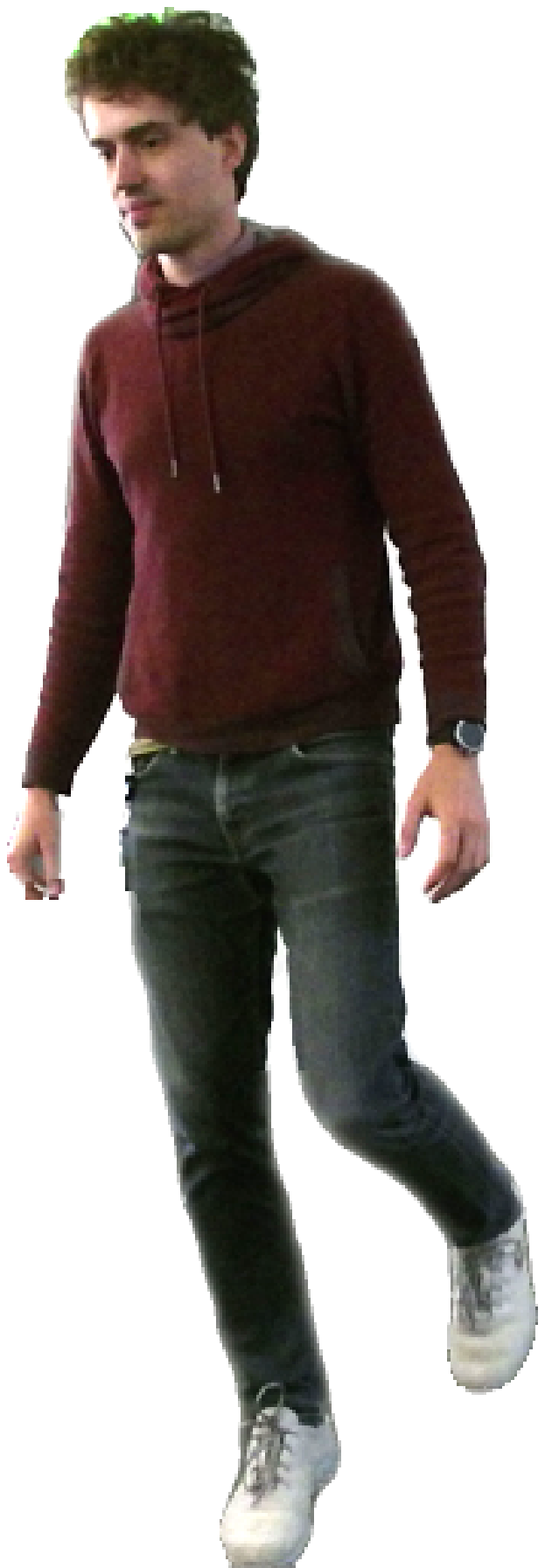}}
\subfloat[(b)]{
\includegraphics[height=0.23\linewidth]{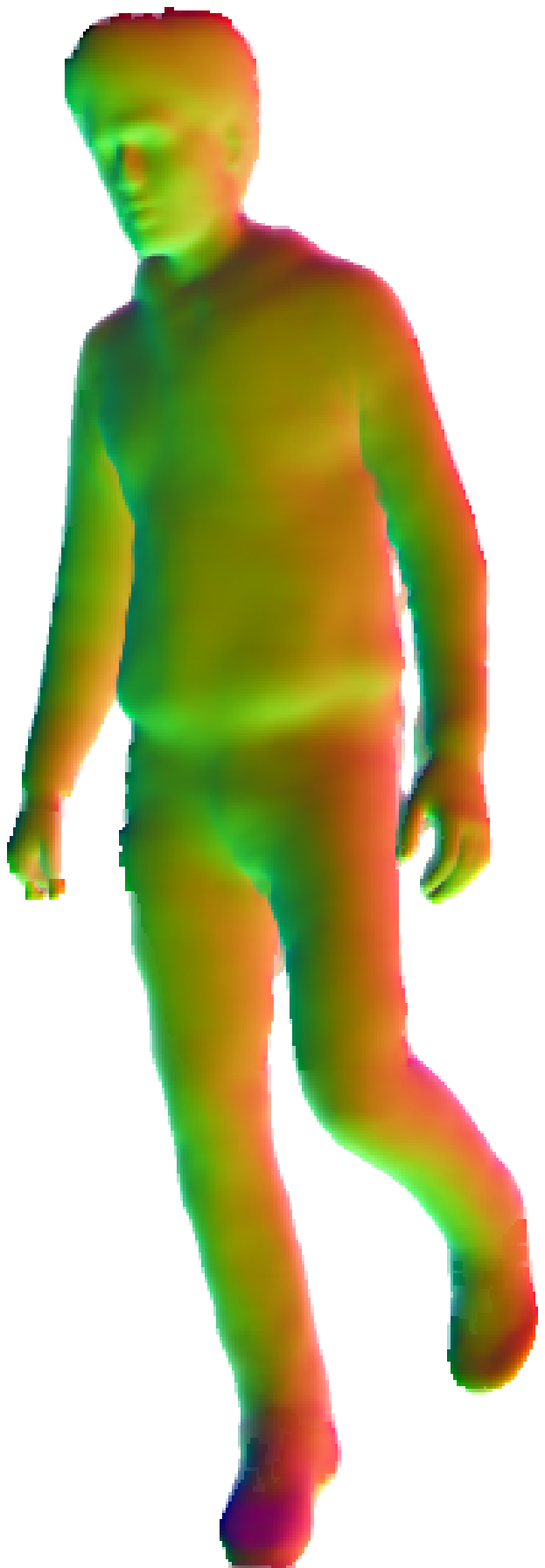}}
\subfloat[(c)]{
\includegraphics[height=0.23\linewidth]{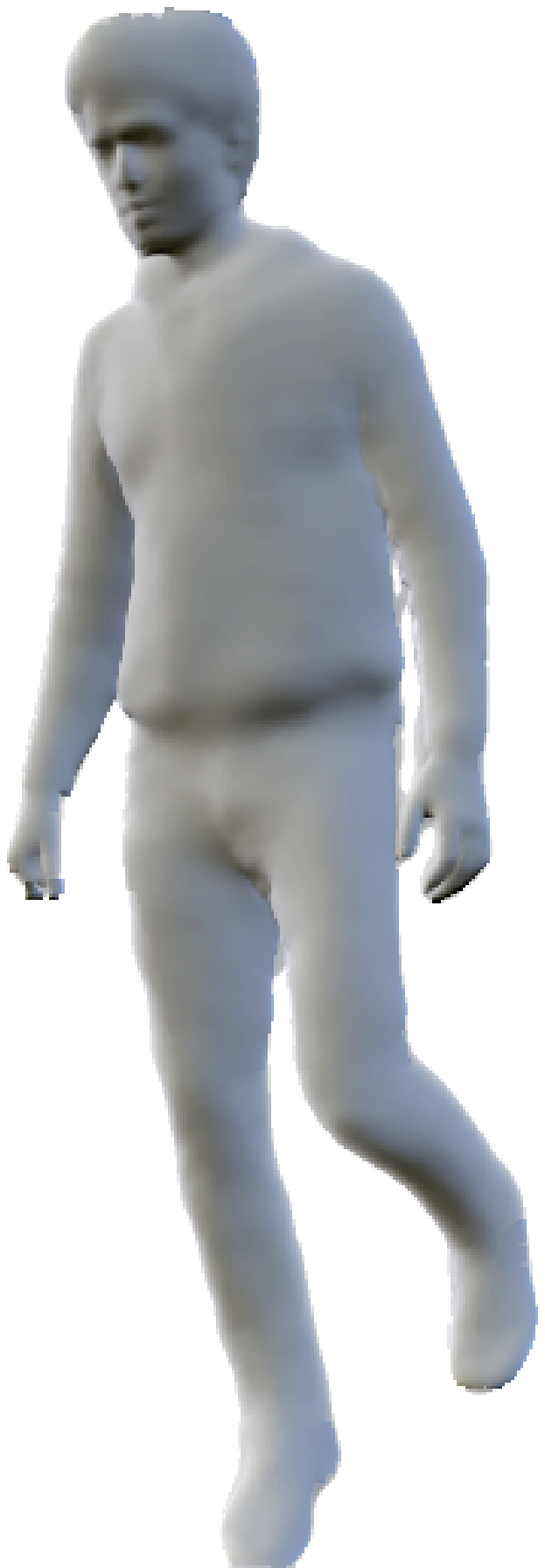}}
\quad
\subfloat[
(d)]{
\includegraphics[height=0.23\linewidth]{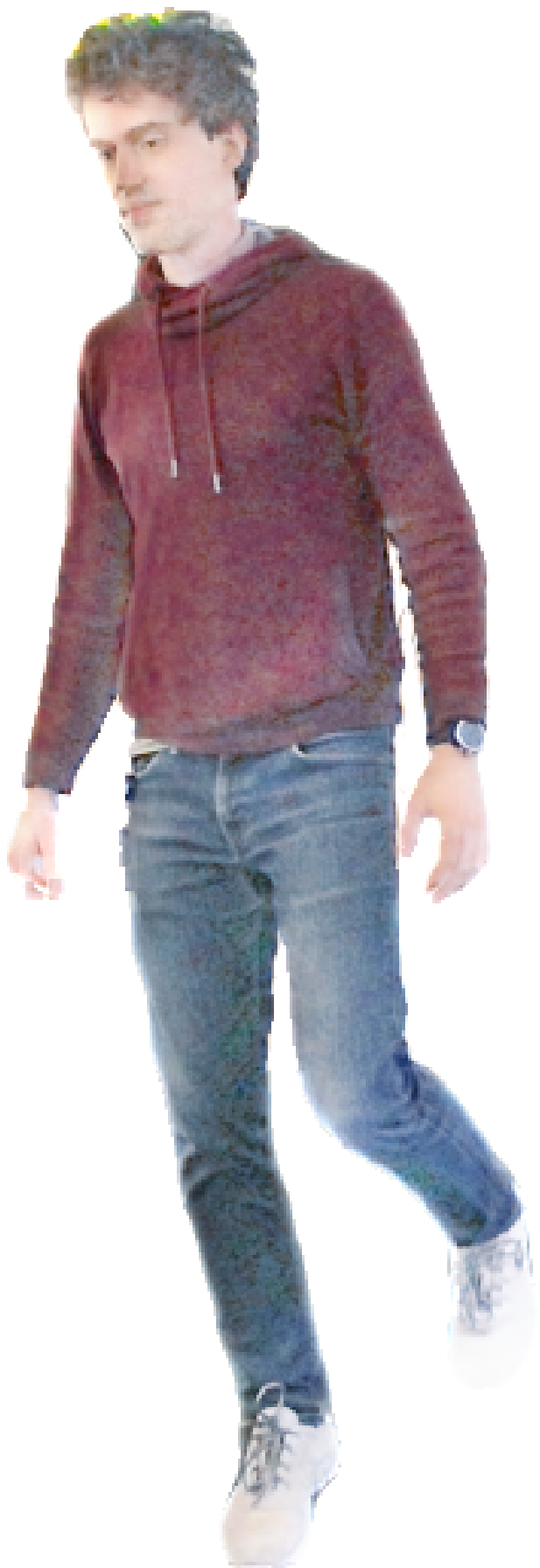}}
\subfloat[(e)]{
\includegraphics[height=0.23\linewidth]{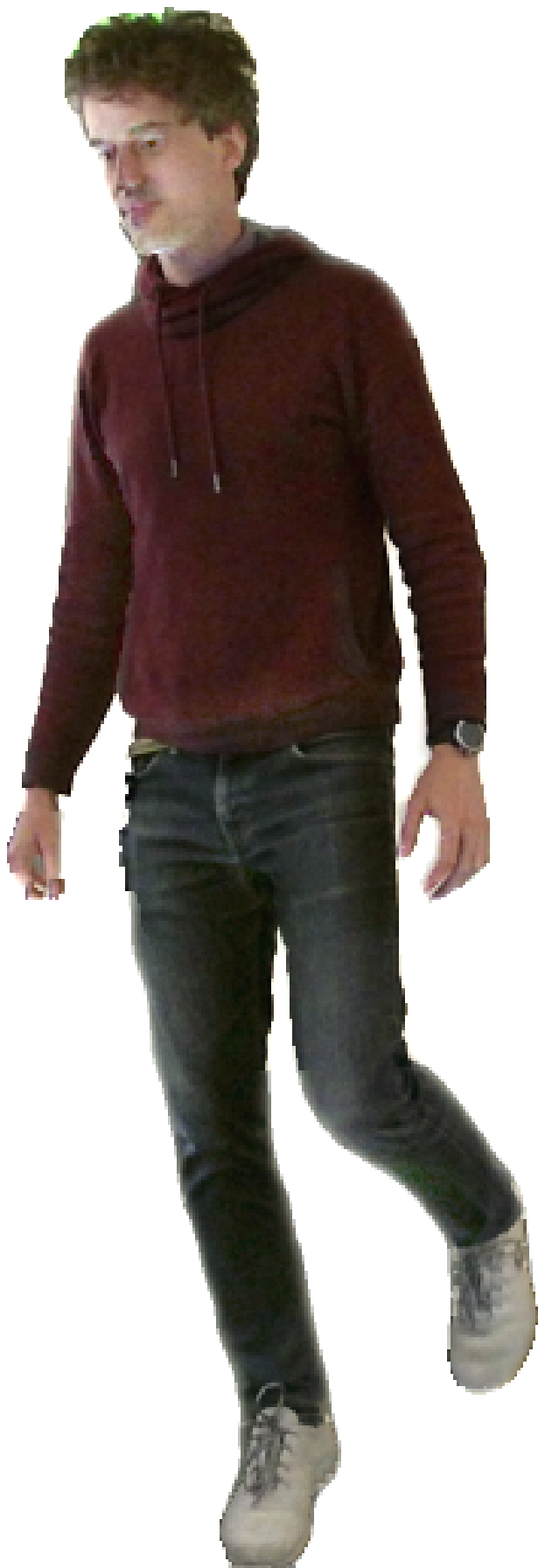}}
\subfloat[(f)]{
\includegraphics[height=0.23\linewidth]{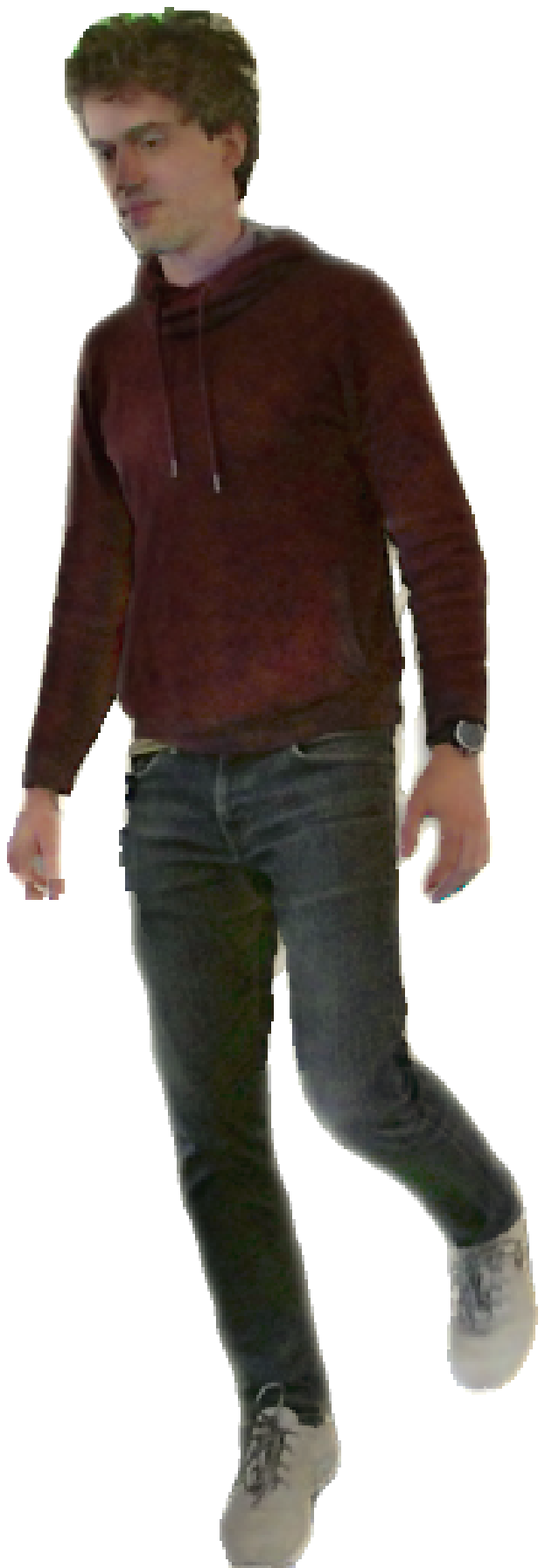}}
\quad
\subfloat[(g)]{
\includegraphics[height=0.23\linewidth]{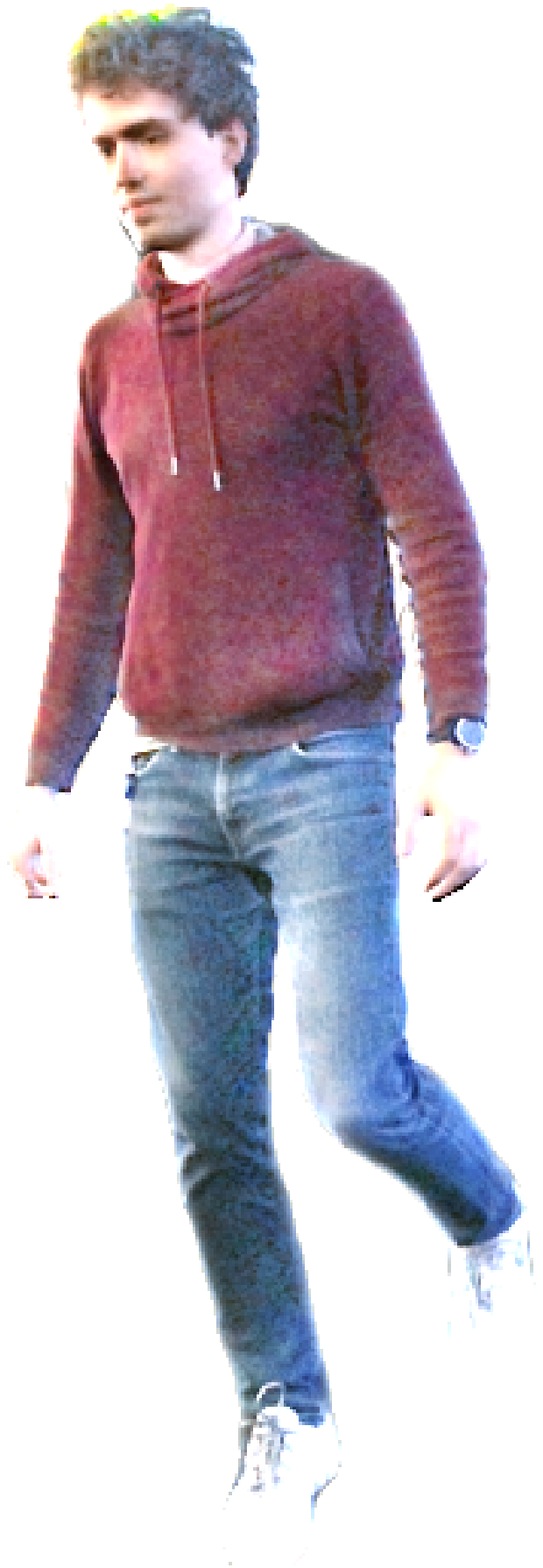}}
\subfloat[(h)]{
\includegraphics[height=0.23\linewidth]{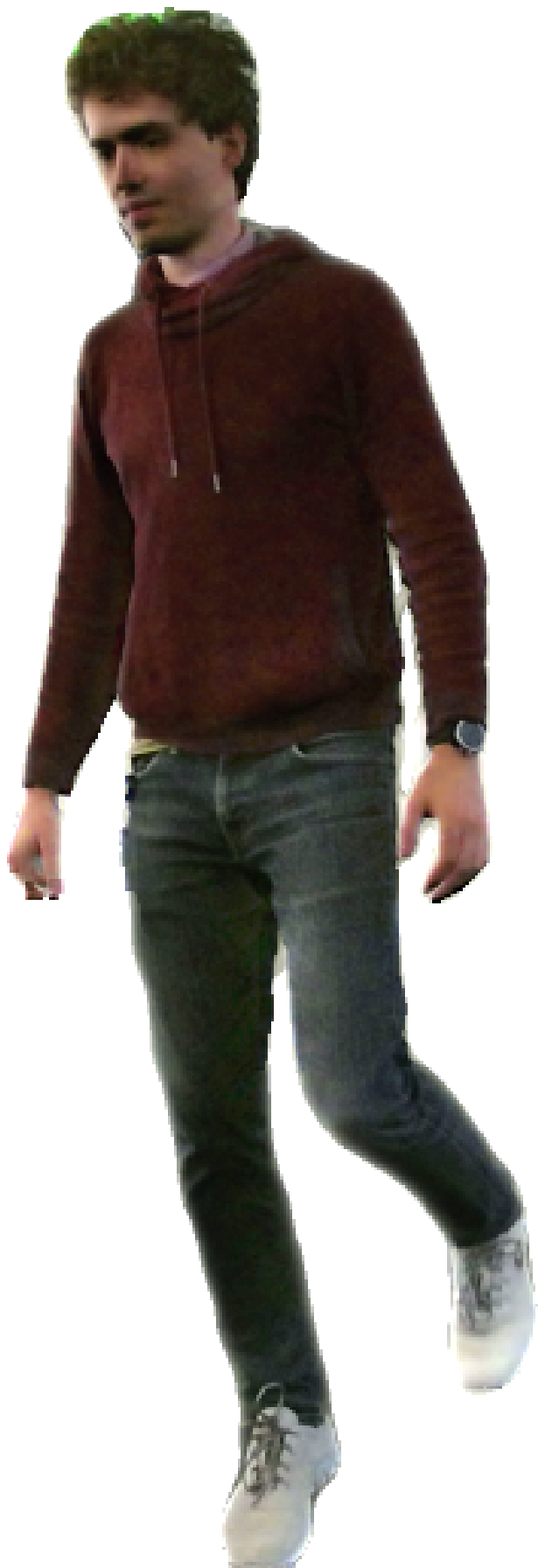}}
\caption{Intermediate results of AlbeNorm. (a) original RGB image; (b) predicted normal; (c) calculated shading; (d) albedo directly from CNN; (e) albedo from dividing RGB by shading; (f) final albedo; (g) rendering using (d); (h) rendering using (f). The final albedo (f) includes less shading information than (e) (e.g., the face region), and (h) resembles the original RGB (a) better than (g). }
\label{fig:albenorm}
\end{figure}

To faithfully recover the color and texture for rendering applications, the albedo and normal should be consistent. In other words, the image synthesized using the albedo and normal should match the original image. However, as shown in Fig.~\ref{fig:albenorm}, due to the domain gap between real and synthetic data, the synthesized image (g) does not have a similar appearance as the original one (a). Another way to obtain consistent albedo is to use the normal $N_{p}$ and the input illumination to estimate the shading~\cite{ramamoorthi2001efficient} (c), and estimate the albedo (e) by dividing the image (a) by this normal estimated shading. This albedo (e) is consistent with the estimated normal $N_p$, and thus has the correct color and global scale. Yet it does not have a ``flat" appearance, which means there is residual shading information included due to incorrectly estimated normals. The estimated albedo $A_p$ in (d) on the other hand correctly factored out the shading, but is not consistent with the normal image. To consolidate the two estimates, we modify (d) by taking the color and scale from (e), to obtain an albedo (f) which is consistent with the normal image and at the same time has a ``flat" appearance. 

Concretely, let $I$ be the per-pixel intensity of R,G,B channels: $I=(R+G+B)/3$, and $med(I)$ be the median intensity within human mask. We first take the color from (e) as $R'=I_d/I_e\times R_e$ and globally scale it to (e) as $R = med(I_e)/med(I')\times R'$. $B$ and $G$ are obtained similarly. The resulting albedo (f) is consistent with the normal image (b) and the newly synthesized image (h) better matches the original image (a).

\subsection{Texture Generation}
\begin{figure}[t]
\centering
\includegraphics[width=\linewidth]{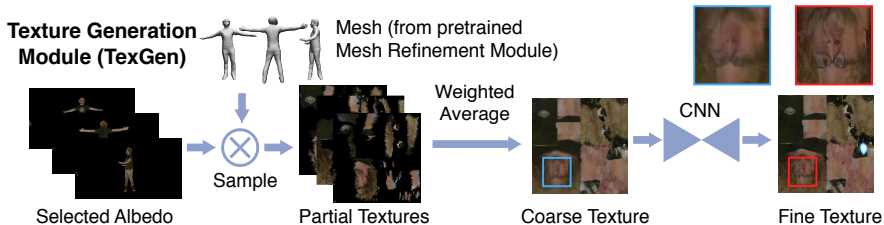}
\caption{Texture Generation Module (TexGen). TexGen selects $K$ albedo images and converts them into partial textures. A coarse full texture is constructed by averaging the partial textures with weights. The texture is then refined by a CNN optimized from scratch for each video. The supervision comes from rasterizing the texture to the image space and comparing it with the input albedo images.}
\label{fig:texture}
\end{figure}

TexGen is used to encode and refine the person texture. Inspired by TNA~\cite{shysheya2019textured}, the texture is obtained by optimizing the photometric loss between rendered images and the original ones. We propose to (1) use albedo instead of the original image to prevent shading leaking into texture, (2) select keyframes to mitigate geometric misalignment, and (3) use a CNN to parameterize the texture to reduce noise and artifacts. We assume the availability of MeshRef (Sec.~\ref{ssec:meshref}) pre-trained on synthetic data. Fig.~\ref{fig:texture} shows the TexGen pipeline.

\label{ssec:texgen}
\boldstart{UV Mapping.}
We follow the common practice in graphics where texture is stored using a UV map~\cite{blinn1976texture}. This map unwraps a mesh into 2D space, called UV space. Each pixel $\mathbf{t}=(u,v)$ in UV space corresponds to a point $\mathbf{v}$ on the mesh, and thus maps to a pixel $\mathbf{p}=(x_{2D},y_{2D})$ in the image via projection. Therefore, we can sample image features and convert them into UV space. For example, we can convert an albedo image into a partial texture via sampling and masking with visibility. See supplementary material for details.

\begin{figure}[t]
\centering
\begin{minipage}{0.16\linewidth}
\subfloat[(a) Ours]{
\includegraphics[width=0.9\linewidth]{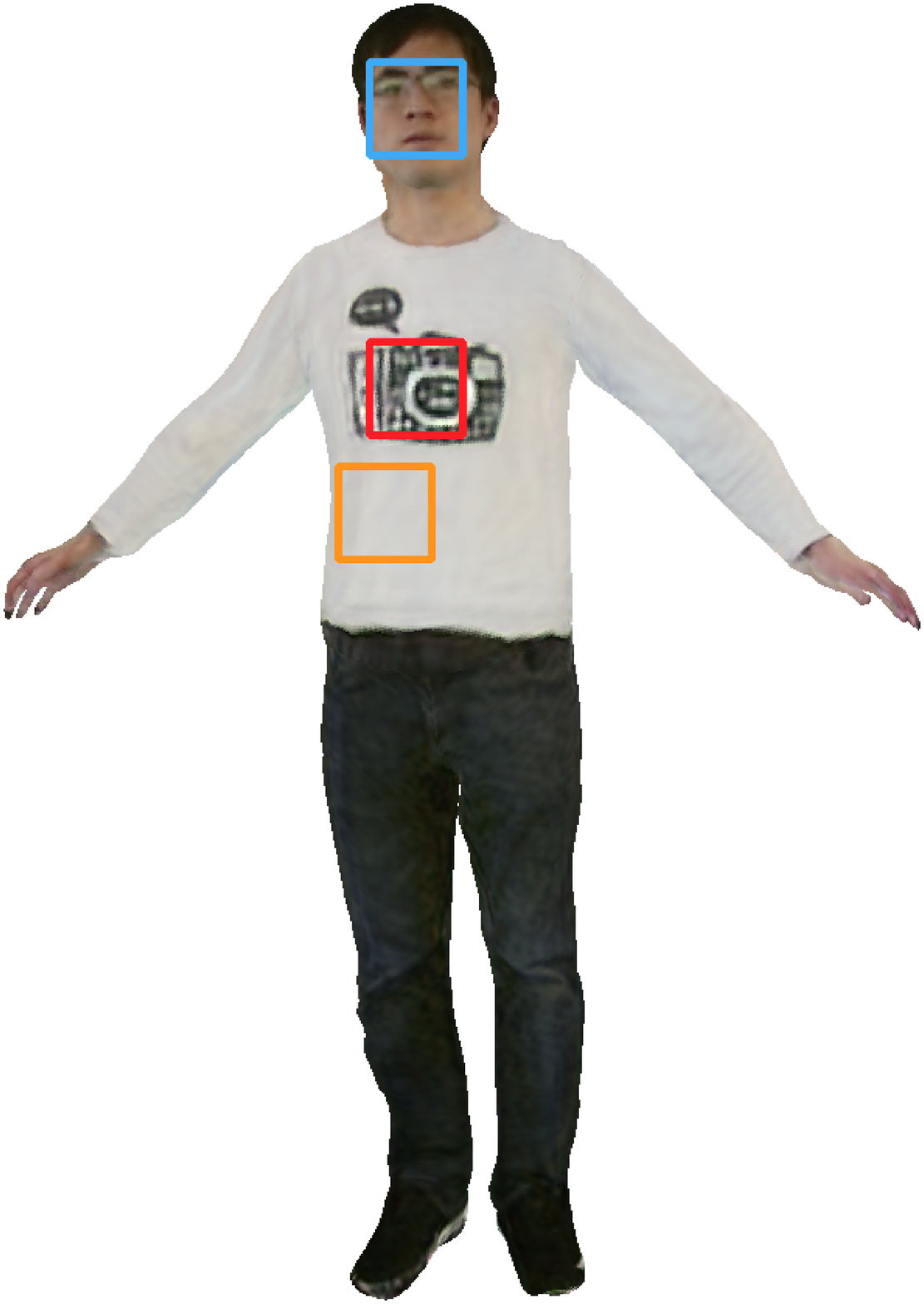}}
\end{minipage}
\begin{minipage}{0.82\linewidth}
\subfloat[(b)  Albedo image]{
\includegraphics[height=30pt]{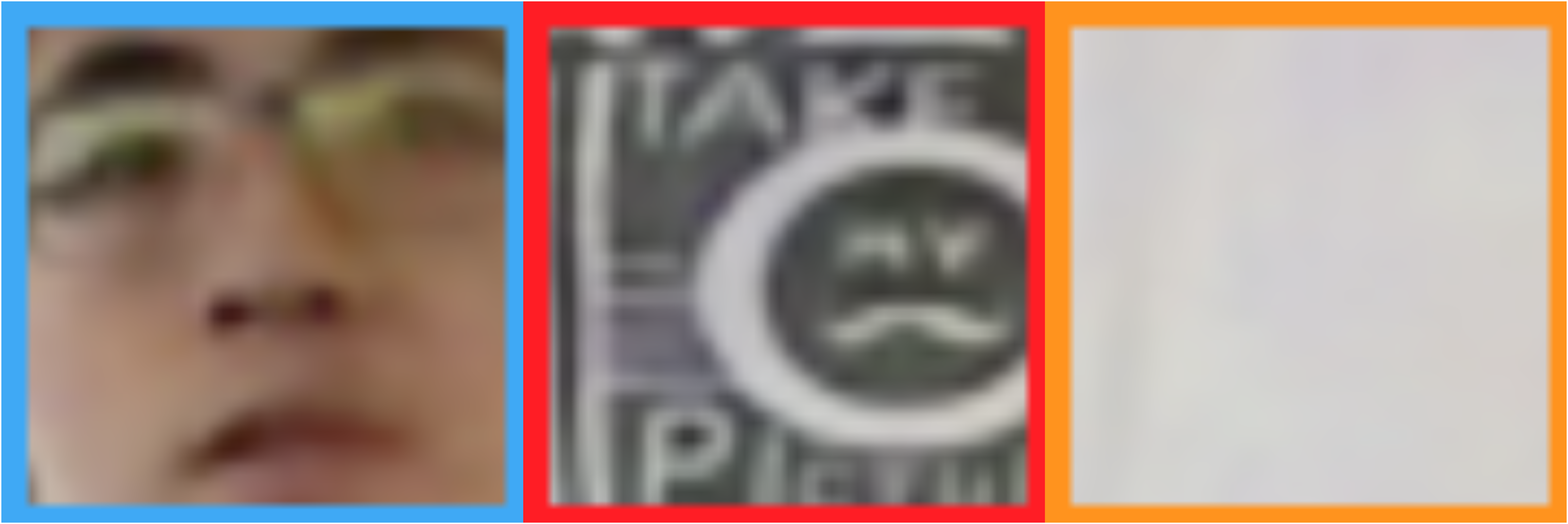}}
\subfloat[(c) Neural Avatar \cite{shysheya2019textured}]{
\includegraphics[height=30pt]{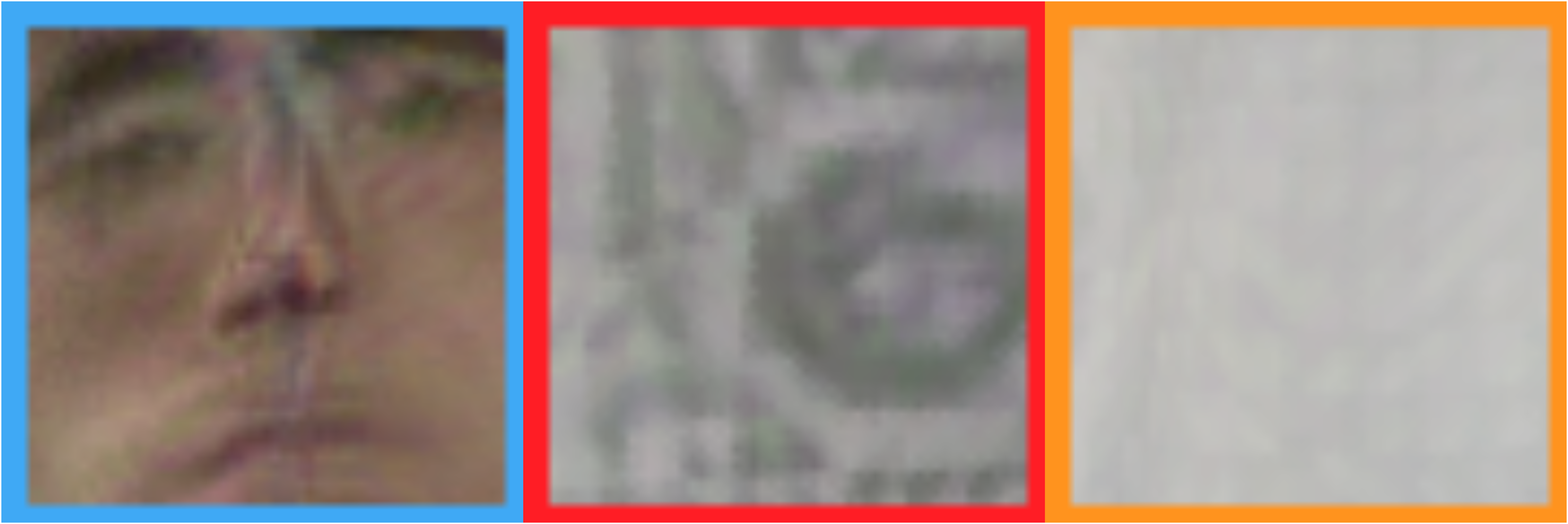}}
\subfloat[(d) Not train on albedo]{
\includegraphics[height=30pt]{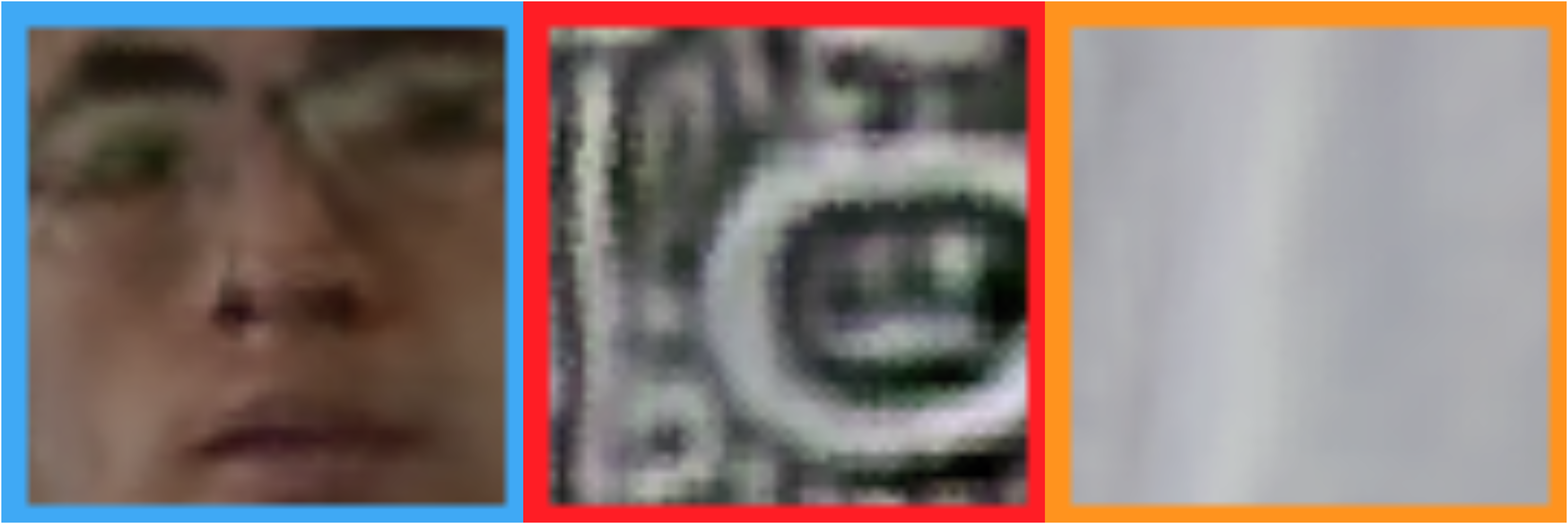}}
\\
\subfloat[(e) No frame selection]{
\includegraphics[height=30pt]{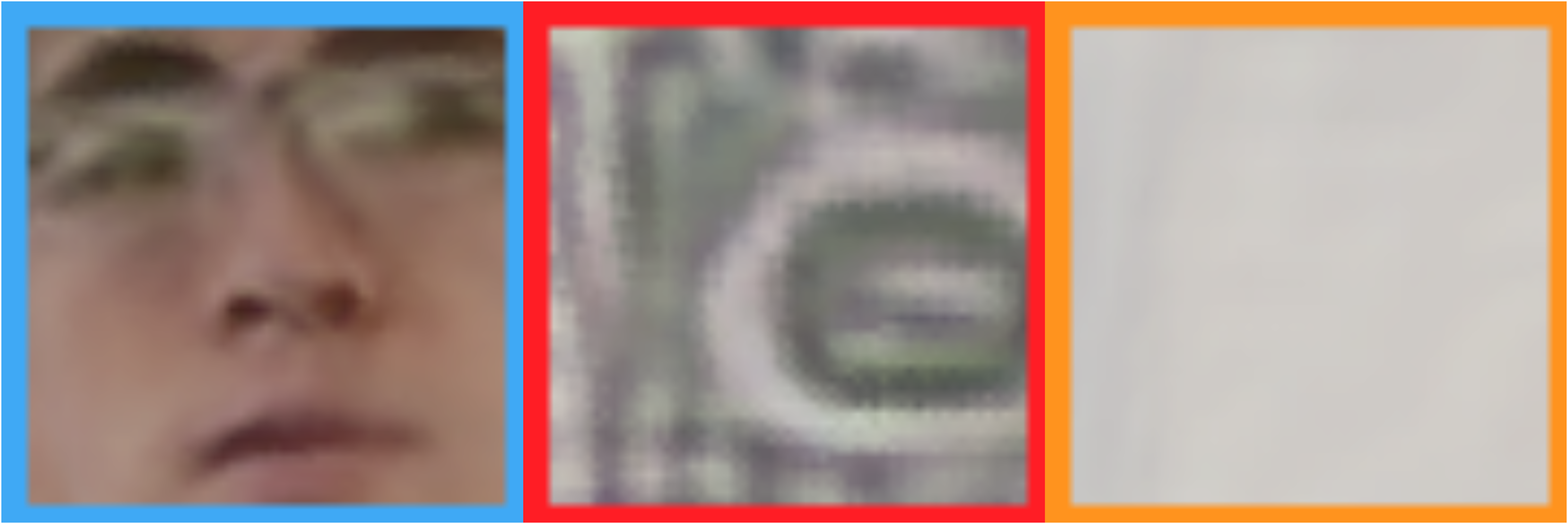}}
\subfloat[(f) Texture stack]{
\includegraphics[height=30pt]{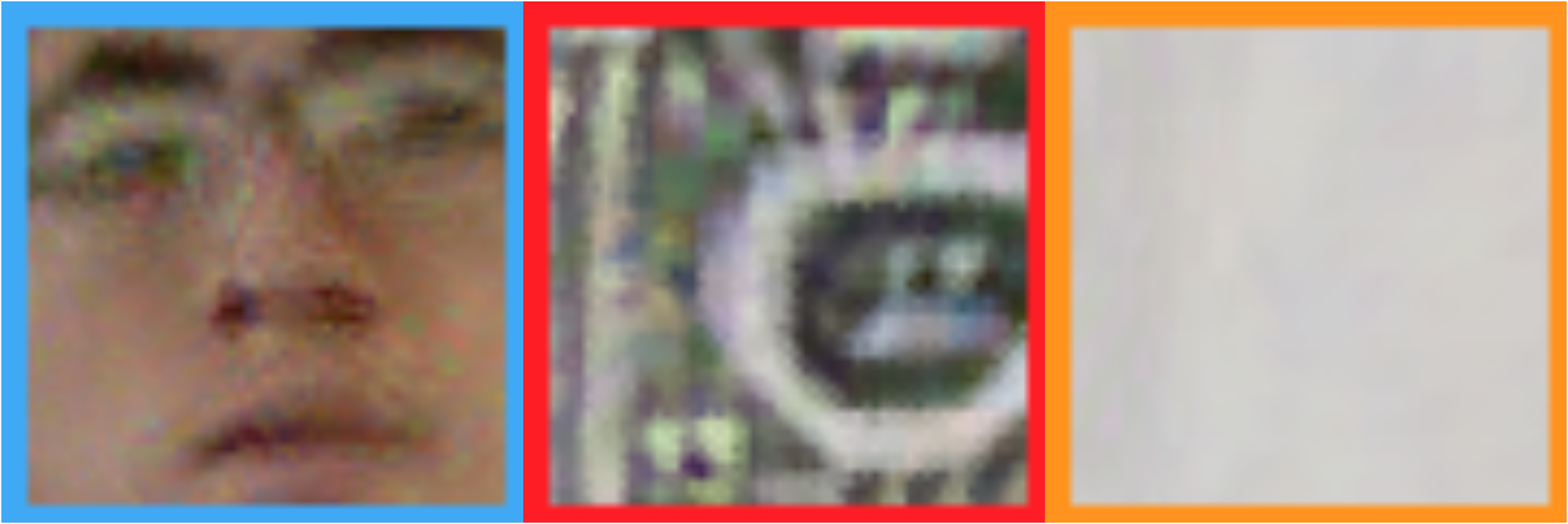}}
\subfloat[(g) Our full method]{
\includegraphics[height=30pt]{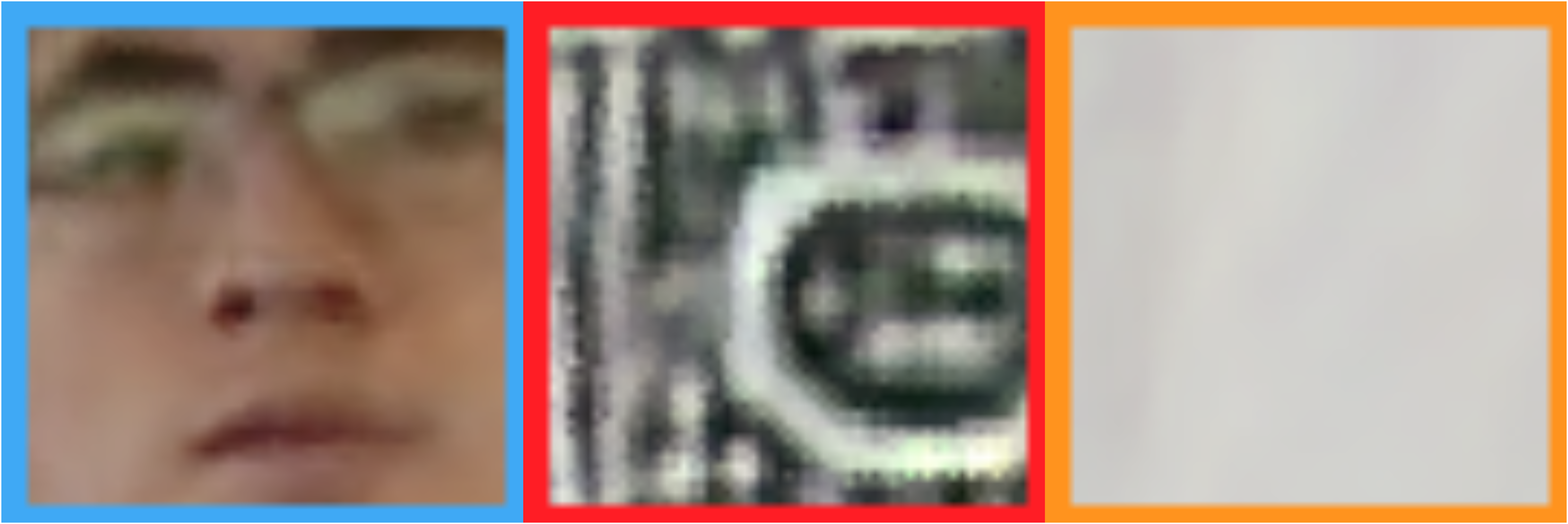}}
\end{minipage}
\caption{Rasterized albedo using generated texture. (b) is the albedo from AlbeNorm, which can be seen as ``ground truth". Training on the original image rather than albedo (d) causes the texture to include shading. No frame selection (e) makes the result blurry. Using a texture stack instead of a CNN (f) creates a noisy face. (c) is TNA~\cite{shysheya2019textured} (trained on original images using a texture stack, without frame selection). These issues are addressed by our full method (a)(g).}
\label{fig:comptex}
\end{figure}

\boldstart{Key Frame Selection.}
We aim to extract a sharp texture from a short video sequence. Inherently, this is difficult because of misalignments. We aim to address these issues through selection of a few, particularly well suited frames for this task. Our selection should cover the whole body using a small number ($K$) of albedo frames based on the visibility of the partial texture image. More concretely, let $V_i$ be the visibility UV map for the $i$-th selected frame and $V_0$ be the visibility map of the rest pose mesh. Since most salient parts (e.g., faces) are visible in the rest pose, we first select the frame closest to the rest pose by minimizing $||V_1-V_0||_{1}$. We then greedily add the $i$-th frame by maximizing the total visibility $||\max_{j=1}^{i}{V_j}||_{1}$, until $K$ frames are selected. We also assign a sampling frequency weight of $W_i = 1 / K + w_i / \sum_{i=1}^K{w_i}$, where $w_i = ||V_i - \max_{j=1}^{i-1}{V_j}||_{1}$, to every $i$-th frame. These weights bias the training to more visible frames and speed up the convergence. In practice, we add two adjacent frames with $w=w_i/2$ of the $i$-th frame for denoising. Fig.~\ref{fig:comptex} shows the benefit of our selection scheme which leads to more detailed and accurate reconstructions.

\boldstart{Texture Refinement CNN.}
From the key albedo frames, we generate partial textures and obtain a coarse texture using a weighted average, where the weight is $w_i V_i$. The coarse texture is processed by a CNN for generating a fine texture, using a deep image prior~\cite{ulyanov2018deep} for denoising and artifact removal (see faces in Fig.~\ref{fig:comptex} (f) and (g) for the benefit of our CNN-based texture paramaterization over the texture stack of TNA~\cite{shysheya2019textured}). The loss comes from rasterizing the texture to the image space and comparing it with the selected albedo images. The gradients are also back-propagated to the mesh and thus MeshRef for a better alignment. This mesh adjustment scheme is a crucial component of the TexGen module.

Concretely, let $R$ be the albedo image rasterized using SoftRas \cite{liu2019soft}, $A$ be the albedo image from AlbeNorm, and $M$ be the human mask from segmentation~\cite{chen2017rethinking,piccardi2004background}. We use an $L_1$ photometric loss and a perceptual loss~\cite{johnson2016perceptual} to compare $R$ and $A$ within $M$. We further regularize the mesh deformation by an $L_1$ loss between the Laplacian coordinates~\cite{sorkine2006differential} of the current vertices and the vertices from the initial pre-trained MeshRef model. 
Our total loss $L_{TG}$ is written as:
\begin{equation}
    \label{loss:TG}
    L_{TG} = \lambda^{tg}_{L1} ||(R - A) \cdot M||_{1} + \lambda^{tg}_{pct} l_{pct}(R \cdot M, A \cdot M) + \\ \lambda^{tg}_{lap} \sum\limits_{i\in V}||\mathbf{v}'_{p,i} - \mathbf{v}'_{o,i}||_{1},
\end{equation}
where $V$ is the vertex index set and $\mathbf{v}'_{p,i}$ and $\mathbf{v}'_{o,i}$ are the Laplacian coordinates of the $i$-th predicted vertex and original vertex, respectively.  $l^{tg}_{pct}(x, y)$ computes an adaptive robust loss function~\cite{barron2019general} over the VGG features \cite{simonyan2014very} as perceptual loss \cite{johnson2016perceptual}. $\lambda^{tg}_{L1}$, $\lambda^{tg}_{pct}$, and $\lambda^{tg}_{lap}$ are weights. In practice, we empirically limit the deformation of small structures such as the head and hands by locally disabling gradients, because of possible large mesh registration errors in these regions. 

\subsection{Mesh Refinement}
\label{ssec:meshref}
\begin{figure}[t]
\centering
\includegraphics[width=\linewidth]{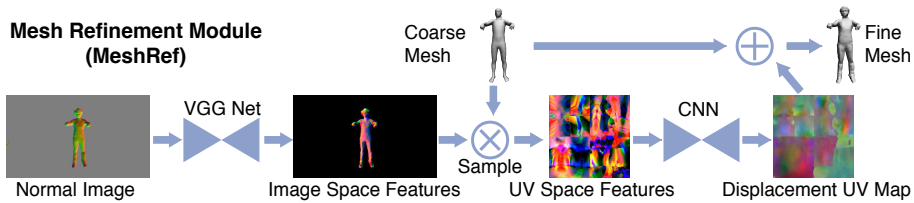}
\caption{Mesh Refinement Pipeline (MeshRef). MeshRef first extracts features from a normal image, then convert those features to UV space. The UV space features are then sent to a CNN to predict a 3D displacement map. We obtain the fine mesh by adding the displacements to the coarse mesh. This module is first trained using synthetic data with ground truth and later self-adapted on a short real sequence via photometric and temporal losses.}
\label{fig:refine}
\end{figure}
The MeshRef module is used to refine the coarse mesh. Fig.~\ref{fig:refine} gives an overview of its design. Inspired by the effectiveness of predicting human shape deformation in UV space~\cite{alldieck2019tex2shape}, MeshRef converts the image features into UV space to predict 3D displacement vectors. Our design takes the normal map from AlbeNorm and extracts VGG features~\cite{simonyan2014very} to obtain a better encoding, before converting the features to UV space. The features can be further augmented by other information such as vertex positions (see supplementary material for details).

To learn human shape priors, we pre-train MeshRef on a synthetic dataset with supervision. However, due to the domain gap, the pre-trained model does not perform well on real data. Thus, after obtaining the texture from TexGen, we adapt MeshRef on a real sequence using a photometric loss between the textured mesh and the original image. We also apply a motion prior loss~\cite{Vo_2016_CVPR} to enhance short-term temporal consistency. Since these losses cannot provide tight supervision for invisible regions, we further use a deformation loss to propagate the deformation from frames where those regions are visible to frames where they are not. This model is trained on batches of consecutive video frames. 



\boldstart{Supervised Training on Synthetic Images.} We supervise the 3D displacement maps using $L_1$ and SSIM losses and regularize the 3D vertices using a Laplacian loss. Let $D_{p}$ and $D_{g}$ be the predicted and ground truth displacement maps and $DSSIM=(1-SSIM)/2$ be the structural dissimilarity function~\cite{wang2004image}. Our loss $L_{MR1}$ is defined as:
\begin{equation}
    L_{MR1} = \lambda^{mr1}_{L1} ||D_{p} - D_{g}||_{1} + \lambda^{mr1}_{ssim} DSSIM(D_{p}, D_{g}) + \lambda^{mr1}_{lap} \sum\limits_{i\in V}||\mathbf{v}'_{p,i} - \mathbf{v}'_{g,i}||_{1},
\end{equation}
where $\mathbf{v}'_{p,i}$ and $\mathbf{v}'_{g,i}$ are Laplacian coordinates defined similar to Eq.~\ref{loss:TG}, and $\lambda^{mr1}_{L1}$, $\lambda^{mr1}_{ssim}$, and $\lambda^{mr1}_{lap}$ are the weights between different losses.

\boldstart{Self-supervised Training on Real Video Data.} 
For self-supervised training, we render the images using the SoftRas differentiable renderer~\cite{liu2019soft} and compare with the original images. Our self-supervised loss is defined as:
\begin{equation}
\begin{aligned}
    L_{MR2} = &\lambda^{mr2}_{pct}L_{pct} + \lambda^{mr2}_{sil}L_{sil} + \lambda^{mr2}_{temp}L_{temp} + \\ & \lambda^{mr2}_{pos}L_{pos}  + \lambda^{mr2}_{lap}L_{lap} + \lambda^{mr2}_{deform}L_{deform} ,
\end{aligned}
\end{equation}
where $L_{pct}, L_{sil}, L_{temp}, L_{pos}, L_{lap}, L_{deform}$ are the perceptual loss, the silhouette loss, the motion consistency loss, the vertex position loss, Laplacian loss, deformation propagation loss, and $\lambda^{mr2}_{pct}, \lambda^{mr2}_{sil}, \lambda^{mr2}_{temp},\lambda^{mr2}_{pos}, \lambda^{mr2}_{lap}, \lambda^{mr2}_{deform}$ are their corresponding weights, respectively. We introduce the losses below. For simplicity, we present the losses for one frame, omitting the summation over all frames.

\boldstart{Perceptual Loss.} Let $R$ be the rendered image, $I$ be the original image, $M_R$ be the rasterized silhouette and $M_I$ be the segmented human mask, and $M=M_R \cdot M_I$. The loss is defined as $L_{pct} = l_{pct}(R \cdot M, I \cdot M)$, where $l_{pct}$ is the robust perceptual loss function~\cite{barron2019general,johnson2016perceptual}.

\boldstart{Silhouette Loss.} This loss compares the rasterized silhouette and the segmented human mask is defined as $L_{sil} = ||(M_R - M_I) \cdot C||_{1}$, where $C$ is the confidence map given by the segmentation algorithm~\cite{chen2017rethinking,piccardi2004background}. 

\boldstart{Motion Consistency Loss.} Let $t$ be the current frame index and $\mathbf{v}^{(t)}_{p,i}$ be the position of the $i$-th vertex in frame $t$. Our motion consistency loss favors constant velocity in adjacent frames~\cite{Vo_2016_CVPR} and is written as $L_{temp} = \sum\limits_{i\in V}||\mathbf{v}^{(t-1)}_{p,i} + \mathbf{v}^{(t+1)}_{p,i} - 2 \mathbf{v}^{(t)}_{p,i} ||_{1}$, where V is the set of vertex indices.

\boldstart{Vertex Position Loss.} This loss prevents large deformation from the original position predicted by the model pre-trained on synthetic data and is defined as  $L_{pos} = \sum\limits_{i \in V'}||\mathbf{v}_{p,i} - \mathbf{v}_{o,i}||_{2}$, where $\mathbf{v}_{p,i}$ and $\mathbf{v}_{o,i}$ are the positions of the $i$-th predicted vertex and original vertex, and $V'$ be the set of visible vertex indices. 

\boldstart{Laplacian Loss.} This loss is not only applied to visible vertices but also head and hand vertices regardless of their visibility because noisy deformation of these vertices can significantly affect the perceptual result, and is defined as $ L_{lap} = \sum\limits_{i \in V}||(\mathbf{v}'_{p,i} - \mathbf{v}'_{o,i}) \cdot u_i ||_{1}$, where $\mathbf{v}'_{p,i}$ and $\mathbf{v}'_{o,i}$ are the Laplacian coordinates of the $i$-th predicted and original vertices, and $u_{i}$ be the weight of the $i$-th vertex. We set $u_i=100$ for head and hand, $1$ for other visible vertices, and $0$ for the rest.

\begin{figure}
\centering
\includegraphics[width=\linewidth]{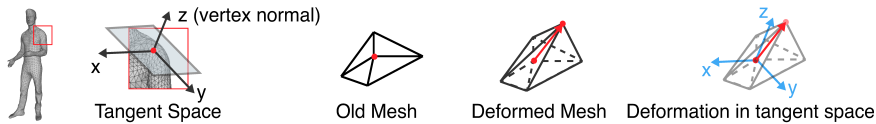}
\caption{Deformation in tangent space. At a local point, the $z$-axis points to the vertex normal direction and the $x$ and $y$ axes complete the orthogonal basis, forming a local coordinate system. We use this coordinate system to represent the vertex deformation. This representation is invariant to pose change, propagating deformation of the same vertex in different frames.}
\label{fig:tangent}
\end{figure}
\begin{figure}[b]
\centering
\subfloat[(a) Source frame]{
\includegraphics[height=0.2\linewidth]{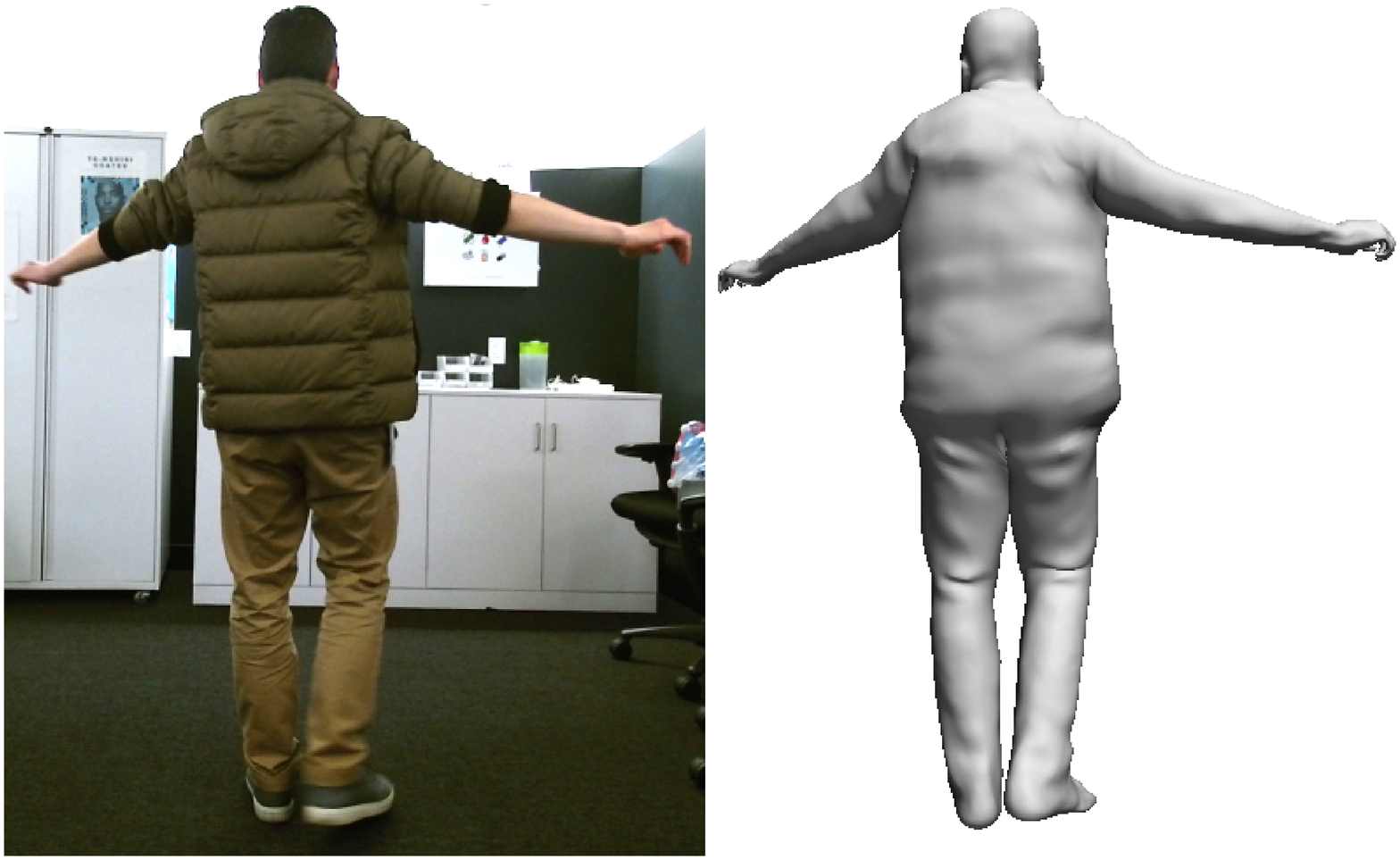}}
\subfloat[(b) Target]{
\includegraphics[height=0.2\linewidth]{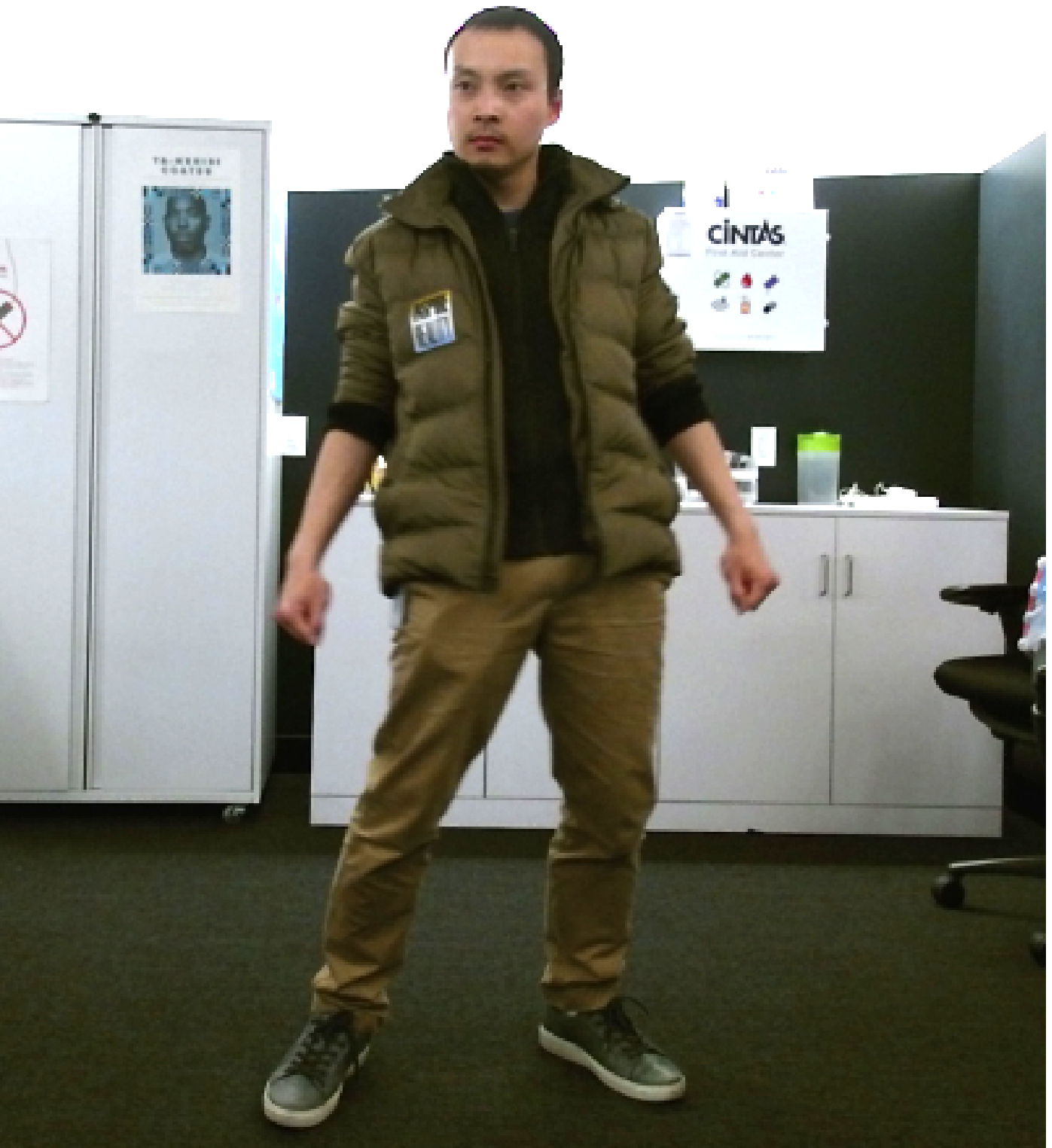}}
\subfloat[(c) No propagation]{
\includegraphics[height=0.2\linewidth]{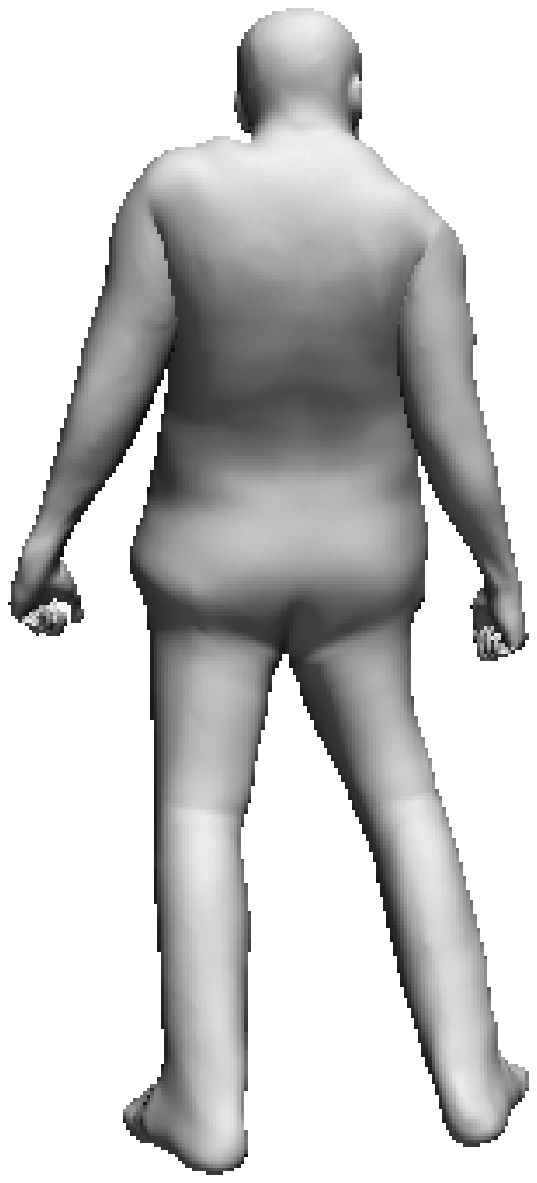}}
\subfloat[(d) Propagation]{
\includegraphics[height=0.2\linewidth]{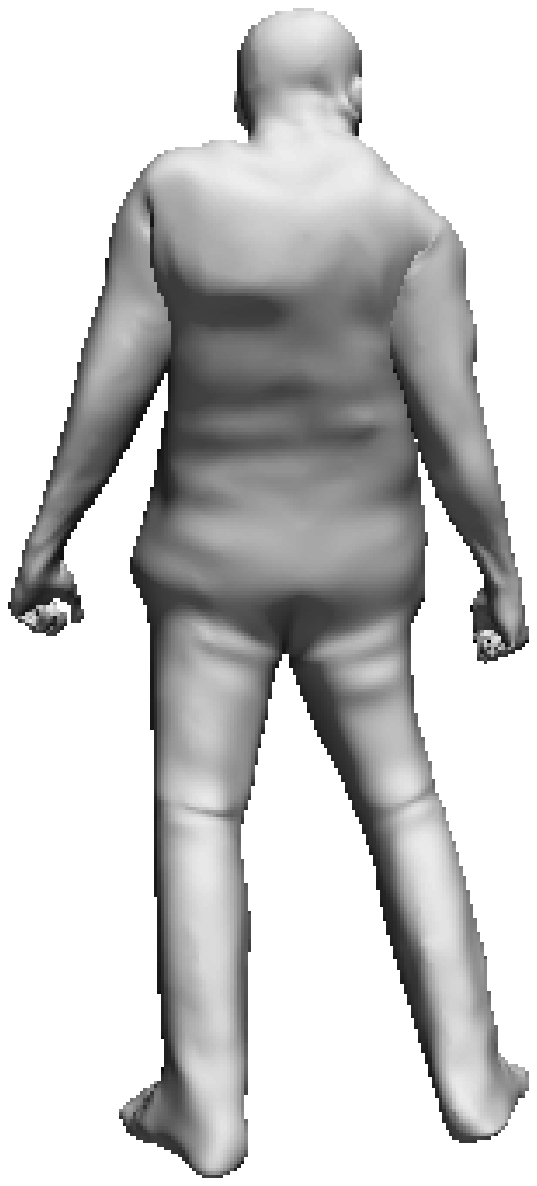}}

\caption{Effect of deformation propagation. (a) is a source frame propagating deformation to frame (b) where the back is not visible. The back of (b) is reconstructed without (c) and with (d) deformation propagation. The one with propagation shows more clothing details.}
\label{fig:deform}
\end{figure}

\boldstart{Deformation Propagation Loss.} To reconstruct an vertex invisible in the current frame, we find a visible counterpart in the set of keyframes computed in Sec. \ref{ssec:texgen} and propagate the deformation from it. This is similar in spirit to the canonical shape model~\cite{newcombe2015dynamicfusion}. However, because the human in the source frame and target frame may have different poses, we can not simply force the deformation in the global coordinates to be similar. We adopt the local tangent space \cite{lengyel2012mathematics} (Fig.~\ref{fig:tangent})
to solve this problem.

Let $\mathbf{d}^{(s)}_{i}$ and $\mathbf{d}^{(t)}_{i}$ be the deformation in tangent space of the $i$-th source vertex visible in one of the selected keyframes and occluded target vertex at the current frame. The deformation loss is defined as $L_{deform} = \sum\limits_{i \in V''}||\mathbf{d}^{(t)}_{i} - \mathbf{d}^{(s)}_{i}||_{1}$, where $V''$ is the set of invisible vertex indices in target frame. $\mathbf{d}^{(s)}_{i}$ does not receive gradients, and is stored in a buffer updated when the source frame is sampled during training. In practice, we extend $V''$ to include head vertices, to enhance head rigidity. Our deformation propagation scheme provides more realistic details on invisible surfaces as shown in Fig.~\ref{fig:deform}.

\section{Experimental Analysis}
\subsection{Implementation Details}
\noindent\textbf{Deep Networks.} CNNs are based on U-Net~\cite{ronneberger2015u}, optimized using Adam~\cite{kingma2014adam}. The full-frame image resolution is 960$\times$540 with the human region resolution around 200$\times$430. Texture resolution is 512$\times$512. Details are in supplementary.

\noindent\textbf{Human Model.} The full-body human model is a variant of SMPL~\cite{loper2015smpl}. The original SMPL model has about 7k vertices, which is too large for the coarse mesh representation, and insufficient to represent fine details such as clothing wrinkles. Thus, we construct a body model with two levels of resolution: the coarse level has 1831 vertices and 3658 faces, and the fine level has 29,290 vertices and 58,576 faces obtained by subdividing the coarse mesh topology.
The vertices of the coarse mesh are a subset of the vertices of the fine mesh and share identical vertex indices.
Both representations also share a unique skeletal rig that contains 74 joints. This design reduces the overhead for generating the coarse mesh, and preserves the fine mesh capability to represent geometric details.


\subsection{Datasets}
Our method requires lighting information, which is not provided by most public datasets.
Thus we capture and render our own data to perform experiments.

\noindent\textbf{Synthetic Images for Pre-training.}
We synthesize 18,282 images using 428 human 3D scans from RenderPeople\footnote{http://renderpeople.com/} and Our Dataset under the lighting from Laval Dataset \cite{gardner2017learning}.
Our Dataset was captured with a 3dMD scanner and contains 48 subjects.
We registered the fine level Human Model to the 3D scans using non-rigid ICP~\cite{li2008global}, initialized with a 3D pose estimator~\cite{bogo2016keep}. To generate the coarse mesh, Gaussian noise scaled by a random factor sampled from uniform distribution is added to the pose and shape parameters, and the position of the character.
The registered model can be set in arbitrary pose with skeletal rig. We render the 3D scans into images of various poses sampled from the Motion Capure data. No video sequences are synthesized due to its high computational demand. Our final dataset contains coarse meshes, fine meshes, displacement maps, environment maps, RGB images, albedos, normals, and human masks.

\noindent\textbf{Synthetic Videos for Quantitative Evaluation.}
We synthesize 6 videos with ground truth measurements that contain dynamic clothing deformation for higher realism. Our clothing is modeled as separate meshes on top of human body scans as in DeepWrinkles~\cite{lahner2018deepwrinkles}. However, we obtain deformation by physics-based simulation. We use the human bodies from AXYZ dataset\footnote{http://secure.axyz-design.com/} and the lighting from HDRI Heaven\footnote{http://hdrihaven.com/}. The videos represent subjects performing different motions such as walking or dancing and has about 3.8k frames each. In each video, we use about half of the frames for model adaptation and do inference on the rest. We treat the naked body as coarse mesh and the clothed body as fine mesh.

\noindent\textbf{Real Videos for Qualitative Evaluation.} We capture 8 videos ($\sim$4min each) using a Kinect along with lighting captured using a ThetaS. The cameras are geometrically and radiometrically calibrated. We use the first 2k frames for model adaption and infer on the whole video. We obtain the coarse mesh in real-time by solving an inverse kinematic problem to fit the posed body shape to the 3D point cloud and detected body keypoints~\cite{cao2018openpose} using an approach similar to~\cite{walsman2017dynamic}.

\subsection{Results}
\noindent\textbf{Texture.}
\label{ssec:texexp}
We compare our texture with a sampling-based method (SBM)~\cite{alldieck2018video} variant and a Textured Neural Avatars (TNA)~\cite{shysheya2019textured} variant re-implemented with our Human Model, which map between image and UV spaces using the mesh.
We render albedo images on synthetic videos, and evaluate average RMSE within valid mask and MS-SSIM~\cite{wang2004image} within human bounding box over subsampled videos. Our method outperforms SBM and TNA on (RMSE, MS-SSIM): (0.124, 0.800) for SBM, (0.146, 0.809) for TNA, and (\textbf{0.119, 0.831}) for ours, respectively. See Fig.~\ref{fig:comptex} and the supplementary material for qualitative results.
\begin{figure}[t]
\centering
\begin{minipage}{0.69\linewidth}
{
\includegraphics[height=75pt]{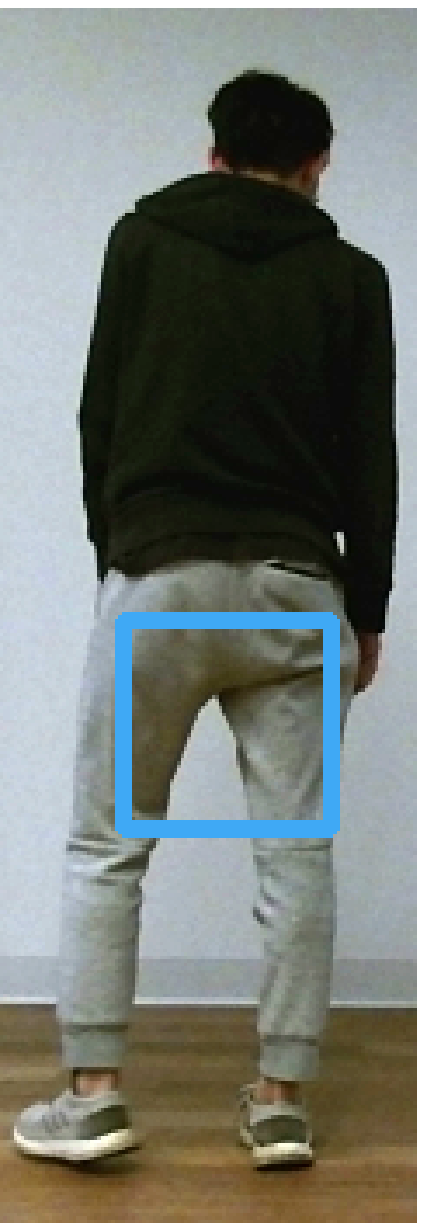}\quad\quad
\includegraphics[height=75pt]{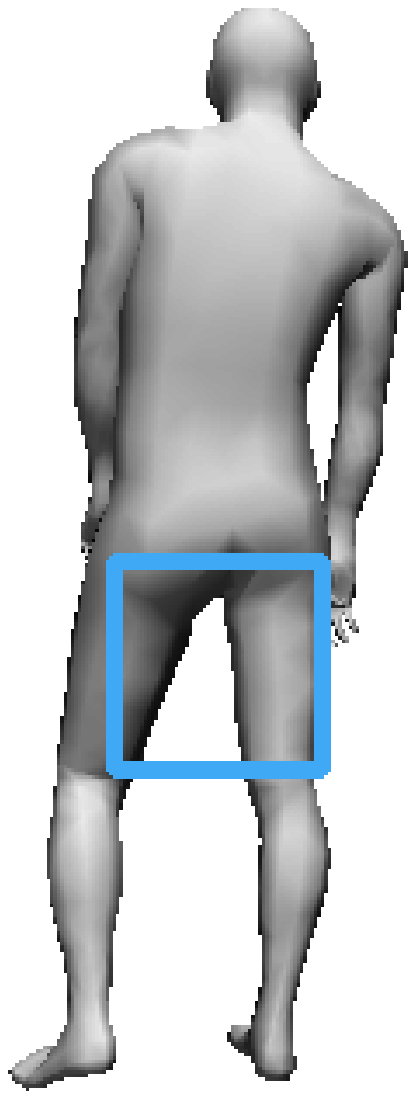}\quad
\includegraphics[height=75pt]{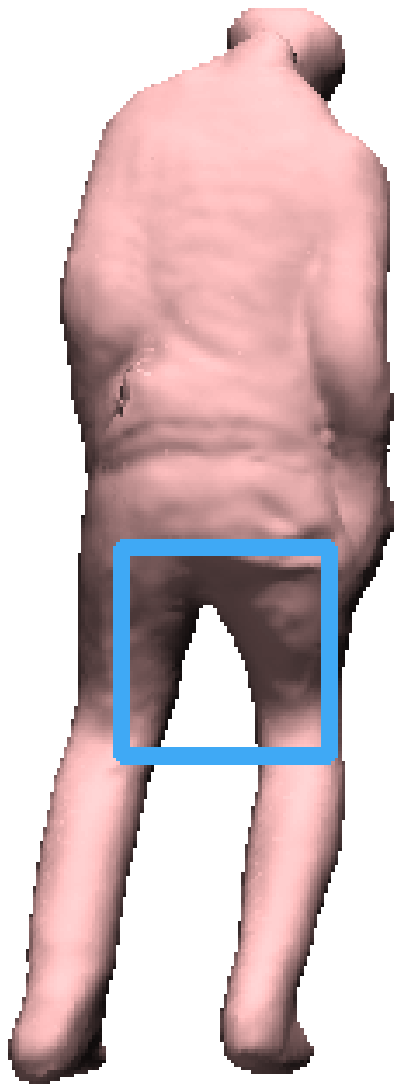}\quad
\includegraphics[height=75pt]{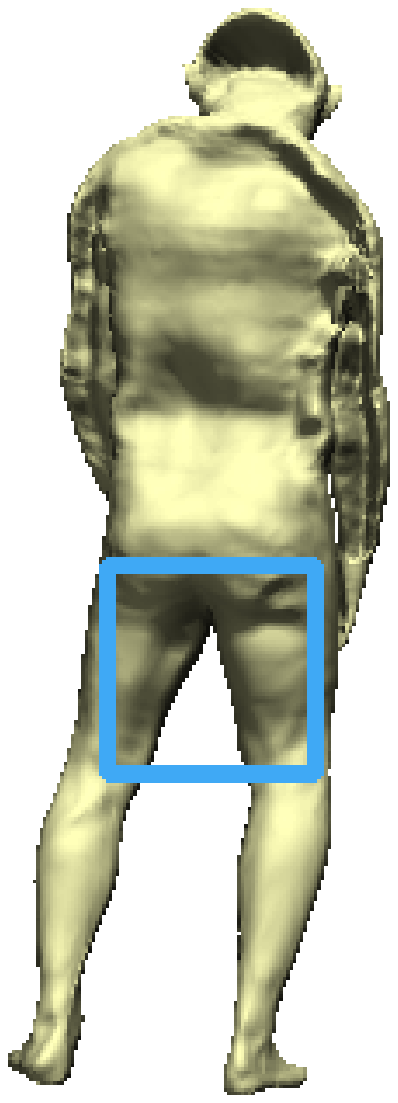}\quad
\includegraphics[height=75pt]{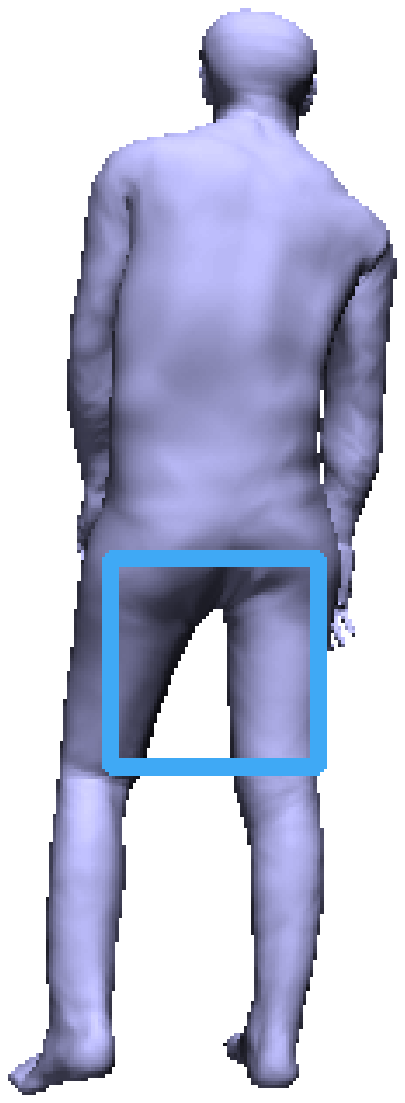}\quad
\includegraphics[height=75pt]{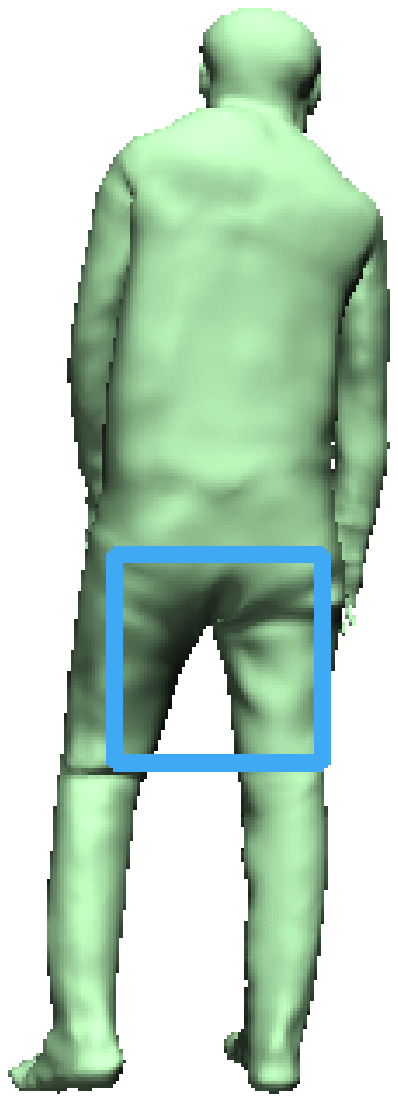}
}
\end{minipage}
\begin{minipage}{0.3\linewidth}
\centering
\includegraphics[height=30pt]{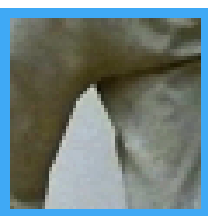}
\includegraphics[height=30pt]{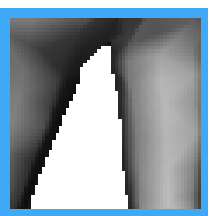}
\includegraphics[height=30pt]{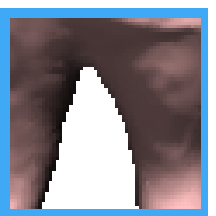}
\includegraphics[height=30pt]{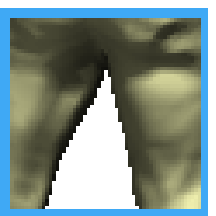}
\includegraphics[height=30pt]{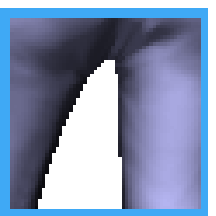}
\includegraphics[height=30pt]{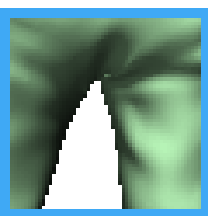}
\end{minipage}

\begin{minipage}{0.69\linewidth}
\includegraphics[height=69pt]{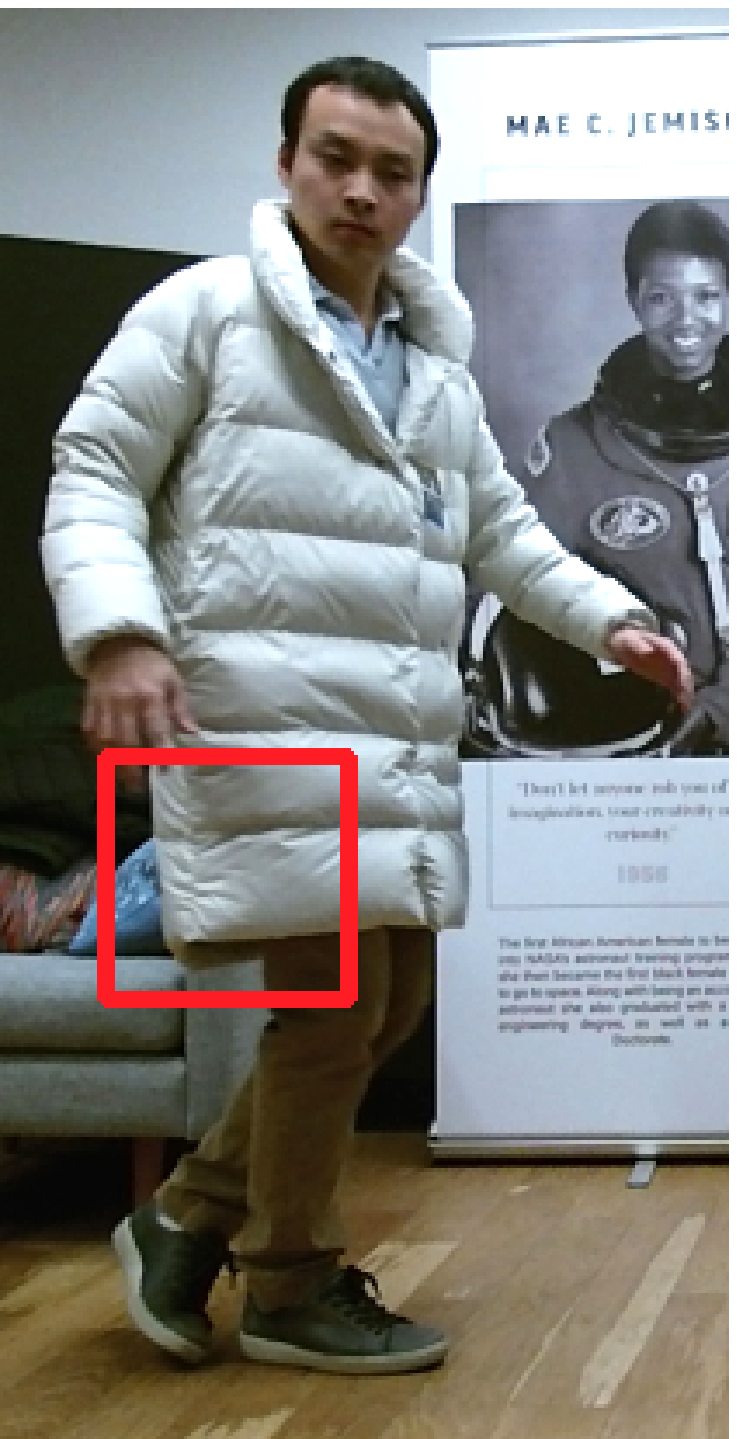}
\includegraphics[height=69pt]{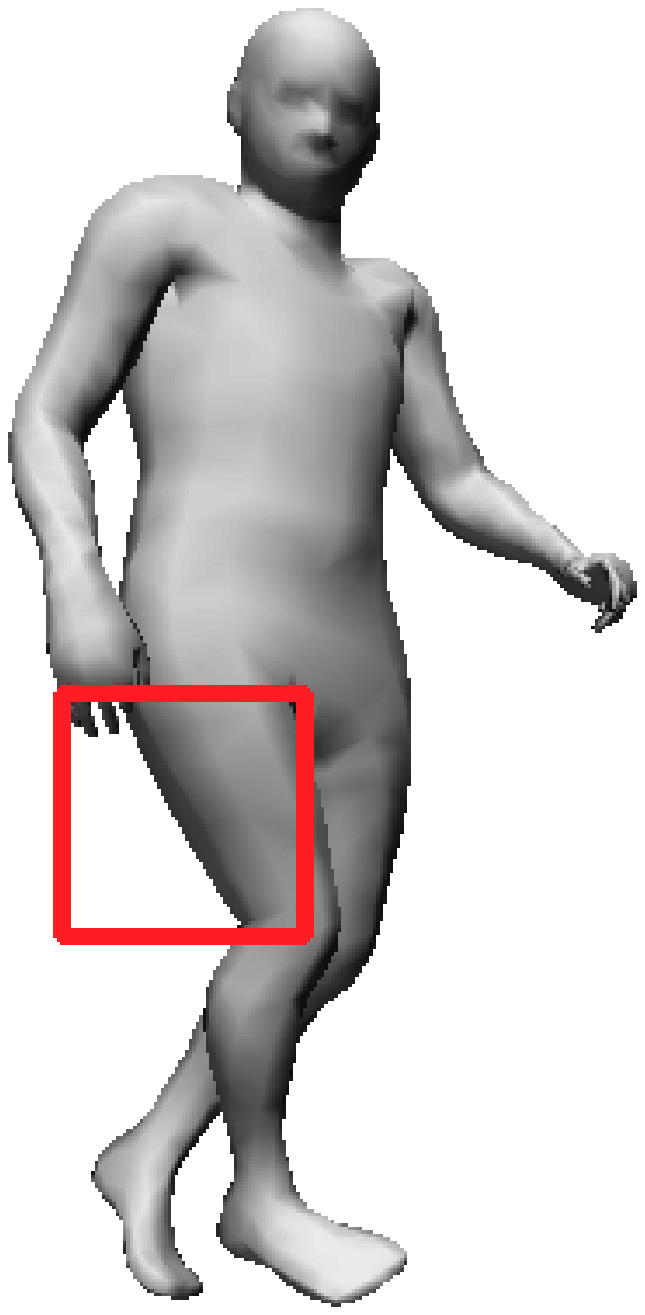}
\includegraphics[height=69pt]{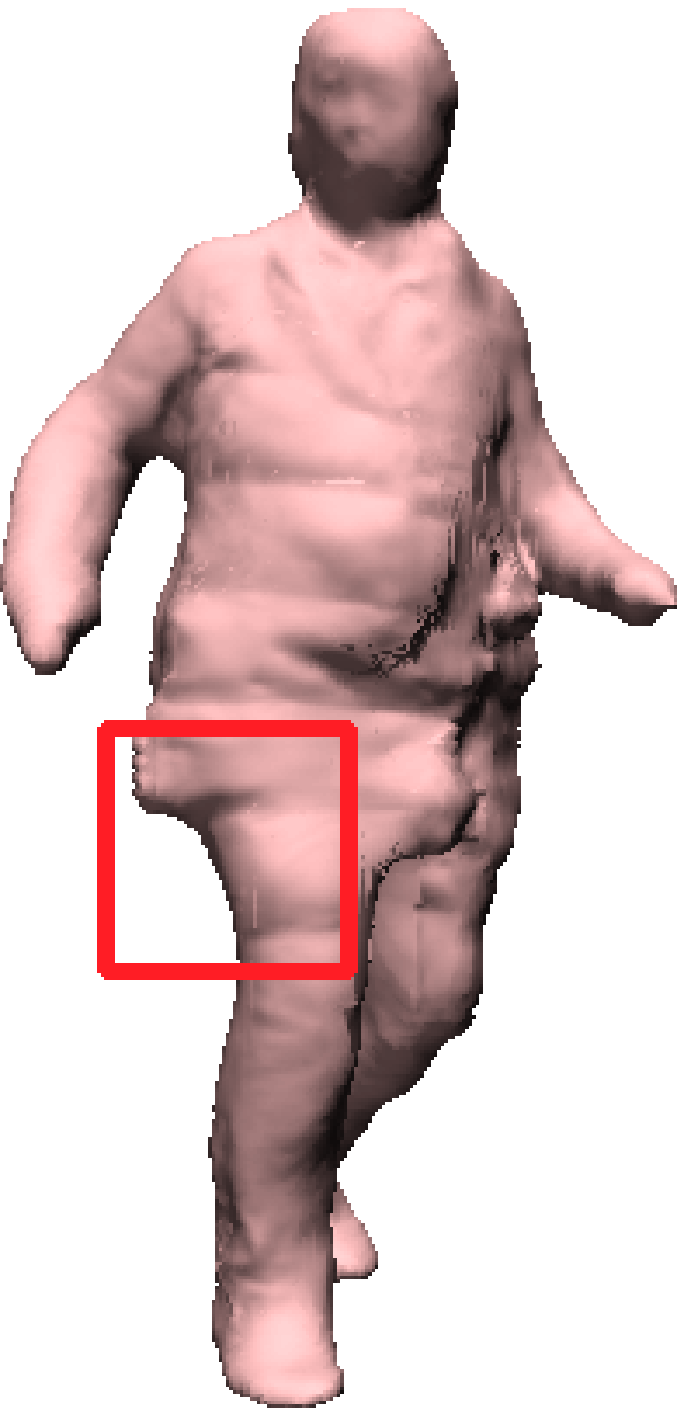}
\includegraphics[height=69pt]{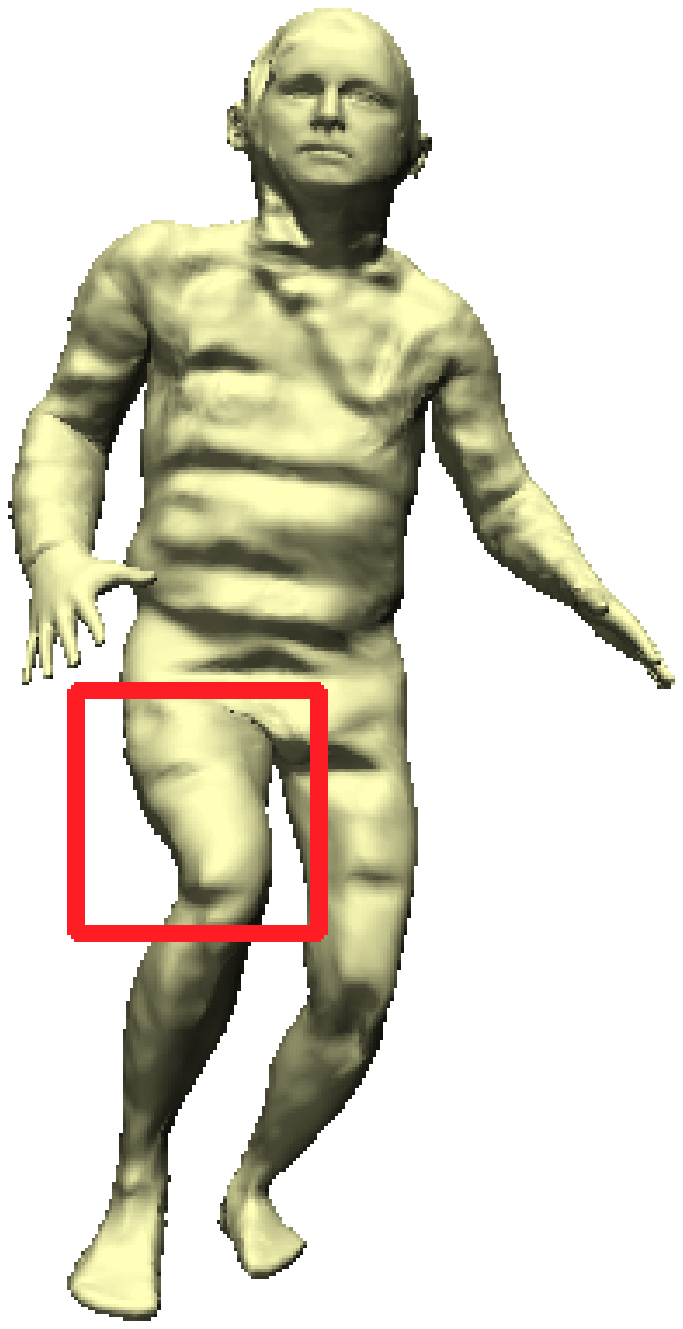}
\includegraphics[height=69pt]{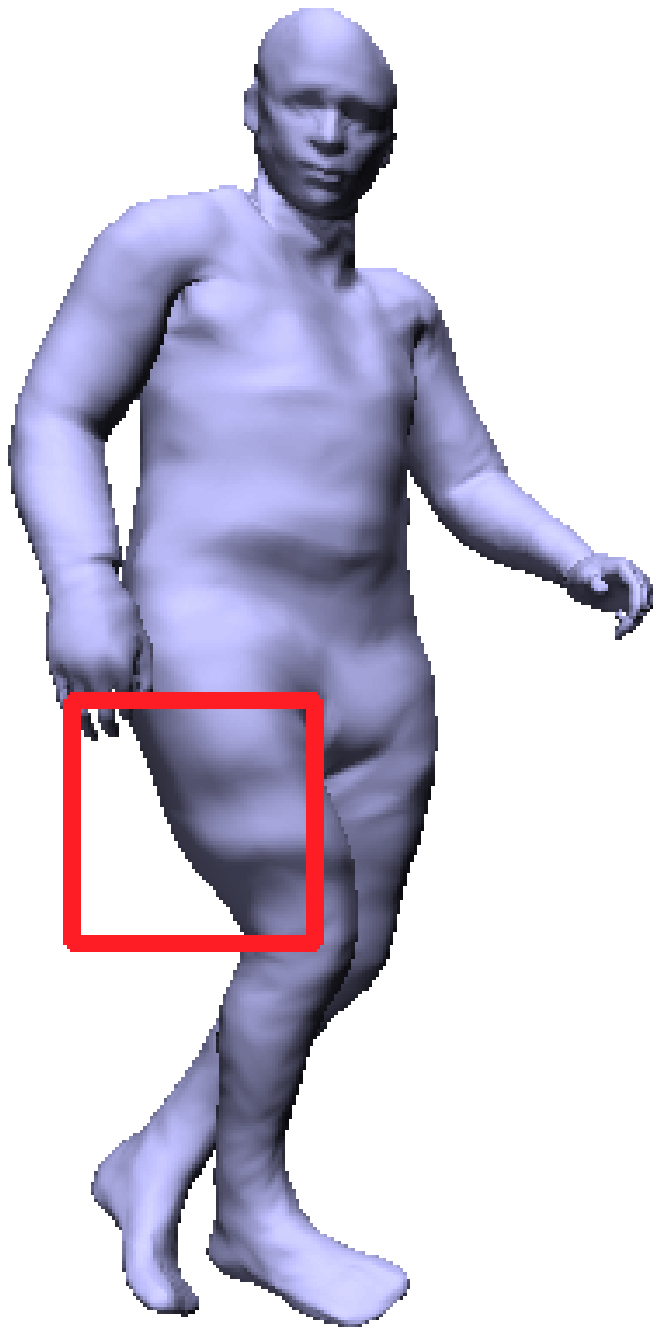}
\includegraphics[height=69pt]{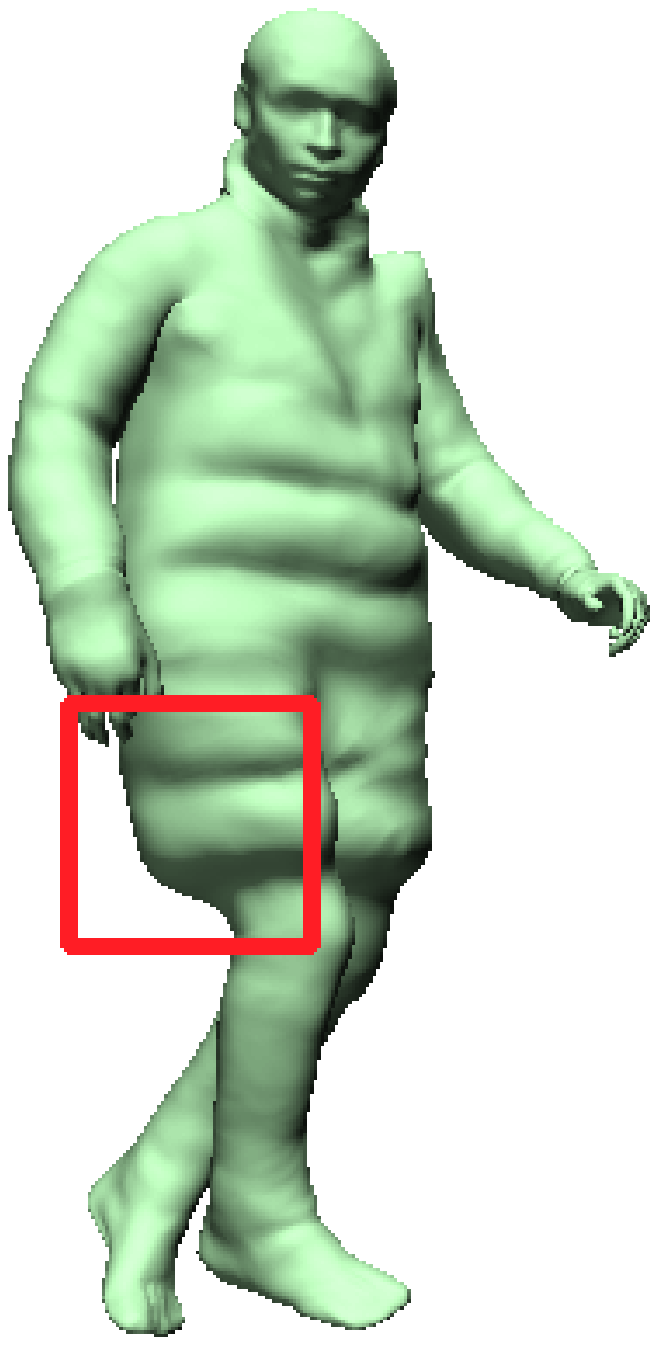}
\end{minipage}
\begin{minipage}{0.3\linewidth}
\centering
\includegraphics[height=30pt]{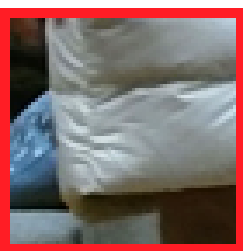}
\includegraphics[height=30pt]{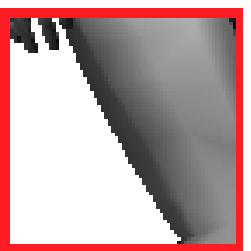}
\includegraphics[height=30pt]{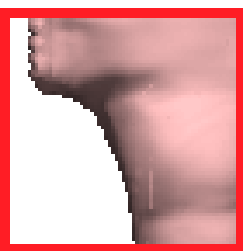}
\includegraphics[height=30pt]{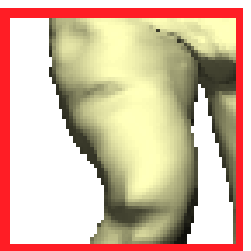}
\includegraphics[height=30pt]{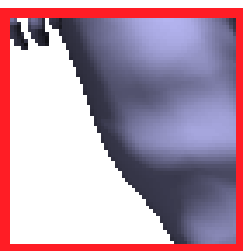}
\includegraphics[height=30pt]{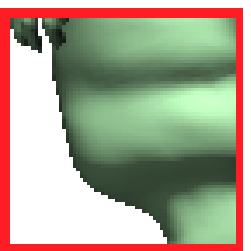}
\end{minipage}
\caption{Comparing mesh with coarse mesh (grey), DeepHuman~\cite{zheng2019deephuman} (red), HMD~\cite{zhu2019detailed} (yellow), Tex2Shape~\cite{alldieck2019tex2shape}. Our method (green) outperforms them in shape and details: DeepHuman is coarse in head and hands. HMD has artifacts in head and body regions. Tex2Shape does not obtain realistic wrinkles.}
\label{fig:compare}
\end{figure}
\setlength{\tabcolsep}{4pt}
\begin{table}[t]
\begin{center}

\caption{Evaluation of mesh reconstruction. Silhouette IoU, rasterized normal RMSE and MS-SSIM are listed. Our method significantly outperforms the compared methods. The ablation study shows the key designs are crucial}
\scalebox{0.95}{
\begin{tabular}{llll}
\hline\noalign{\smallskip}
Method &  IoU &  RMSE &  MS-SSIM\\
\noalign{\smallskip}
\hline
\noalign{\smallskip}
\label{tab:compare}

DeepHuman \cite{zheng2019deephuman} & 0.650 & 0.399 & 0.421 \\
DeepHuman \cite{zheng2019deephuman} variant & 0.779 & 0.309 & 0.587 \\
HMD \cite{zhu2019detailed} & 0.667 & 0.417 & 0.684 \\
HMD \cite{zhu2019detailed} variant & 0.790 & 0.344 & 0.779 \\
Tex2Shape \cite{alldieck2019tex2shape} variant & 0.926 & 0.192 & 0.857 \\
\hline
Ours (no fine-tuning on video) & 0.940 & 0.186&0.857\\
Ours (replace input normal by RGB) & 0.928 &0.190&0.852\\
Ours (no VGG feature) &0.932&0.185&0.865\\
Ours (no deformation propagation) &\textbf{0.941}&0.174&0.869 \\
\hline
Our full method & \textbf{0.941} & \textbf{0.173} & \textbf{0.870} \\
\hline
\end{tabular}}
\end{center}
\end{table}
\setlength{\tabcolsep}{1.4pt}
\begin{figure}
\centering
\begin{minipage}{\linewidth}
\centering
\includegraphics[height=68pt]{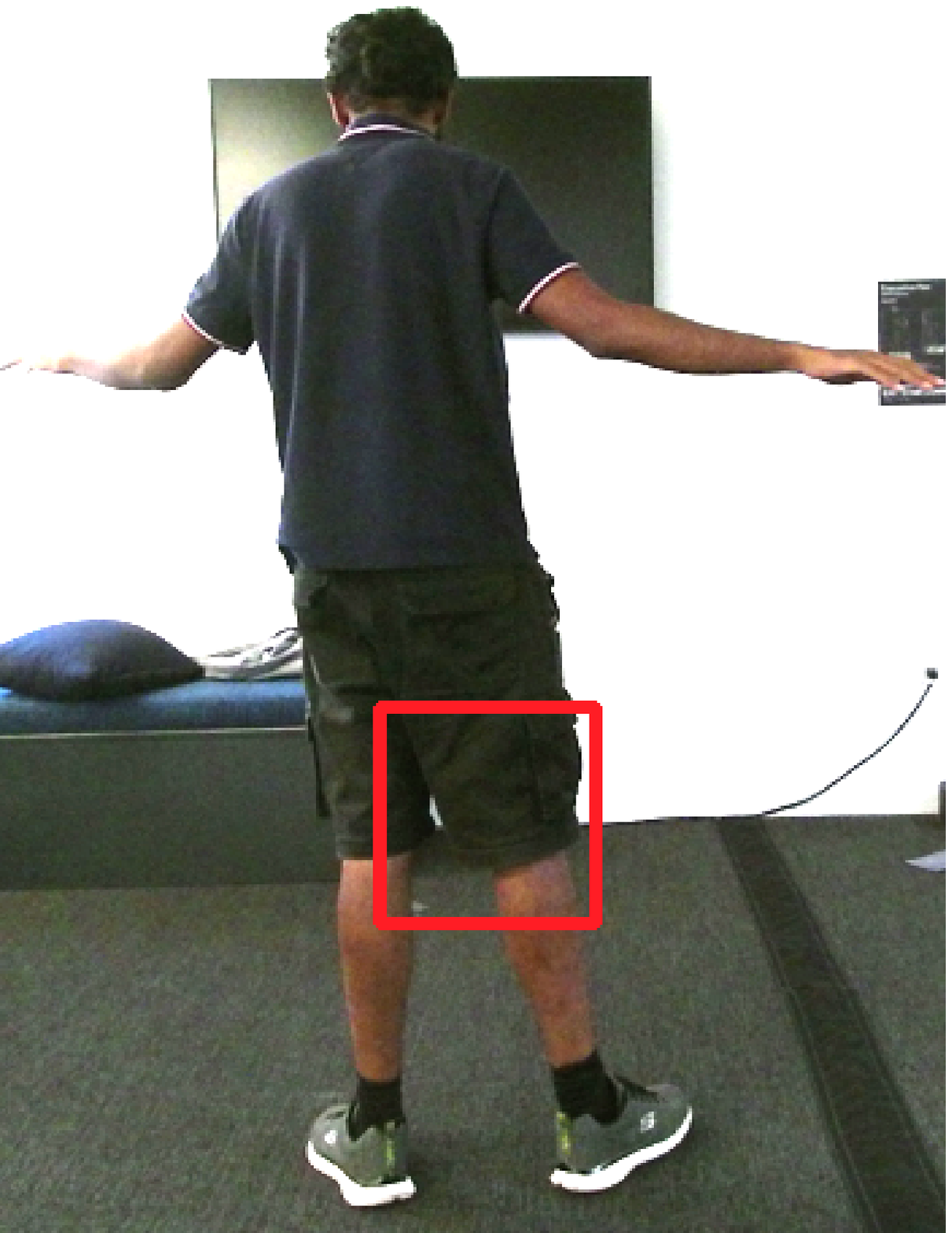}
\includegraphics[height=68pt]{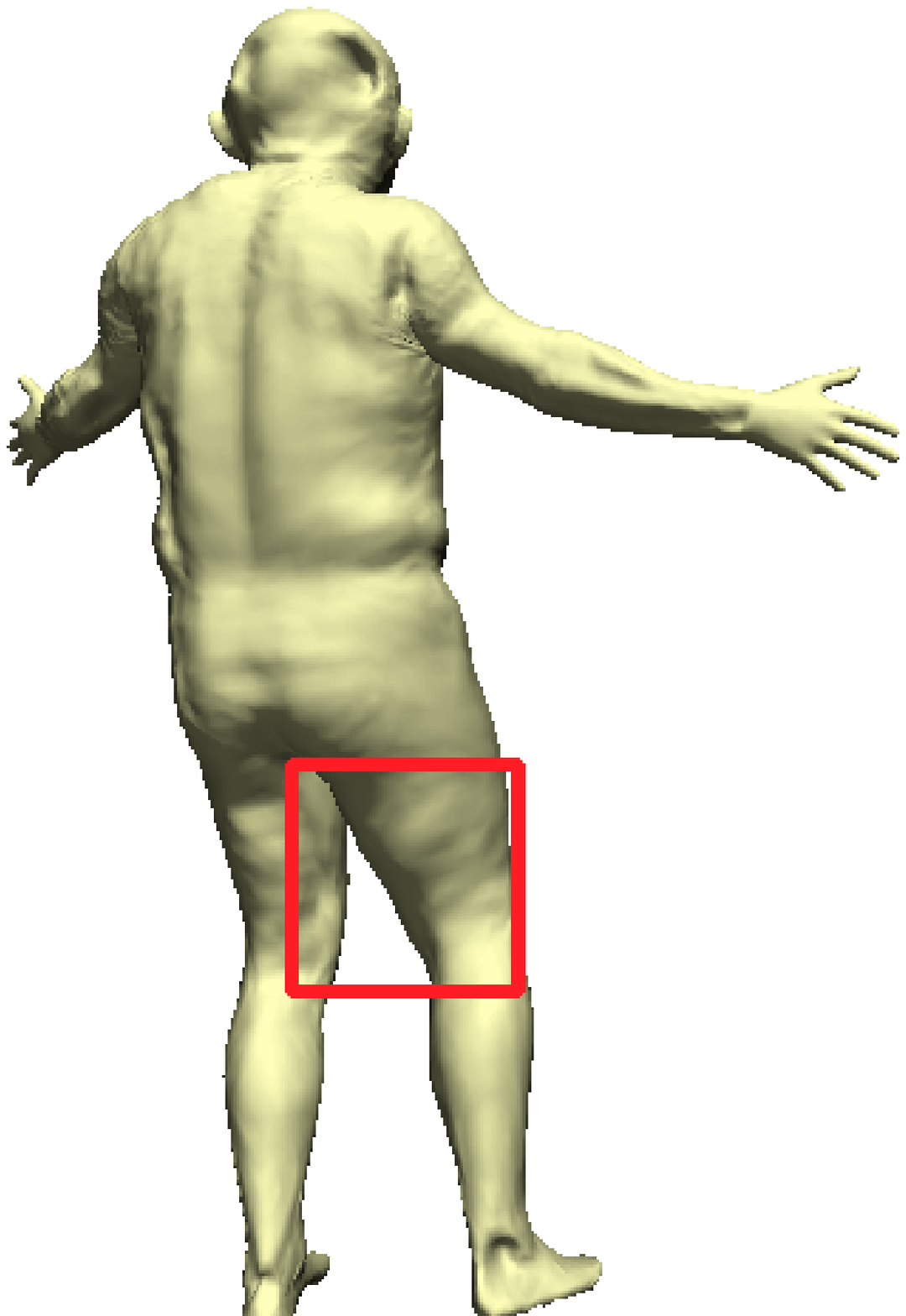}
\includegraphics[height=68pt]{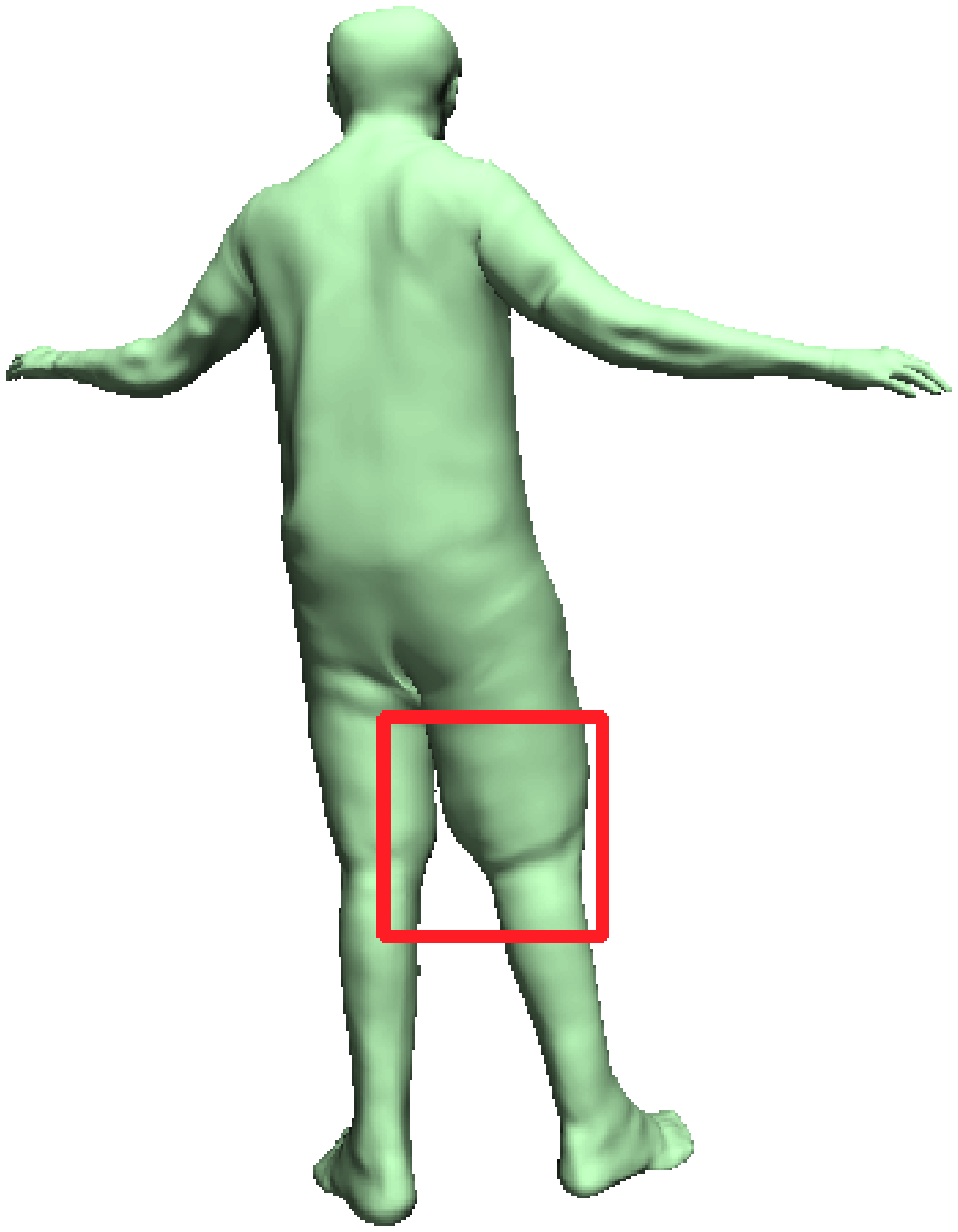}
\includegraphics[height=68pt]{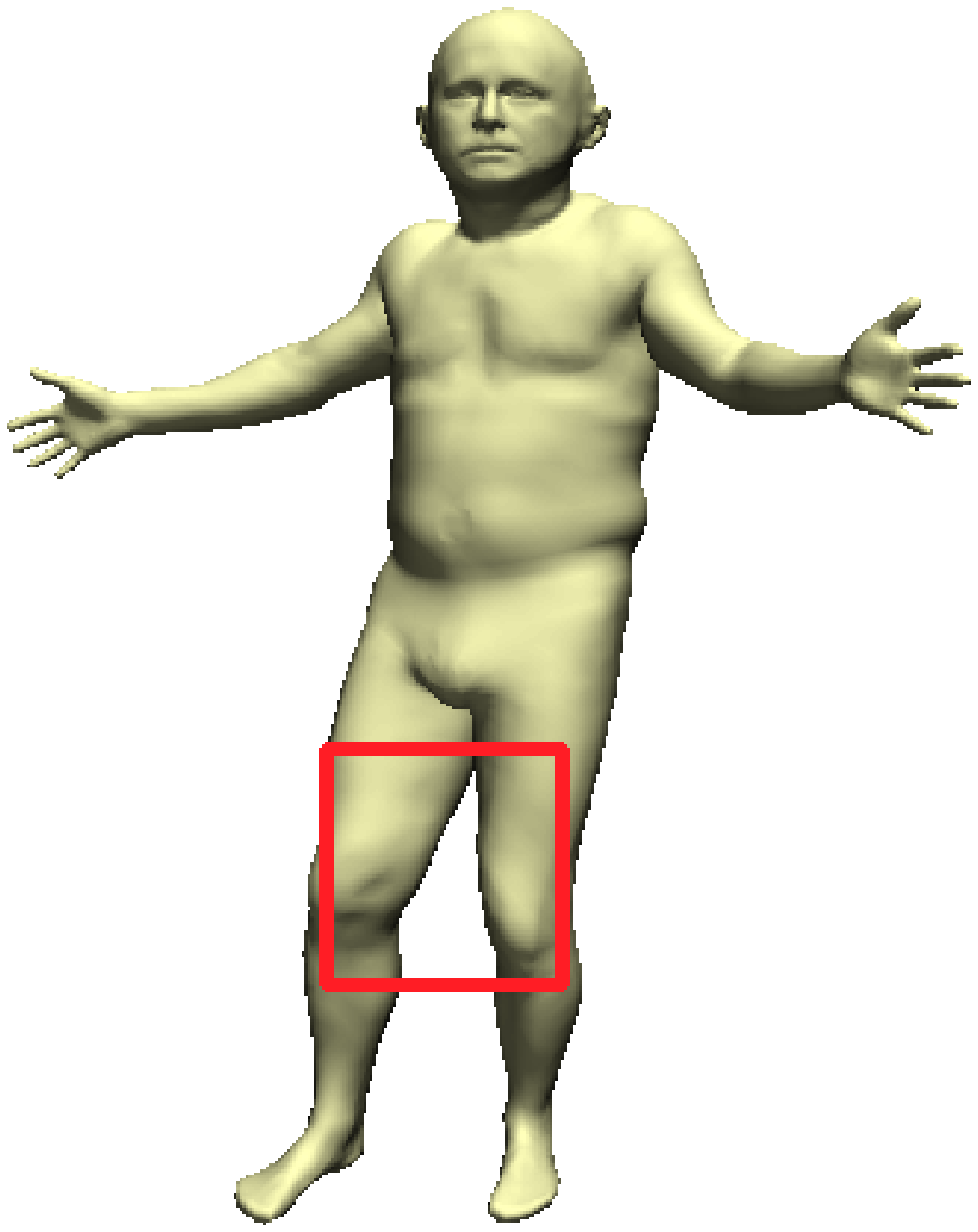}
\includegraphics[height=68pt]{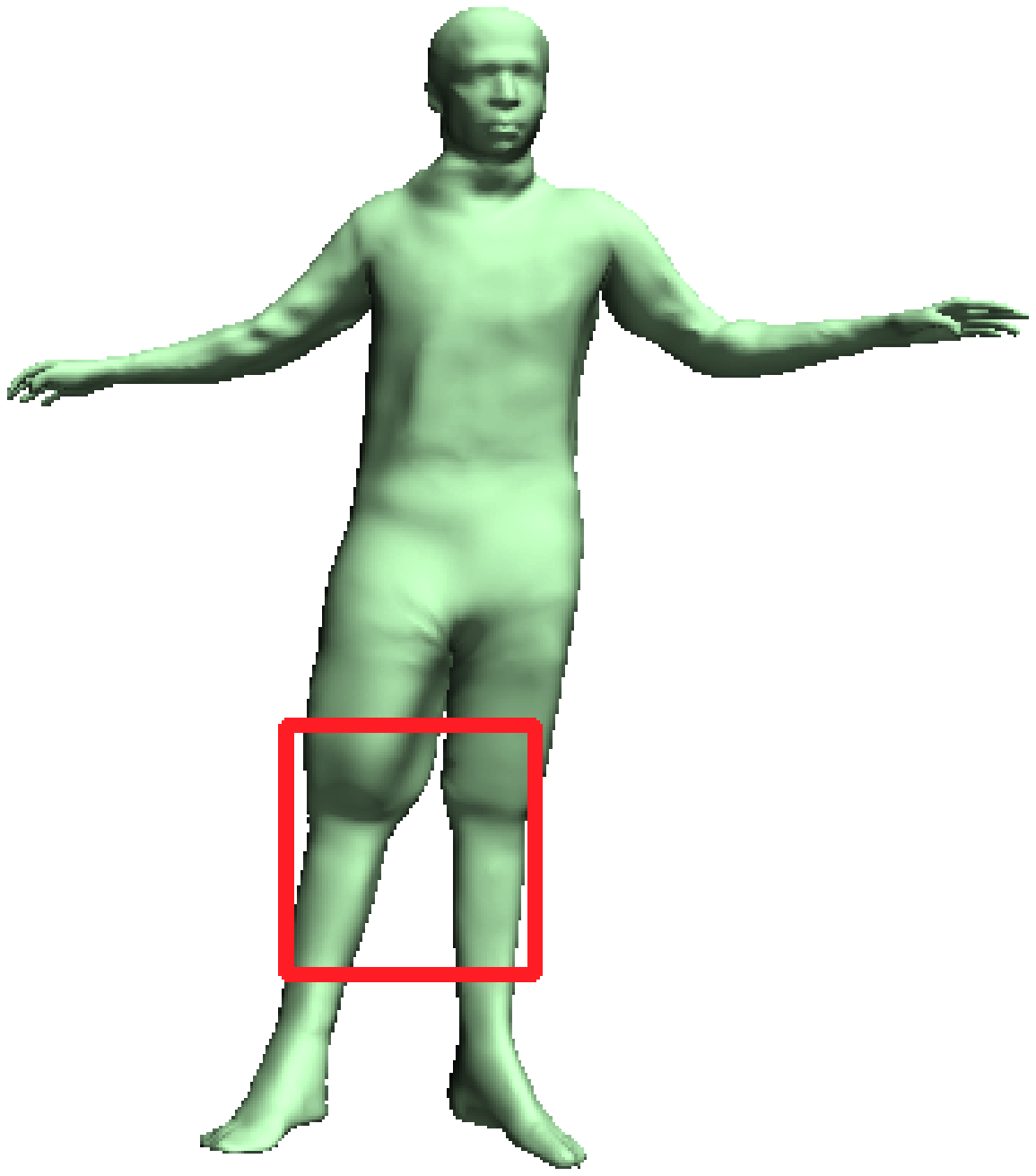}
\end{minipage}
\begin{minipage}{160pt}
\centering
\includegraphics[height=75pt]{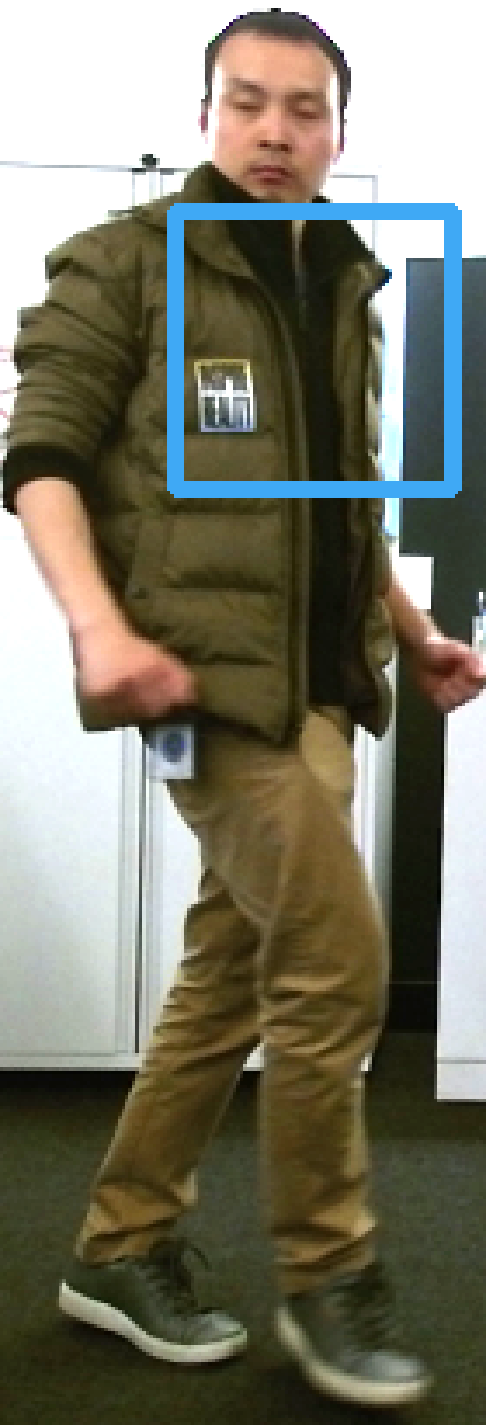}\includegraphics[height=75pt]{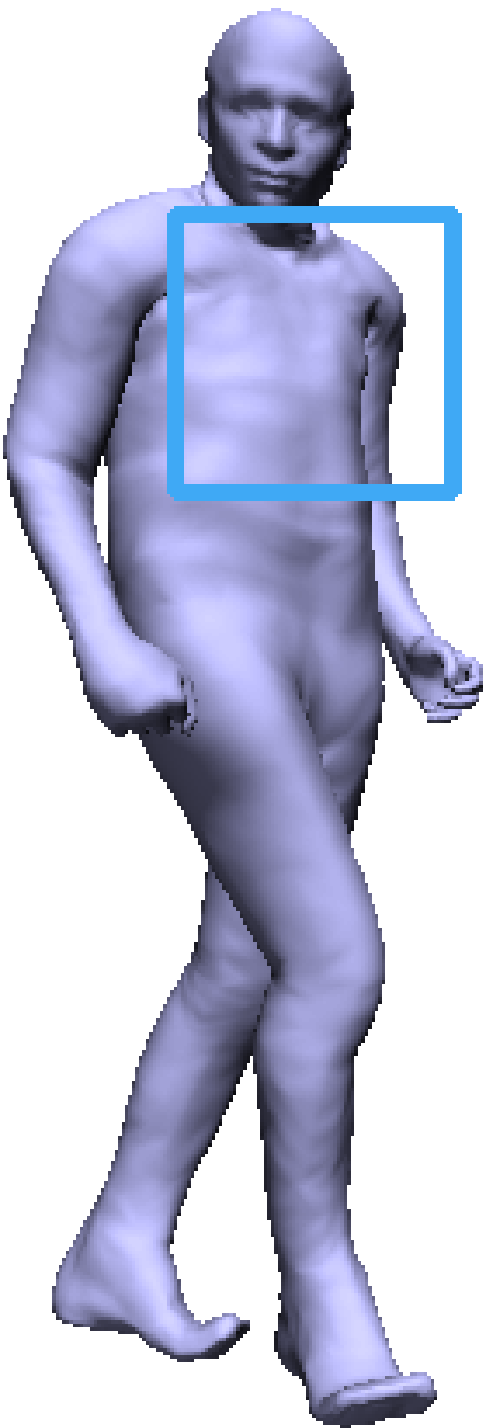}\includegraphics[height=75pt]{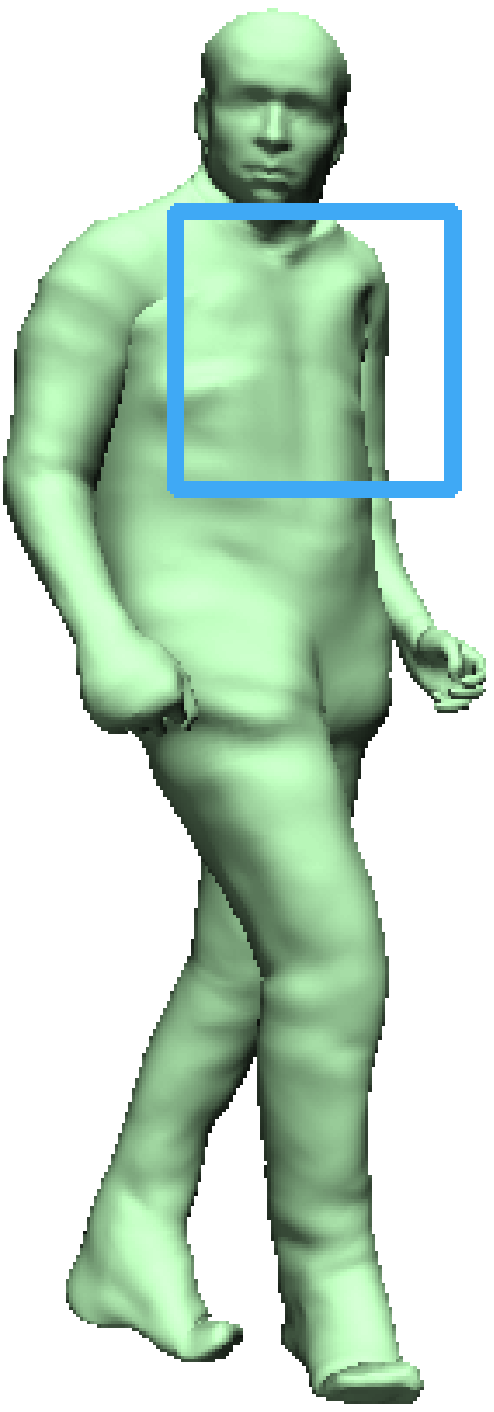}\includegraphics[height=75pt]{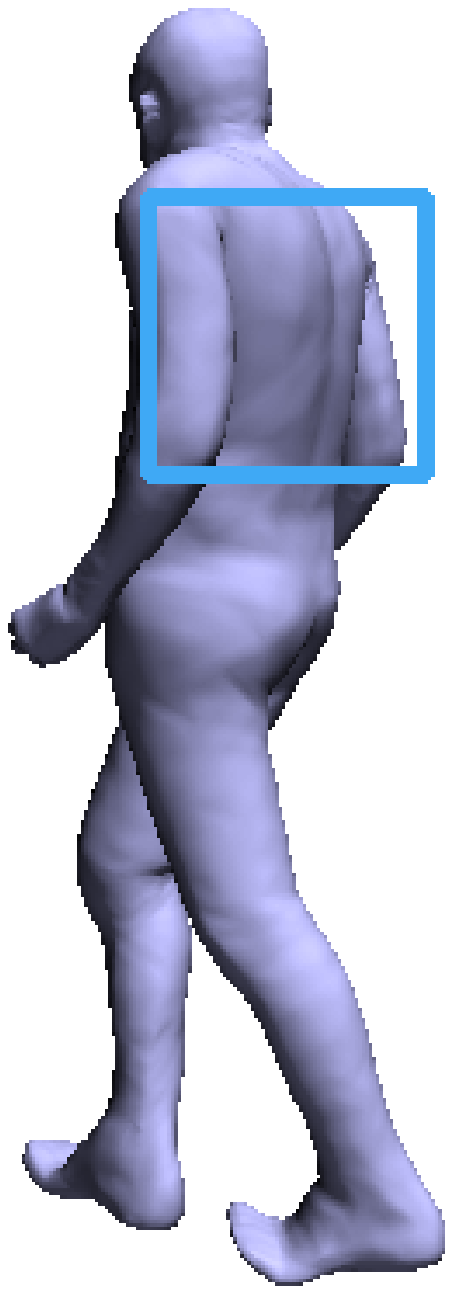}\includegraphics[height=75pt]{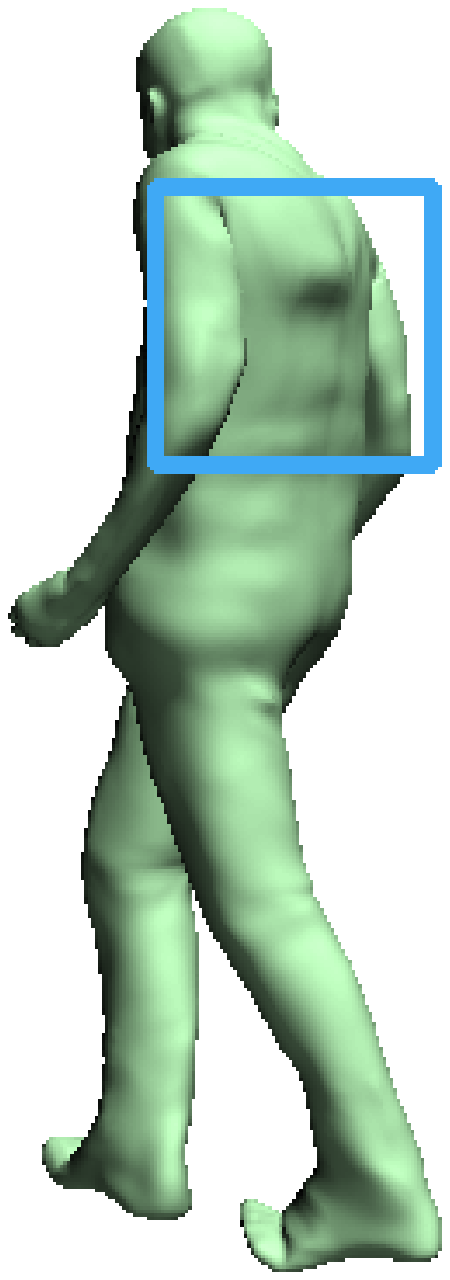}
\end{minipage}
\begin{minipage}{170pt}
\centering
\includegraphics[height=30pt]{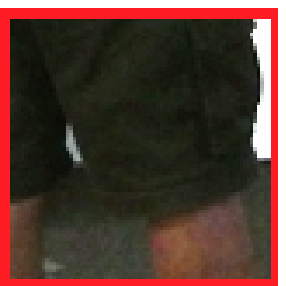}
\includegraphics[height=30pt]{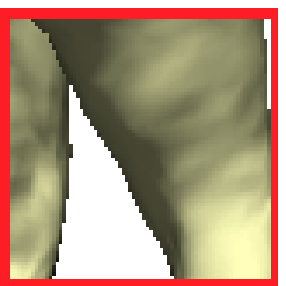}
\includegraphics[height=30pt]{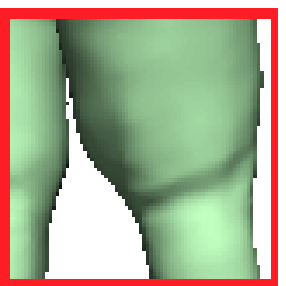}
\includegraphics[height=30pt]{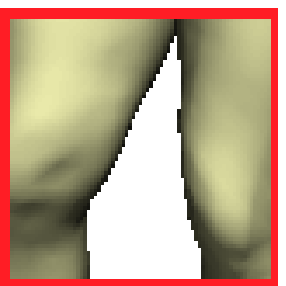}
\includegraphics[height=30pt]{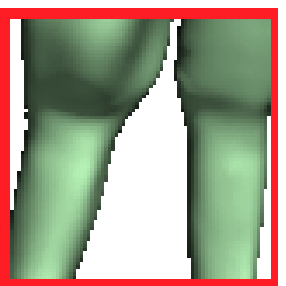}
\includegraphics[height=30pt]{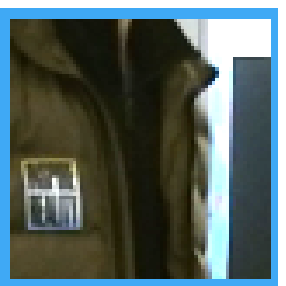}
\includegraphics[height=30pt]{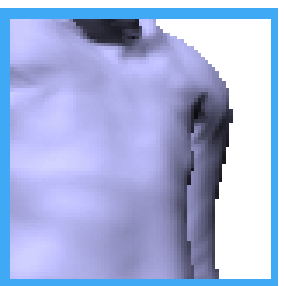}
\includegraphics[height=30pt]{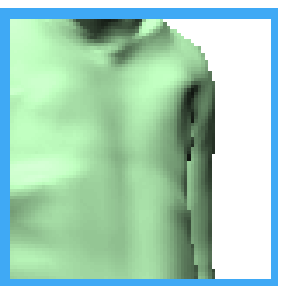}
\includegraphics[height=30pt]{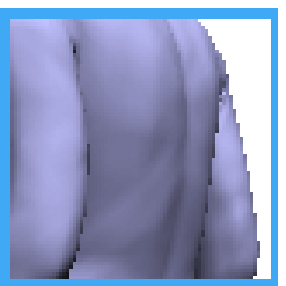}
\includegraphics[height=30pt]{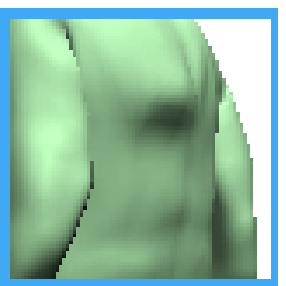}
\end{minipage}
\caption{Viewing from front and back. Tex2Shape (blue), HMD (yellow), Ours (green). Ours captures the shape of occluded jacket hood and shorts deformation.}
\label{fig:back}
\end{figure}

\noindent\textbf{Mesh.}
We compare our method with DeepHuman~\cite{zheng2019deephuman}, HMD~\cite{zhu2019detailed}, and a variant of Tex2Shape \cite{alldieck2019tex2shape} trained on our synthetic images for our Human Model, predicting 3D displacements on top of the posed coarse mesh, using our network and loss settings. For fairness, we compare with both the original version and the variants of DeepHuman and HMD where the initial mesh is replaced by our coarse mesh. See the supplementary material about how to resolve the mismatch of camera parameters and human models.
In Fig.~\ref{fig:compare}, while DeepHuman and HMD provide unrealistic heads and Tex2Shape fails to produce faithful clothing details, our method is shape-preserving and generates better fine geometry. We also recover the geometry of shorts and jacket in occluded regions (Fig.~\ref{fig:back}). The supplemental video shows our result is temporal consistent.
Quantitatively, we evaluate on the synthetic videos by rasterizing 2D normal images. The metrics are silhouette IoU, RMSE, and MS-SSIM\cite{wang2004image} within human mask/bounding box.
Tab.~\ref{tab:compare} shows that our method outperforms the others on all metrics.

\noindent\textbf{Ablation Study.} We quantify the effect of domain finetuning, replacing the normal image by RGB image in MeshRef, removing the VGGNet, and removing the deformation propagation scheme in Tab.~\ref{tab:compare}. Evidently, the first three components are crucial and ignoring them hurts the performance. As expected, removing deformation propagation has little effect because it focuses mainly on the occluded regions (see Fig.~\ref{fig:deform} for its qualitative effect). 

\begin{figure}[t]
\centering
\includegraphics[height=90pt]{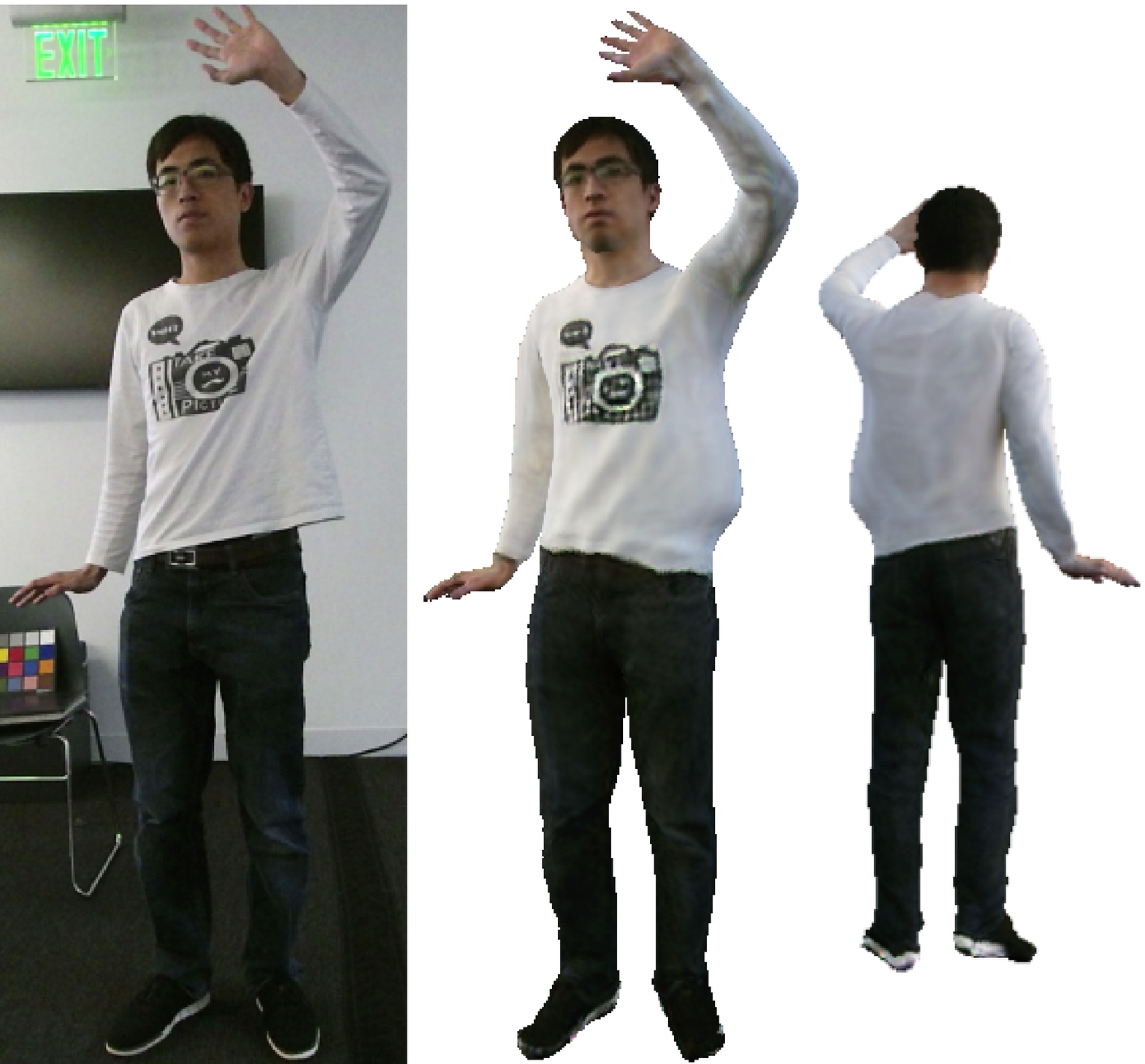}
\includegraphics[height=90pt]{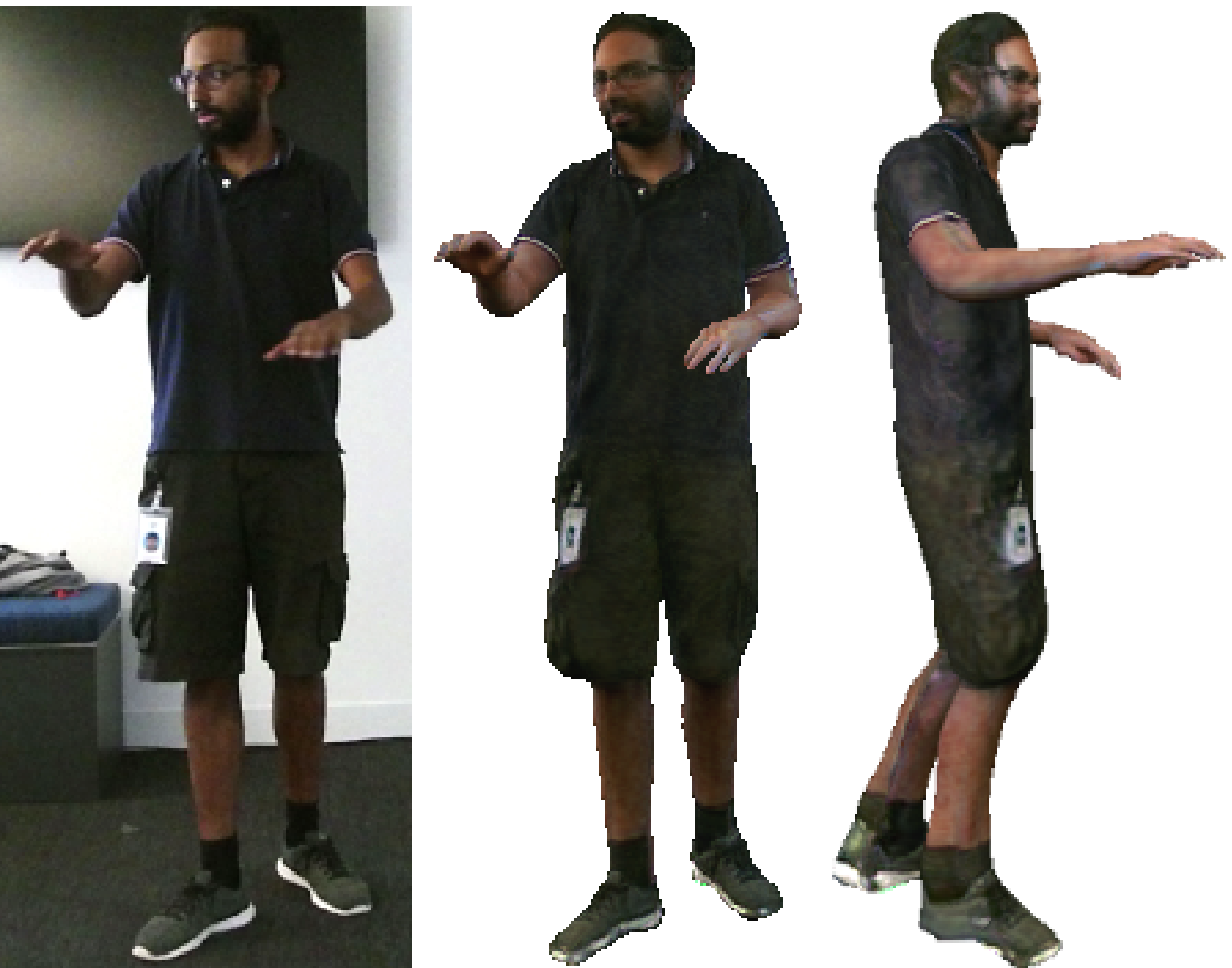}
\includegraphics[height=90pt]{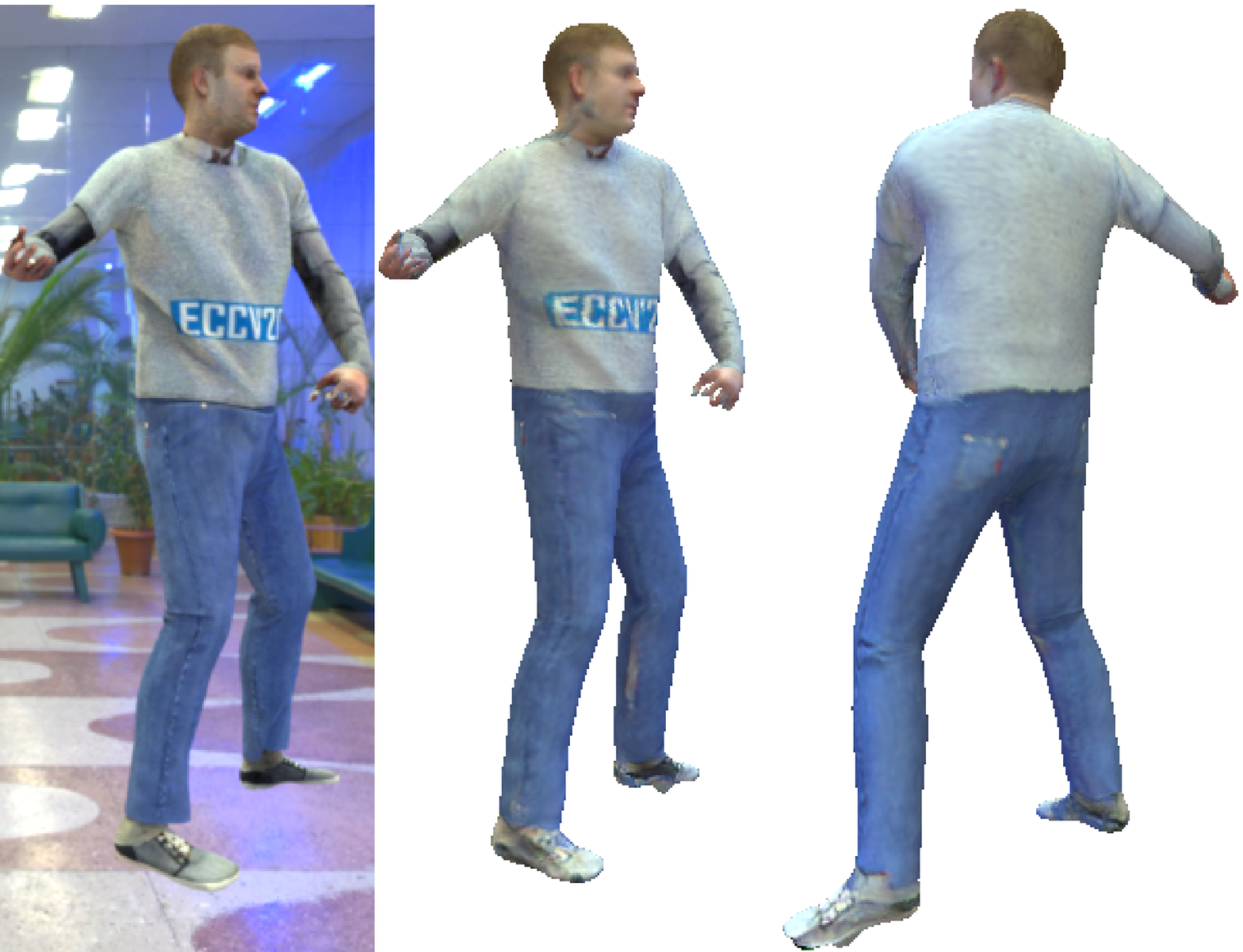}
\caption{Rendering results. The right-most scene is from synthetic video. The rendering has both high fidelity and perceptual quality, from different viewpoints. The clothing wrinkles, logo, and text are clearly recovered.}
\label{fig:render}
\end{figure}
\begin{figure}
\centering
\begin{minipage}{0.06\linewidth}
\subfloat[]{
\includegraphics[height=80pt]{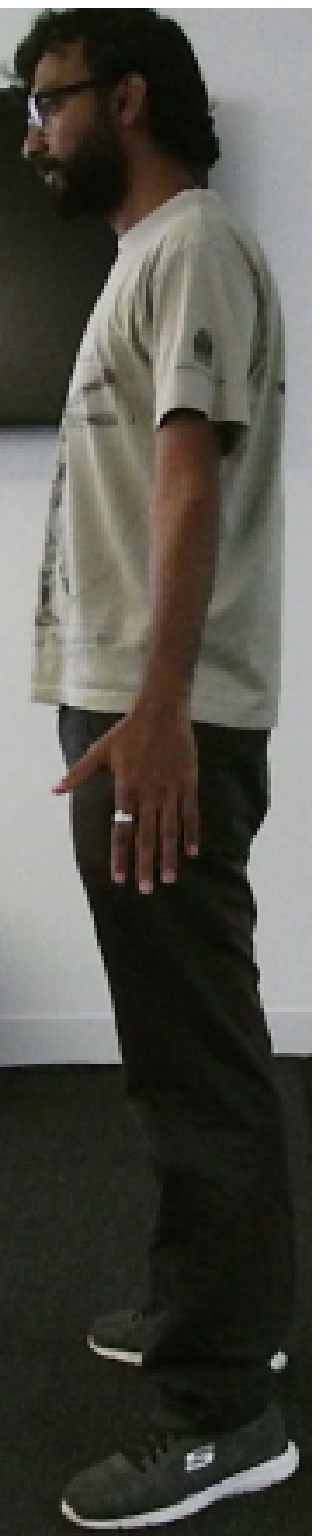}}
\end{minipage}
\begin{minipage}{0.4\linewidth}
\subfloat[Rendering (original lighting)]{
\includegraphics[height=80pt]{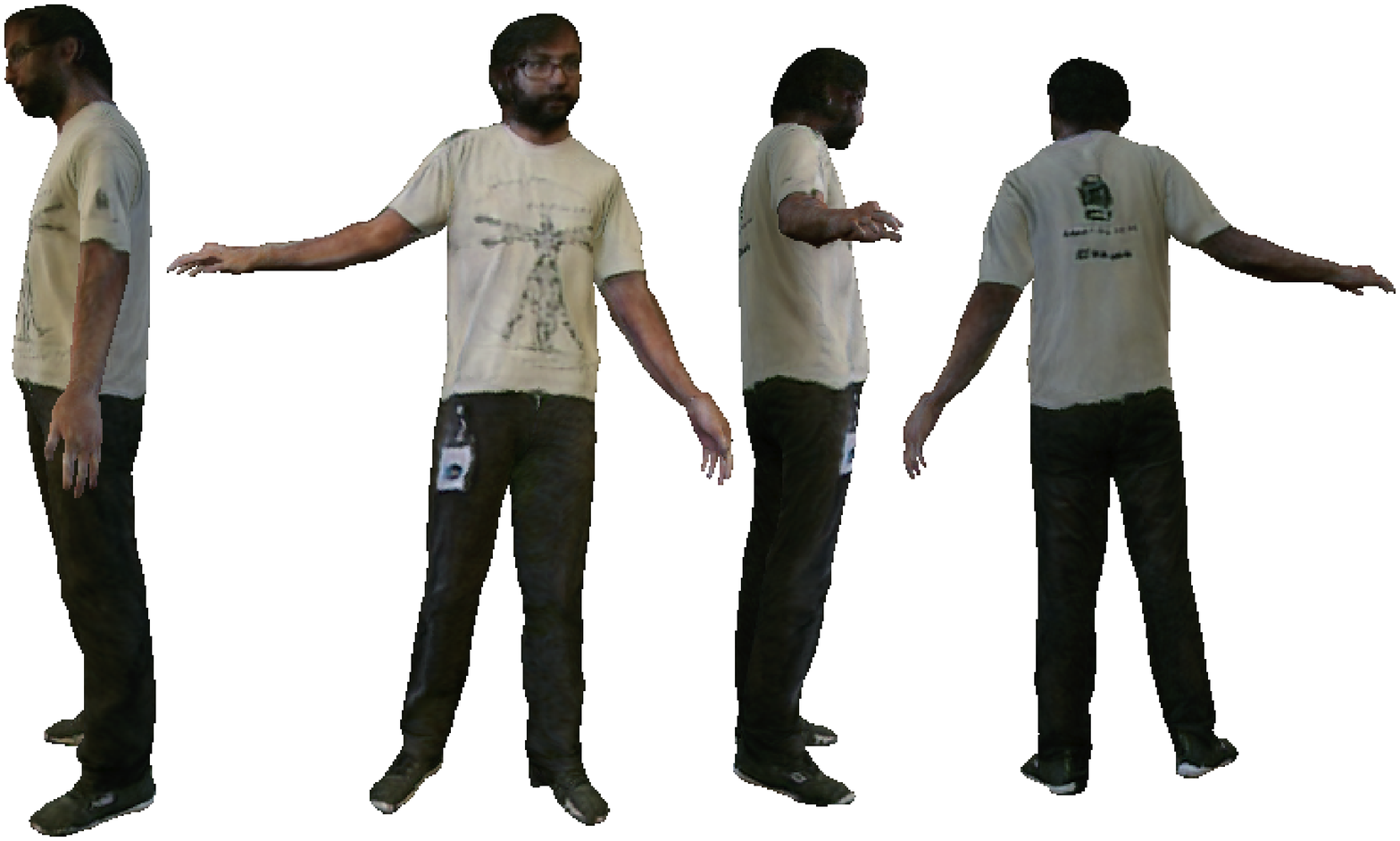}}
\end{minipage}
\begin{minipage}{0.52\linewidth}
\subfloat[Relighting]{
\includegraphics[height=80pt]{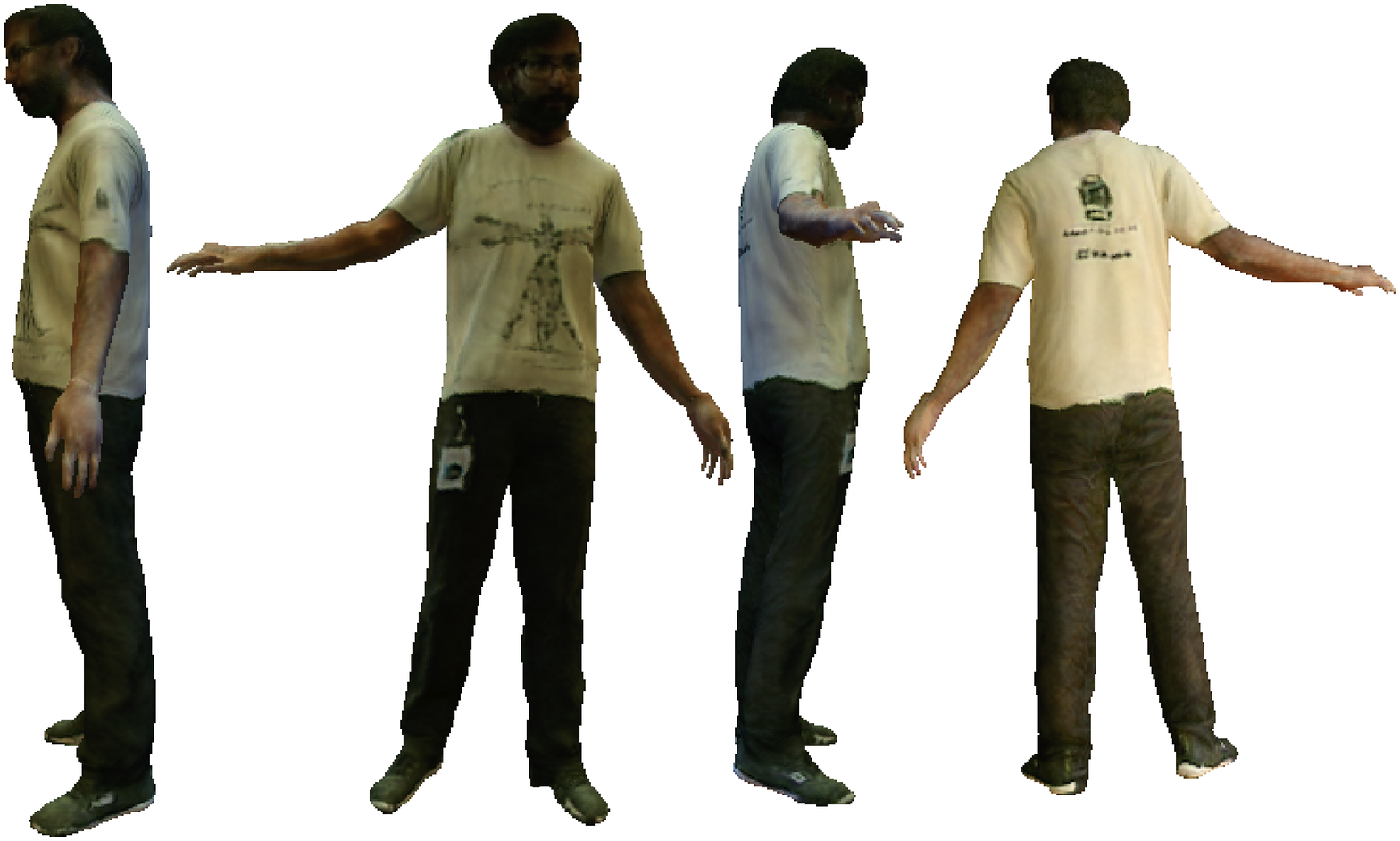}
\includegraphics[width=40pt]{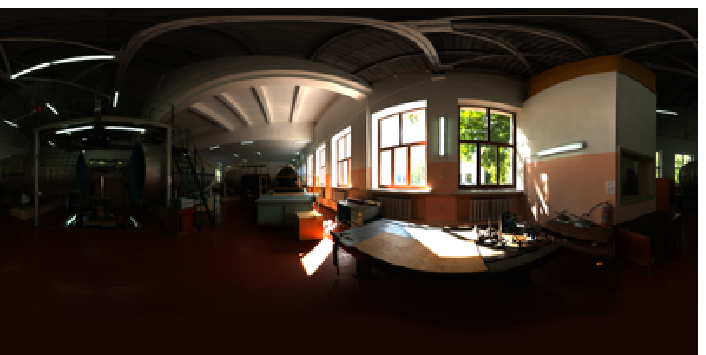}}
\end{minipage}
\caption{Free-viewpoint rendering and relighting. The detailed pattern on the shirt is clearly reconstructed. The shading varies in the clothing wrinkles implying that the wrinkles are correctly estimated as geometry rather than texture.}
\label{fig:relight}
\end{figure}
\noindent\textbf{Applications.} We show rendering from novel viewpoints (Fig.~\ref{fig:render}) and relighting in a different environment (Fig.~\ref{fig:relight}) using our outputs. The results have clear textures and realistic shading variation around clothing wrinkles.
\section{Conclusion}
In summary, we present TexMesh, a state of art method for high fidelity human texture and geometry reconstruction, enabling high quality free-viewpoint human rendering on real videos. Our key idea is a self-supervised learning framework to adapt a synthetically-trained model to real videos. After adaptation, our model runs at interactive frame rate. In the future, we will train the model on many real videos concurrently to learn a generic shape prior as it could allow faster adaptation to new sequences or even require no adaptation at all.

Besides, we will address two limitations of our method in the future. First, we rely on a spherical camera to capture the lighting and represent it using only low frequency Spherical Harmonics. Exploring recent techniques on using a single narrow FOV image to estimate high frequency environmental lighting~\cite{sengupta2018sfsnet,gardner2017learning} together with subtle lighting cues from human appearance~\cite{yi2018faces} is an exciting future direction to tackle this problem. Second, we assume the deformation in invisible regions is the same or similar as in the keyframes, which is not always true. We will consider using clothing simulation~\cite{yu2019simulcap} to alleviate this problem.
\\

\boldstart{Acknowledgements} We thank Junbang Liang, Yinghao Huang, and Nikolaos Sarafianos for their help with the synthetic training data generation.
\clearpage
\begin{center}
{\Large \bf Supplementary Material}
\end{center}
\setcounter{section}{0}
\section{UV Mapping}
\begin{figure}
\centering
\includegraphics[width=\linewidth]{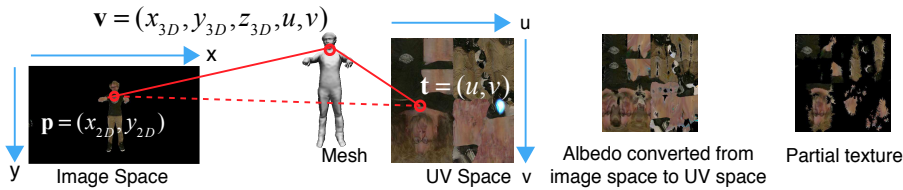}
\caption{The mesh bridges image space and UV space. Assume point $\mathbf{v}$ on the mesh is associated with its 3D position $(x_{3D},y_{3D},z_{3D})$ and UV coordinates $(u, v)$. The corresponding point $\mathbf{p}=(x_{2D},y_{2D})$ in image space can be obtained via camera projection. It also corresponds to point $\mathbf{t}=(u, v)$ in UV space. To convert features from image space to UV space, for each $\mathbf{t}$, we first obtain its 3D position $(x_{3D},y_{3D},z_{3D})$ by barycentric interpolation, project it to image coordinates $(x_{2D},y_{2D})$, and finally sample features from the image. }
\label{fig:convert}
\end{figure}
We follow the common practice in graphics where texture is stored using a UV mapping~\cite{blinn1976texture}. This map unwraps a mesh into 2D space, called UV space. Each pixel $\mathbf{t}=(u,v)$ in UV space corresponds to a point $\mathbf{v}$ on the mesh. Its 3D position is defined by the barycentric interpolation of the vertices of the face where the point is on. With the 3D position, we can project it to the image space of a calibrated camera. Thus, we can sample image features and convert them into UV space. Fig.~\ref{fig:convert} shows an example of converting albedo to UV space. It is further converted into a partial texture by masking with visibility.
\section{Visibility Map}
\begin{figure}
\centering
\includegraphics[width=\linewidth]{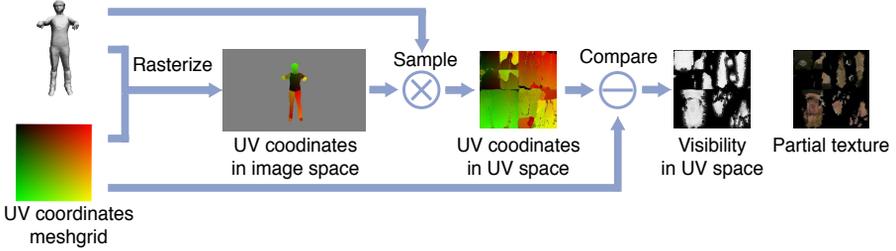}
\caption{Visibility and partial texture generation. By rasterizing UV coordinates to image space and sample it back to UV space, we obtain a UV coordinates map where only the visible parts are "correct". By comparing it with the ground truth UV meshgrid, we obtain the visibility map in UV space. We further calculate partial texture by masking the sampled albedo.}
\label{fig:visibility}
\end{figure}
To calculate visibility, as shown in Fig. \ref{fig:visibility}, we rasterize a image with UV coordinates, then sample that to UV space, and compare with the correct UV coordinates. The pixels whose sampled UVs are consistent with its position in UV space are visible. By masking the sampled albedo with visibility, we obtain the partial texture.
\section{Augment MeshRef Features}
In addition to VGG \cite{simonyan2014very} features, we can further augment the features by including vertex position information. Specifically, we can rasterize coarse vertex position into both image spcae and UV space, to become a vertex position image $I_{vp}$ and a vertex position map $T_{vp}$.
We append $I_{vp}$ to the input of VGGNet, and append $T_{vp}$ to the input of the UV space CNN.
\section{Implementation Details}
\subsubsection{Network Architecture.}

All the CNNs are U-Net sharing similar architectures, except for the VGG16 Network \cite{simonyan2014very} in MeshRef module. See Tab. \ref{tab:components} for the shared components and Tab. \ref{tab:albenorm}, \ref{tab:texgen}, and \ref{tab:meshref} for architectures of AlbeNorm, TexGen, and MeshRef CNNs. Specially, the CNN in TexGen predicts the residual between the coarse texture and the fine texture.
\setlength{\tabcolsep}{4pt}
\begin{table}
\begin{center}
\caption{Network Components. We use ReLU~\cite{nair2010rectified} for activation, and Instance Normalization~\cite{ulyanov2016instance} for normalization}
\begin{tabular}{lllll}
\hline\noalign{\smallskip}
Type & Components\\
\noalign{\smallskip}
\hline
\noalign{\smallskip}
\label{tab:components}
inconv & [Conv3$\times$3 + ReLU + InstanceNorm]$\times$2\\
down & [Conv3$\times$3 + ReLU + InstanceNorm]$\times$2 + MaxPool2$\times$2\\
up & Upsample + [Conv3$\times$3 + ReLU + InstanceNorm]$\times$2 \\
outconv & Conv1$\times$1 \\
\end{tabular}
\end{center}
\end{table}
\setlength{\tabcolsep}{1.4pt}
\setlength{\tabcolsep}{4pt}
\begin{table}
\begin{center}
\caption{Network Architecture of AlbeNorm CNN}
\begin{tabular}{lllll}
\hline\noalign{\smallskip}
Name & Type & Input & Output Channels\\
\noalign{\smallskip}
\hline
\noalign{\smallskip}
\label{tab:albenorm}
inc & inconv & RGB+SH lighting & 64\\
down1 & down & inc & 128\\
down2 & down & down1 & 256\\
down3 & down & down2 & 512\\
down4 & down & down3 & 512\\
up1a & up & down4, down3 & 256\\
up2a & up & up1a, down2 & 128\\
up3a & up & up2a, down1 & 64\\
up4a & up & up3a, inc & 64\\
outca (Normal Output) & up & up4a & 3 \\
up1b & up & down4, down3 & 256 \\
up2b & up & up1b, down2& 128\\
up3b & up & up2b, down1& 64\\
up4b & up & up3b, inc& 64\\
outcb (Albedo Output) & outconv & up4b & 3\\
\end{tabular}
\end{center}
\end{table}
\setlength{\tabcolsep}{1.4pt}
\setlength{\tabcolsep}{4pt}
\begin{table}
\begin{center}
\caption{Network Architecture of TexGen CNN}
\begin{tabular}{lllll}
\hline\noalign{\smallskip}
Name & Type & Input & Output Channels\\
\noalign{\smallskip}
\hline
\noalign{\smallskip}
\label{tab:texgen}
inc & inconv & Coarse Texture & 64\\
down1 & down & inc & 128\\
down2 & down & down1 & 256\\
down3 & down & down2 & 512\\
down4 & down & down3 & 512\\
up1 & up & down4, down3 & 256\\
up2 & up & up1, down2 & 128\\
up3 & up & up2, down1 & 64\\
up4 & up & up3, inc & 64\\
outc  & outconv & up4 & 3 \\
\end{tabular}
\end{center}
\end{table}
\setlength{\tabcolsep}{1.4pt}
\setlength{\tabcolsep}{4pt}
\begin{table}
\begin{center}
\caption{Network Architecture of MeshRef CNN. ``feat0'', ``feat1'', ``feat2'', ``feat3'', ``feat4'' are features converted from VGGNet input, conv1\_2, conv2\_2, conv3\_3, and conv4\_3 features}
\begin{tabular}{lllll}
\hline\noalign{\smallskip}
Name & Type & Input & Output Channels\\
\noalign{\smallskip}
\hline
\noalign{\smallskip}
\label{tab:meshref}
inc & inconv & feat0 & 64\\
down1 & down & inc, feat1 & 128\\
down2 & down & down1, feat2 & 256\\
down3 & down & down2, feat3 & 512\\
down4 & down & down3, feat4 & 512\\
up1 & up & down4, down3 & 256\\
up2 & up & up1, down2 & 128\\
up3 & up & up2, down1 & 64\\
up4 & up & up3, inc & 64\\
outc  & outconv & up4 & 3 \\
\end{tabular}
\end{center}
\end{table}
\setlength{\tabcolsep}{1.4pt}
\subsubsection{Hyperparameters.}
We use $K=30$ for the number of selected frames, and $\lambda^{an}_{a}=1, \lambda^{an}_{n}=1, \lambda^{tg}_{L1}=20, \lambda^{tg}_{pct}=1, \lambda^{tg}_{lap}=10, \lambda^{mr1}_{L1}=1, \lambda^{mr1}_{ssim}=1, \lambda^{mr1}_{lap}=20,  \lambda^{mr2}_{pct}=1, \lambda^{mr2}_{sil}=100, \lambda^{mr2}_{temp}=10, \lambda^{mr2}_{pos}=10, \lambda^{mr2}_{lap}=10, \lambda^{mr2}_{deform}=10$ for the loss weights. We use learning rate $10^{-5}$ for pretraining AlbeNorm, $10^{-4}$ for pretraining MeshRef, $3\times10^{-4}$ for optimizing TexGen, and $5\times10^{-5}$ for finetuning MeshRef. We use batch size $4$ for pretraining AlbeNorm, $1$ for pretraining MeshRef, $1$ for optimizing TexGen, and $3$ for finetuning MeshRef (as a triplet for motion smoothness loss). VGGNet is trained from scratch with MeshRef CNN, and kept fixed during finetuning. To speed up finetuning, we use a smaller image size $480\times270$ for photometric losses, but the image features are from the $960\times540$ original image.

\subsubsection{Adaptive Robust Perceptual Loss.}
We use the adaptive robust loss~\cite{barron2019general} for perceptual losses~\cite{johnson2016perceptual}. We use VGG16 conv1\_2, conv2\_2, conv3\_3, and conv4\_3 features for TexGen, and conv3\_3, and conv4\_3 features for MeshRef. We use learning rate $3\times 10^{-4}$ for the adaptive robust function.
\section{Qualitative Texture Generation Results}
\begin{figure}
\centering
\subfloat[(a) Albedo]{
\includegraphics[width=0.15\linewidth]{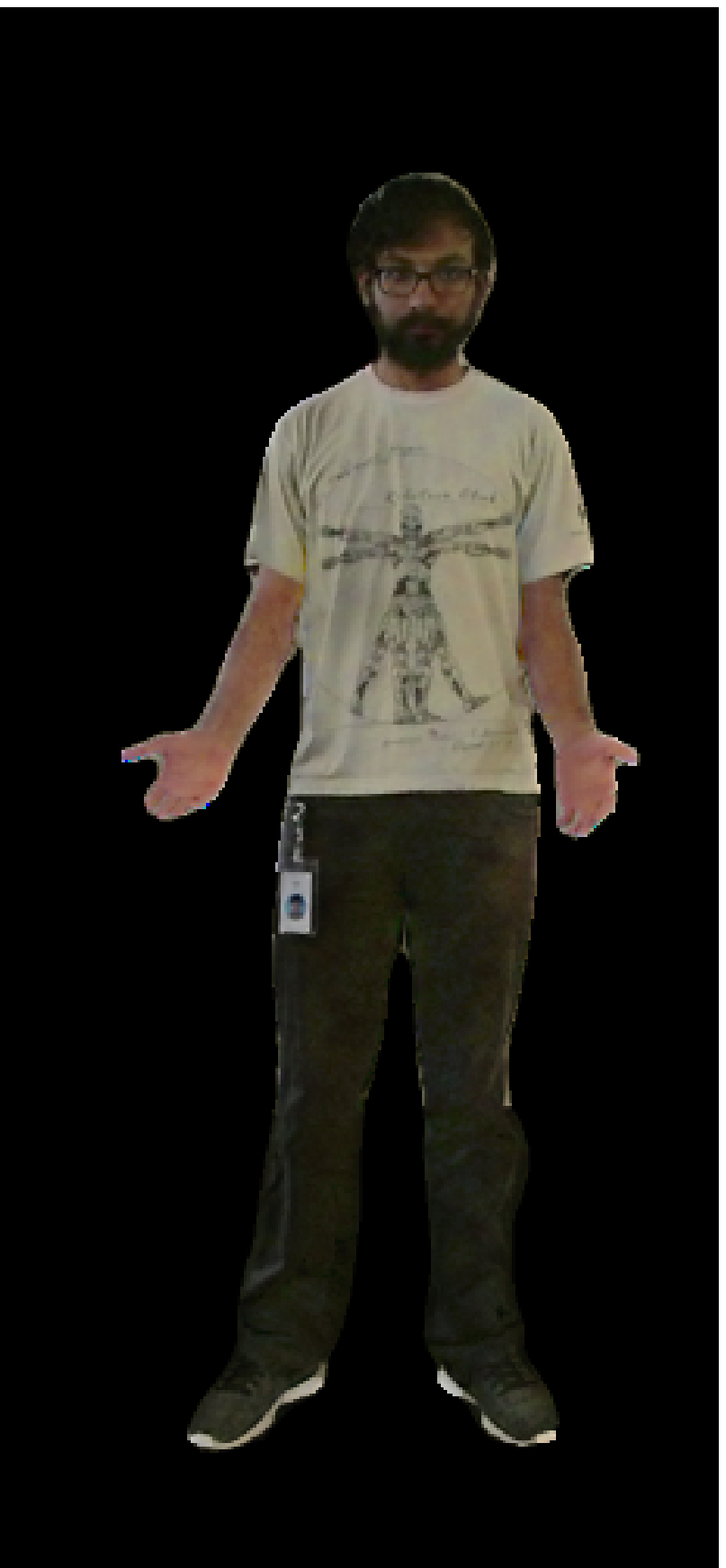}}
\subfloat[(b) SBM~\cite{alldieck2018video}]{
\includegraphics[width=0.15\linewidth]{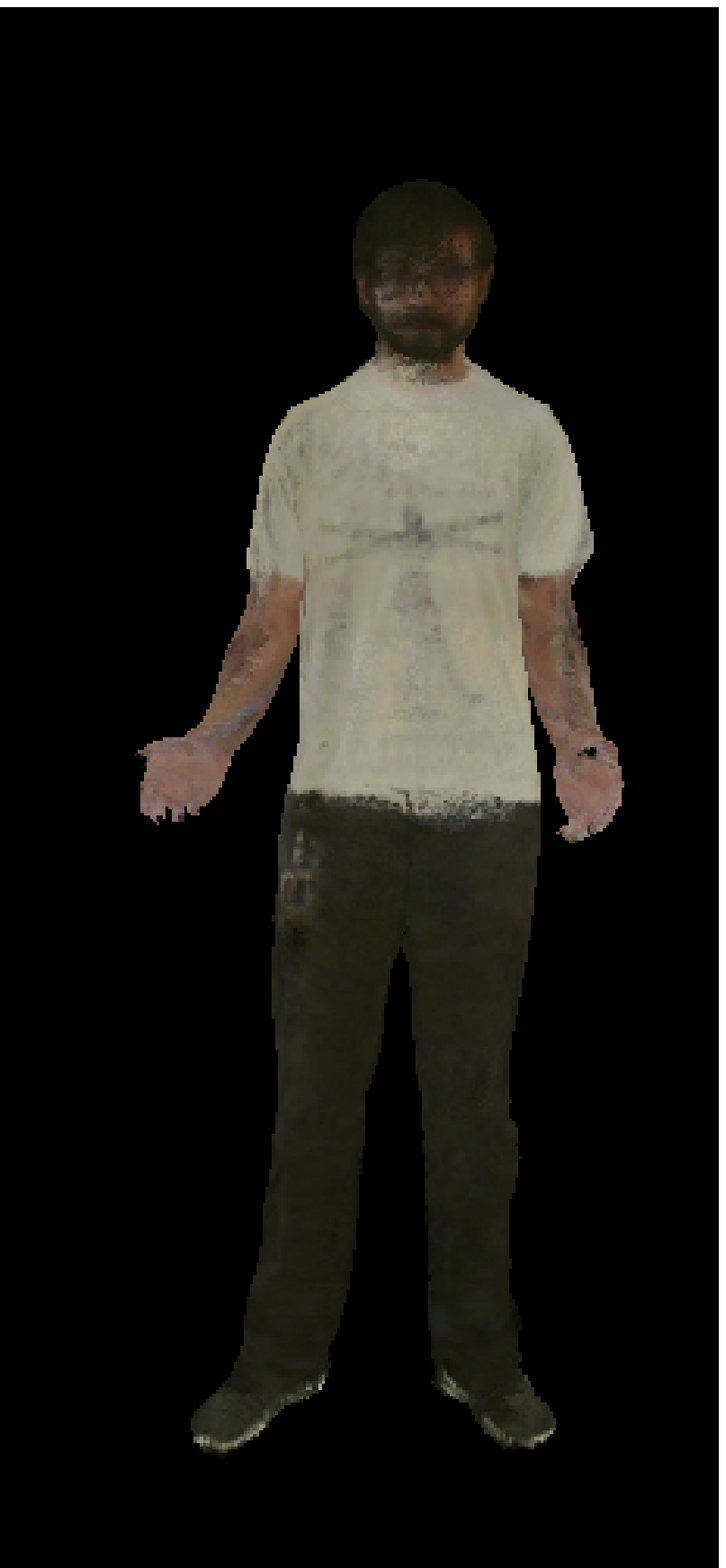}}
\subfloat[(c) TNA~\cite{shysheya2019textured}]{
\includegraphics[width=0.15\linewidth]{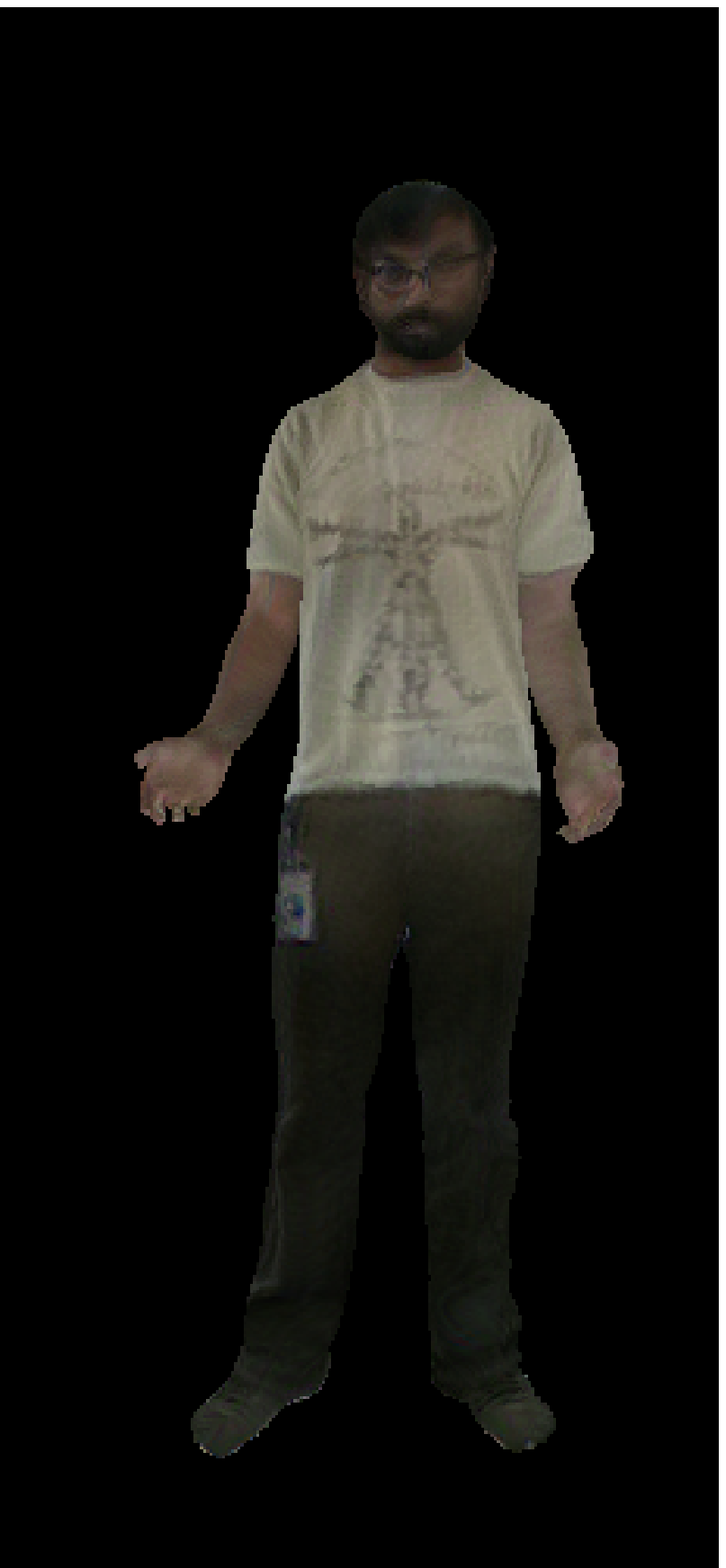}}
\subfloat[(d) Ours]{
\includegraphics[width=0.15\linewidth]{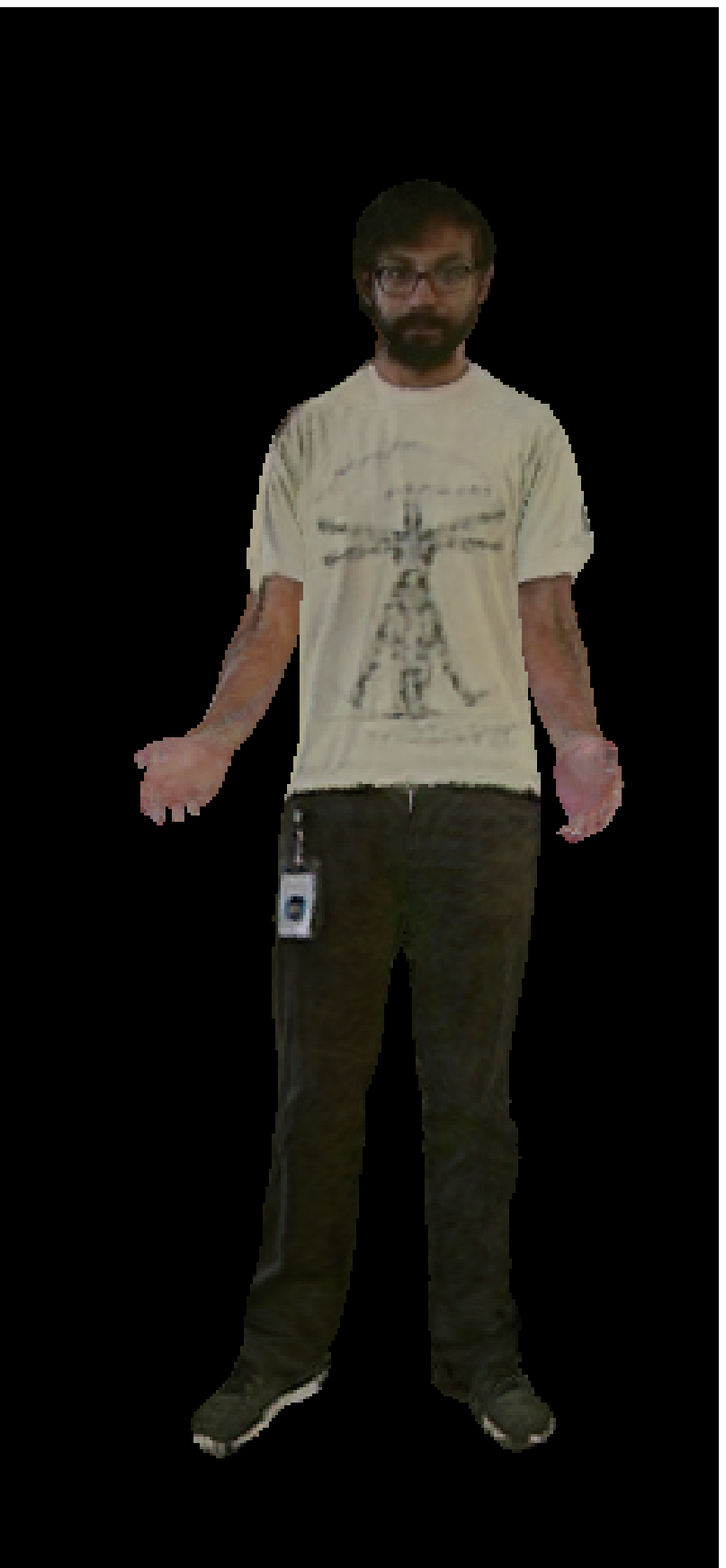}}

\caption{Comparing texture generation by visualizing no-shading image on real data. (a) is albedo image from AlbeNorm, which can be seen as ``ground truth''. Our result has clearer texture, and less baked-in shading.}
\label{fig:texture1}
\end{figure}
\begin{figure}
\centering
\subfloat[(a) Albedo]{
\includegraphics[width=0.15\linewidth]{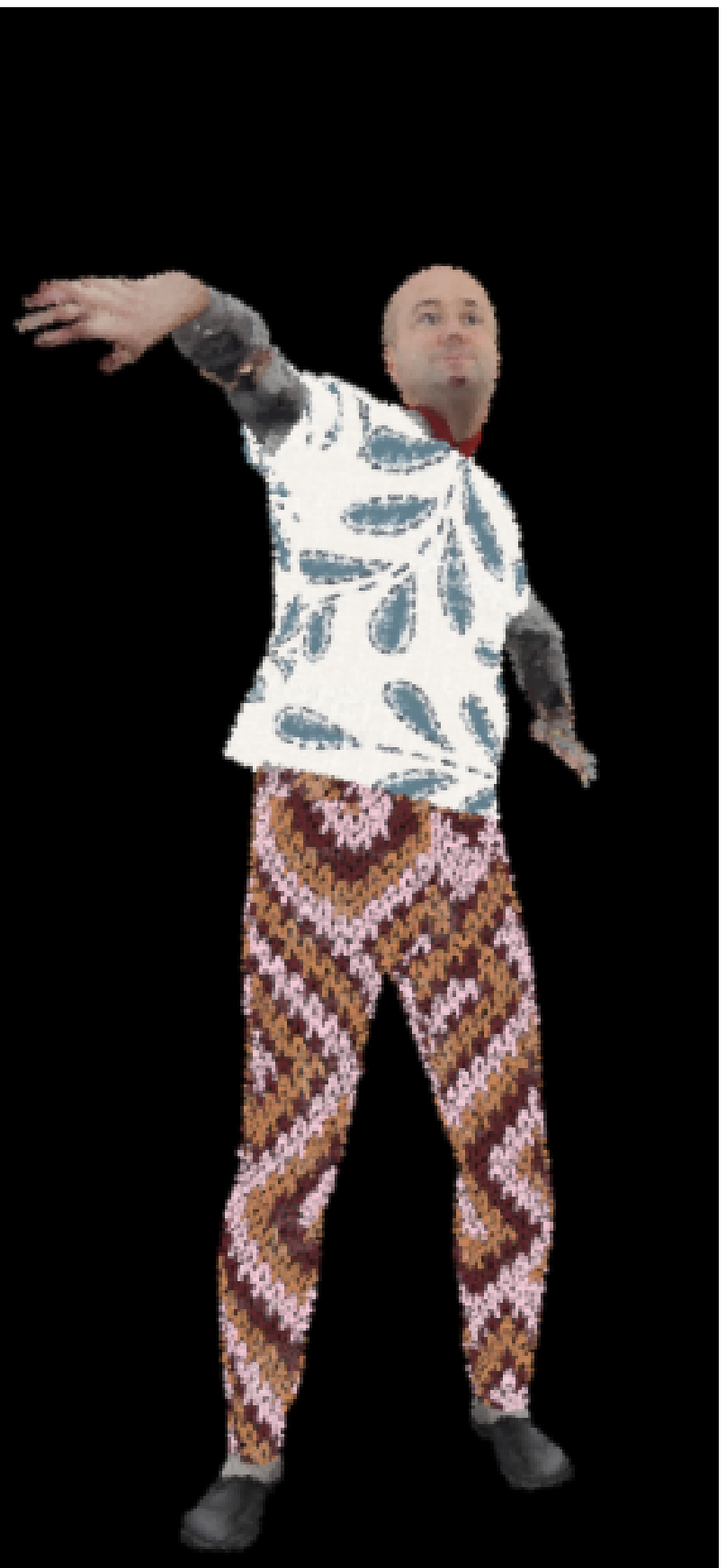}}
\subfloat[(b) SBM~\cite{alldieck2018video}]{
\includegraphics[width=0.15\linewidth]{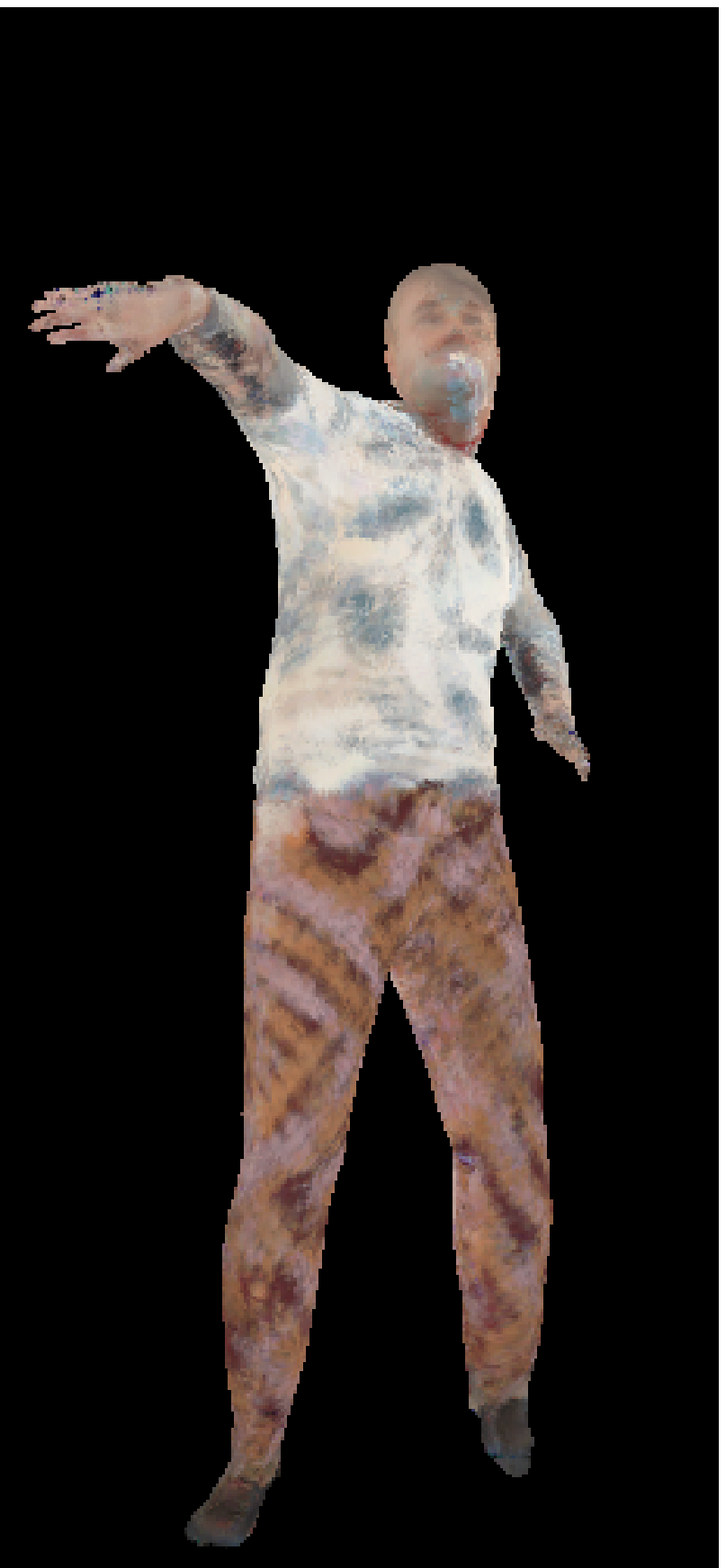}}
\subfloat[(c) TNA~\cite{shysheya2019textured}]{
\includegraphics[width=0.15\linewidth]{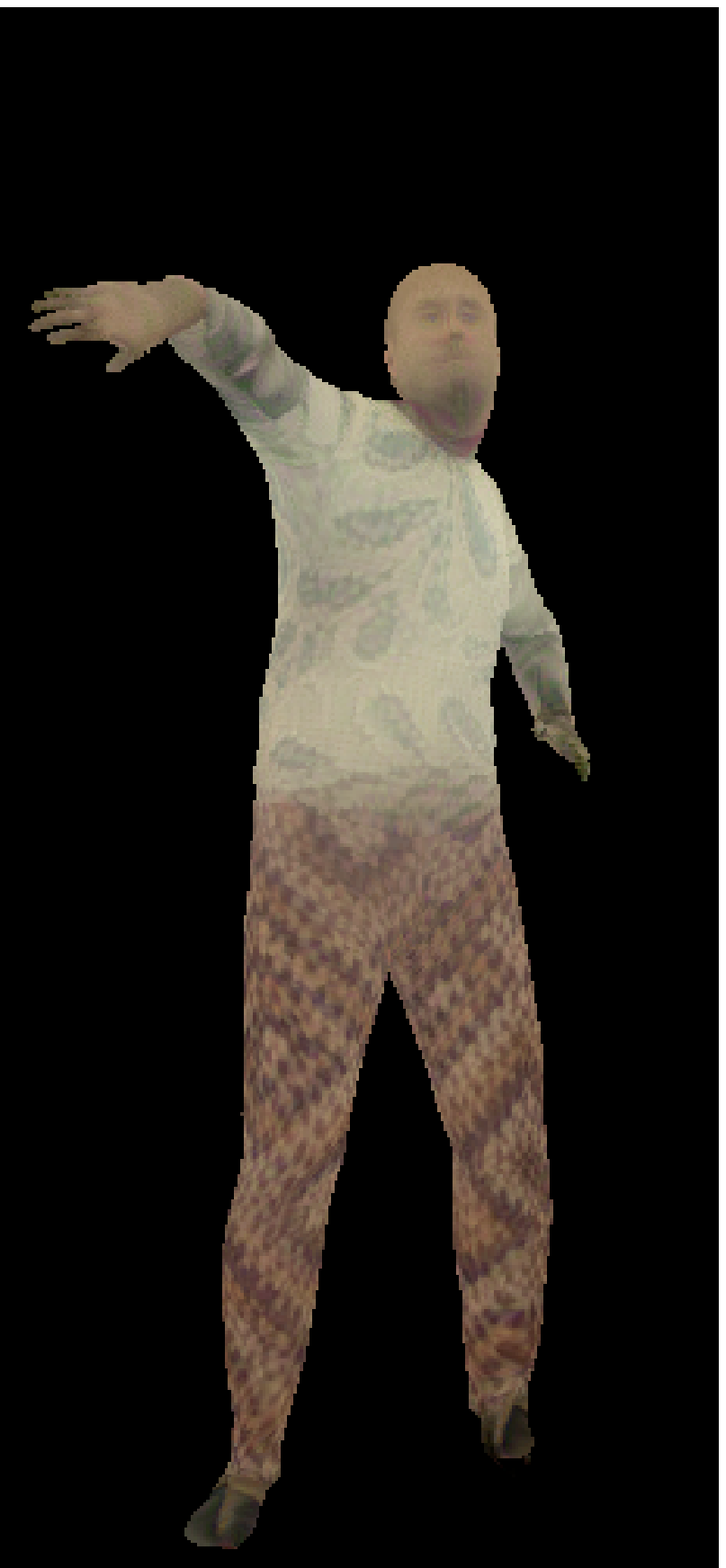}}
\subfloat[(d) Ours]{
\includegraphics[width=0.15\linewidth]{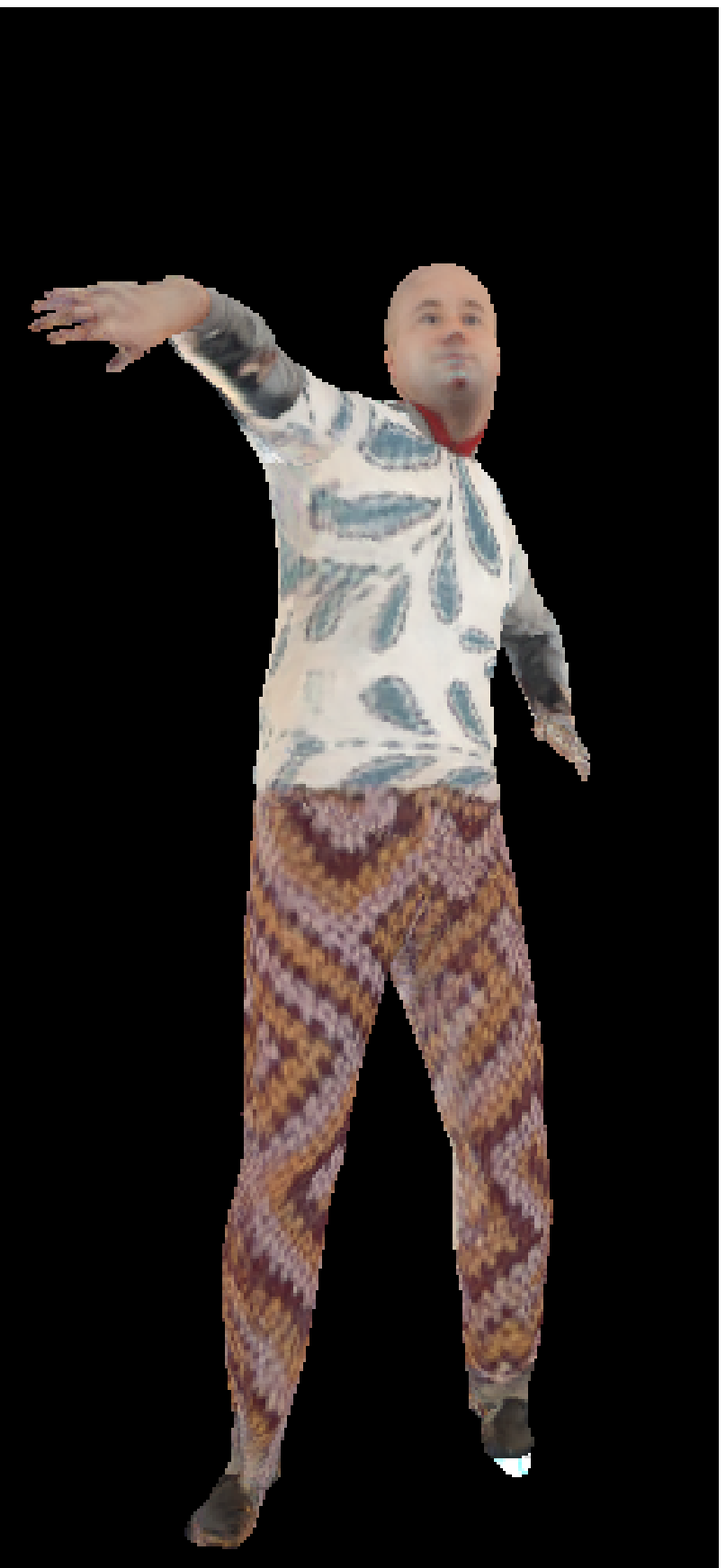}}
\caption{Comparing texture generation by visualizing no-shading image on synthetic data. (a) is ground truth albedo image. Our result has better contrast and clearer texture.}
\label{fig:texture2}
\end{figure}

See Fig.~\ref{fig:texture1} and~\ref{fig:texture2} for qualitative results. Our method provides clearer texture, better contrast, and less baked-in shading.
\section{Handle Different Human Models and Camera Parameters}
We establish a mapping between SMPL~\cite{loper2015smpl} and Our Human Model, so that we can replace the initial meshes of DeepHuman~\cite{zheng2019deephuman} and HMD~\cite{zhu2019detailed} by our coarse mesh. Besides, DeepHuman and HMD use different camera parameters from ours, and the cropping makes it hard to simply transform the mesh from one setting to the other. Thus, we adopt an approximate way to do this: We transform our coarse mesh to match the position and scale of their original initial mesh. Similarly, after obtaining their reconstruction result, we transform the mesh to match the position and scale of our coarse mesh, for DeepHuman, HMD, and their variants.
\section{Synthetic Training Data}
\begin{figure}[t]
\centering
\subfloat[]{
\includegraphics[height=45pt]{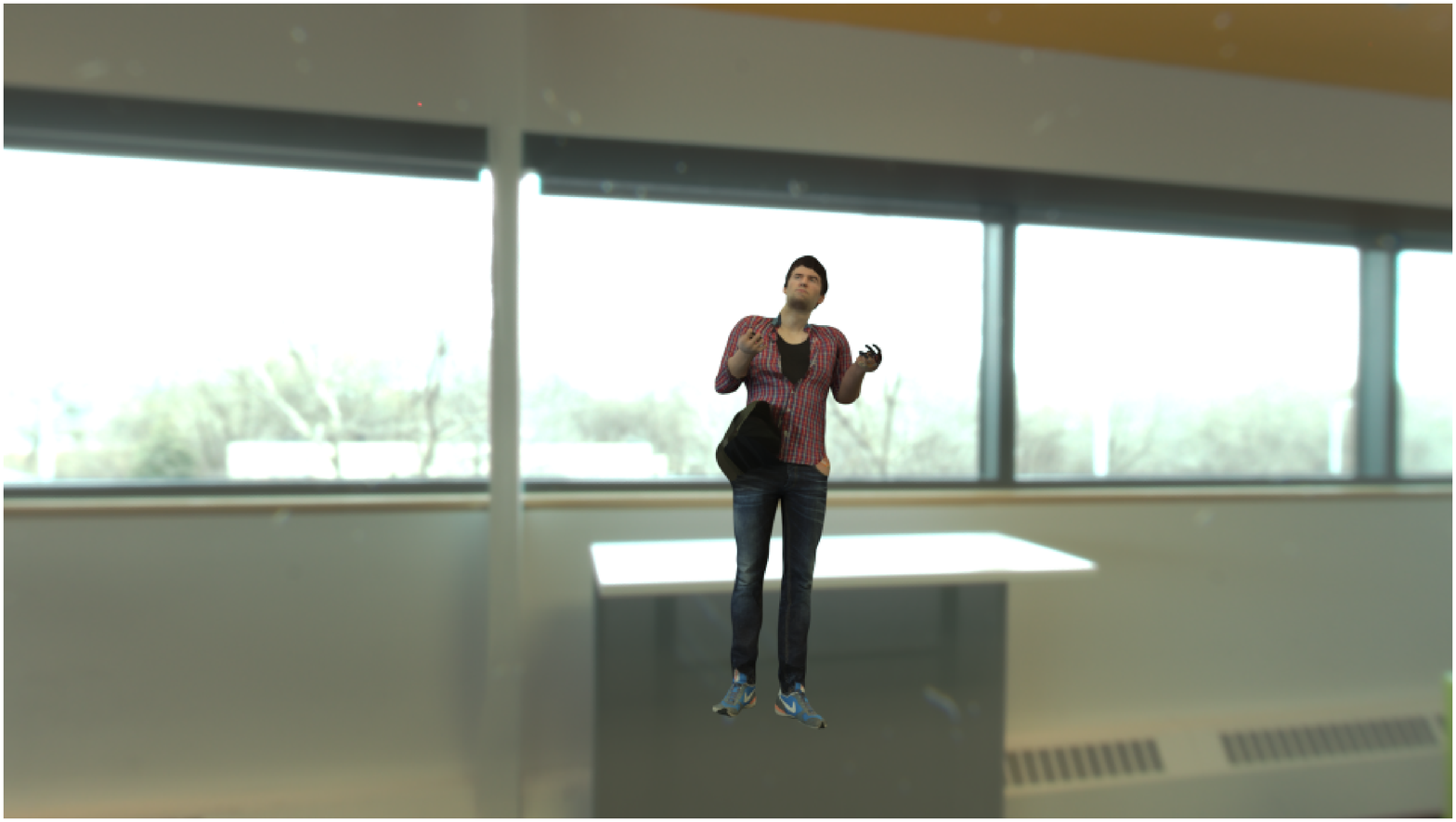}}
\subfloat[]{
\includegraphics[height=45pt]{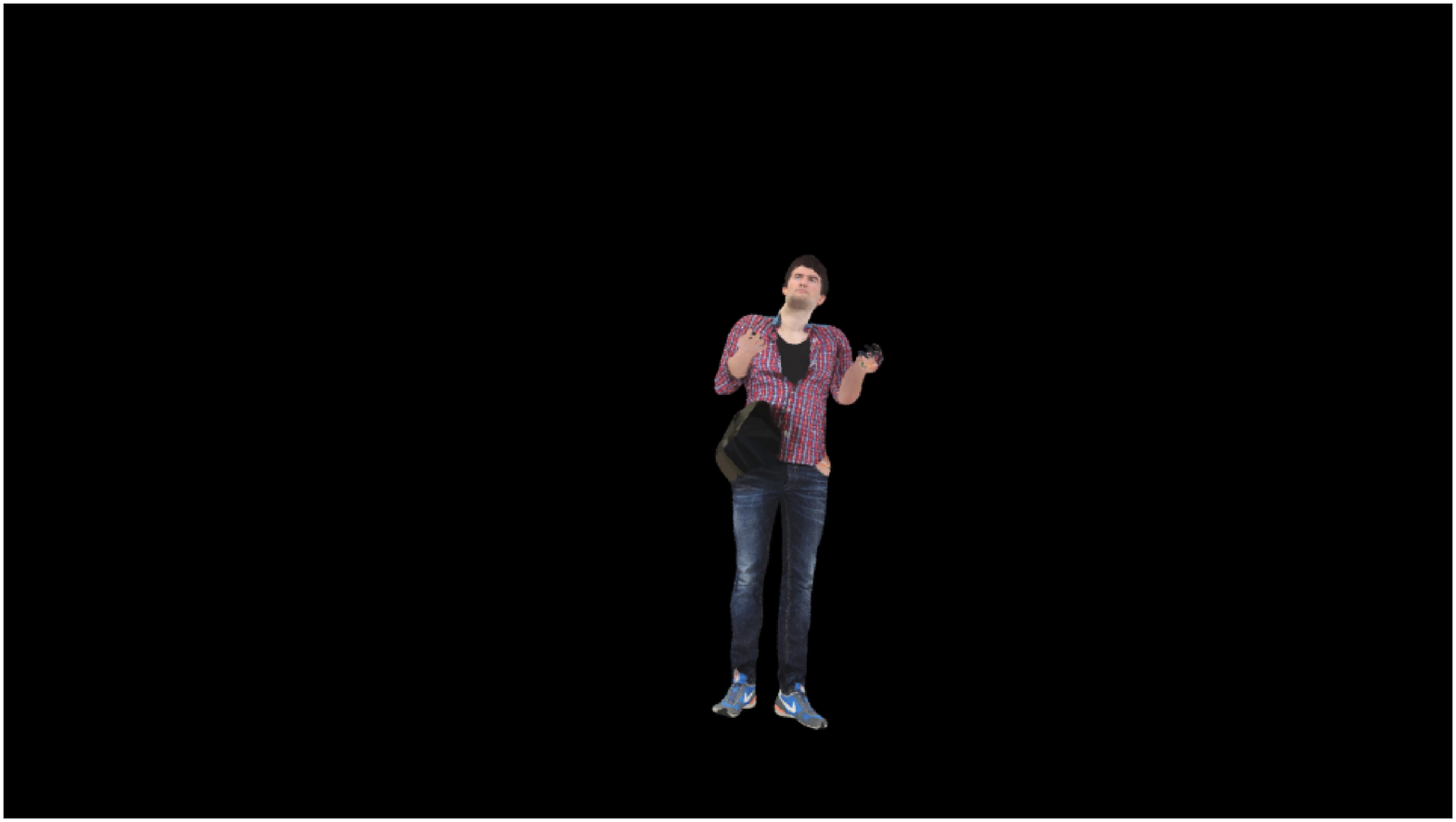}}
\subfloat[]{
\includegraphics[height=45pt]{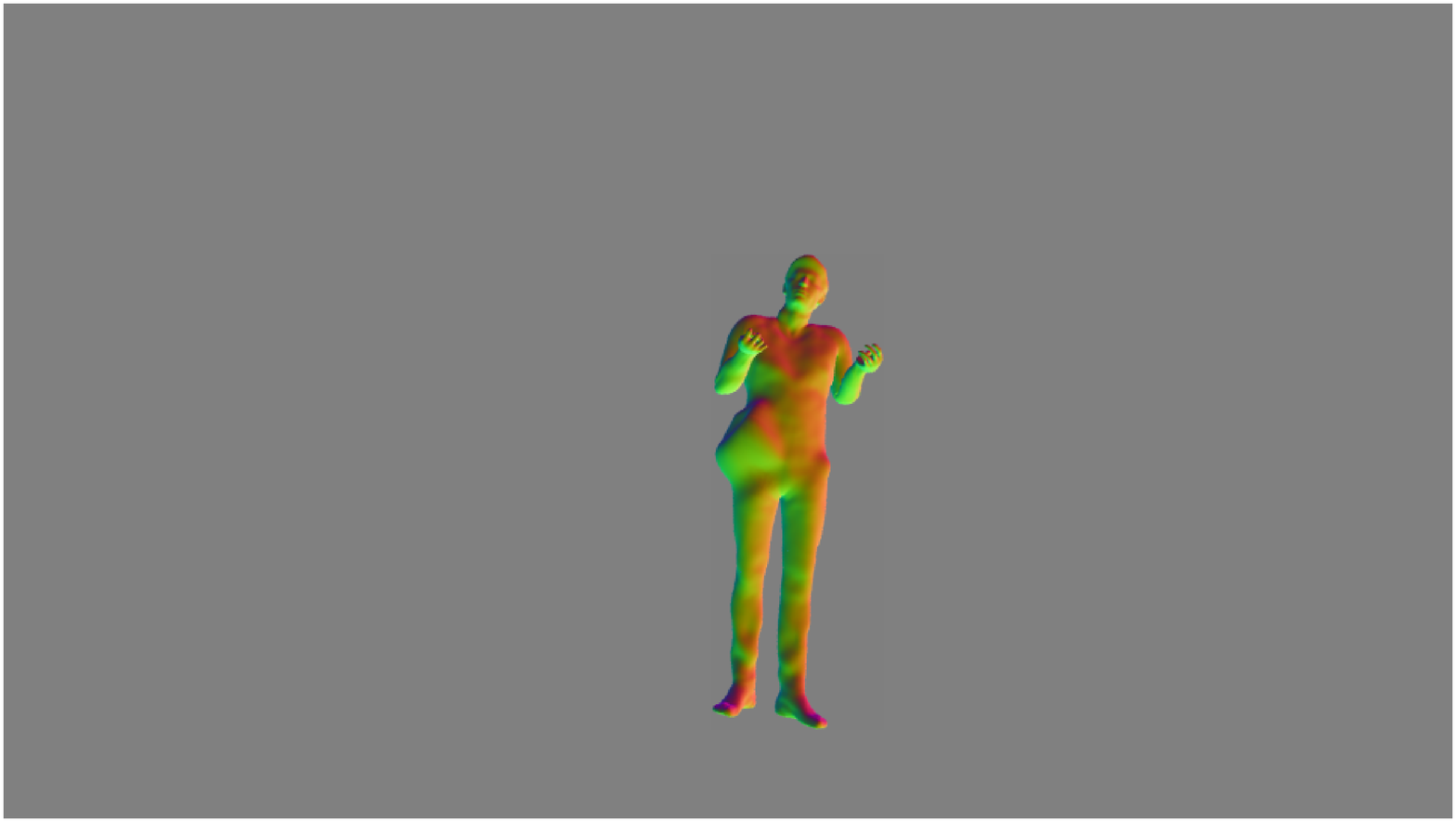}}
\subfloat[]{
\includegraphics[height=45pt]{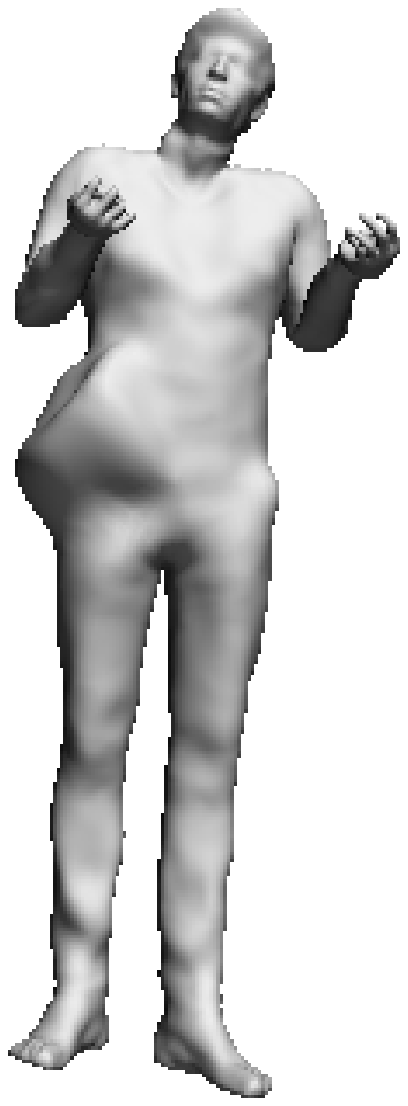}}
\vspace{-15pt}
\subfloat[]{
\includegraphics[height=45pt]{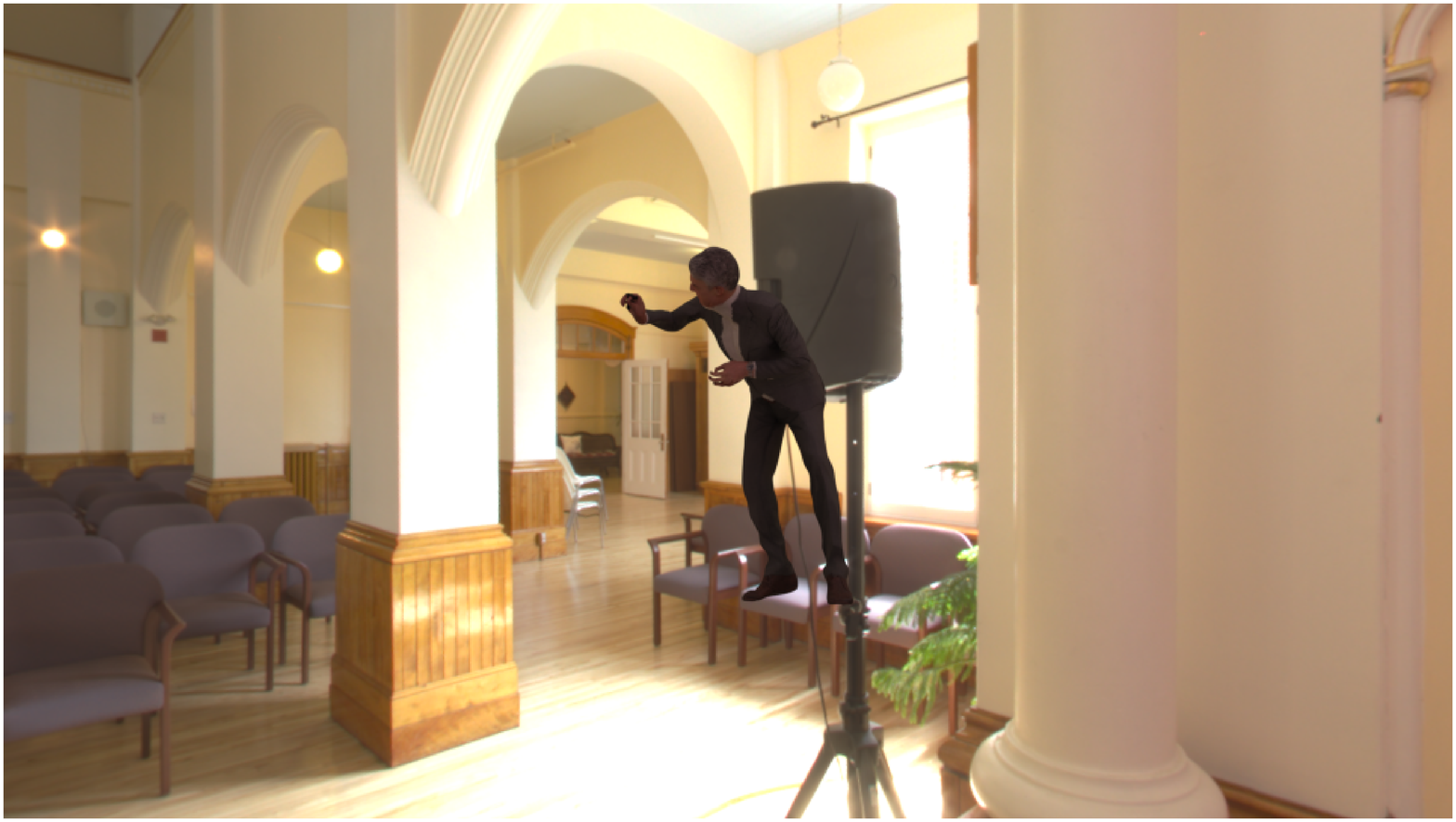}}
\subfloat[]{
\includegraphics[height=45pt]{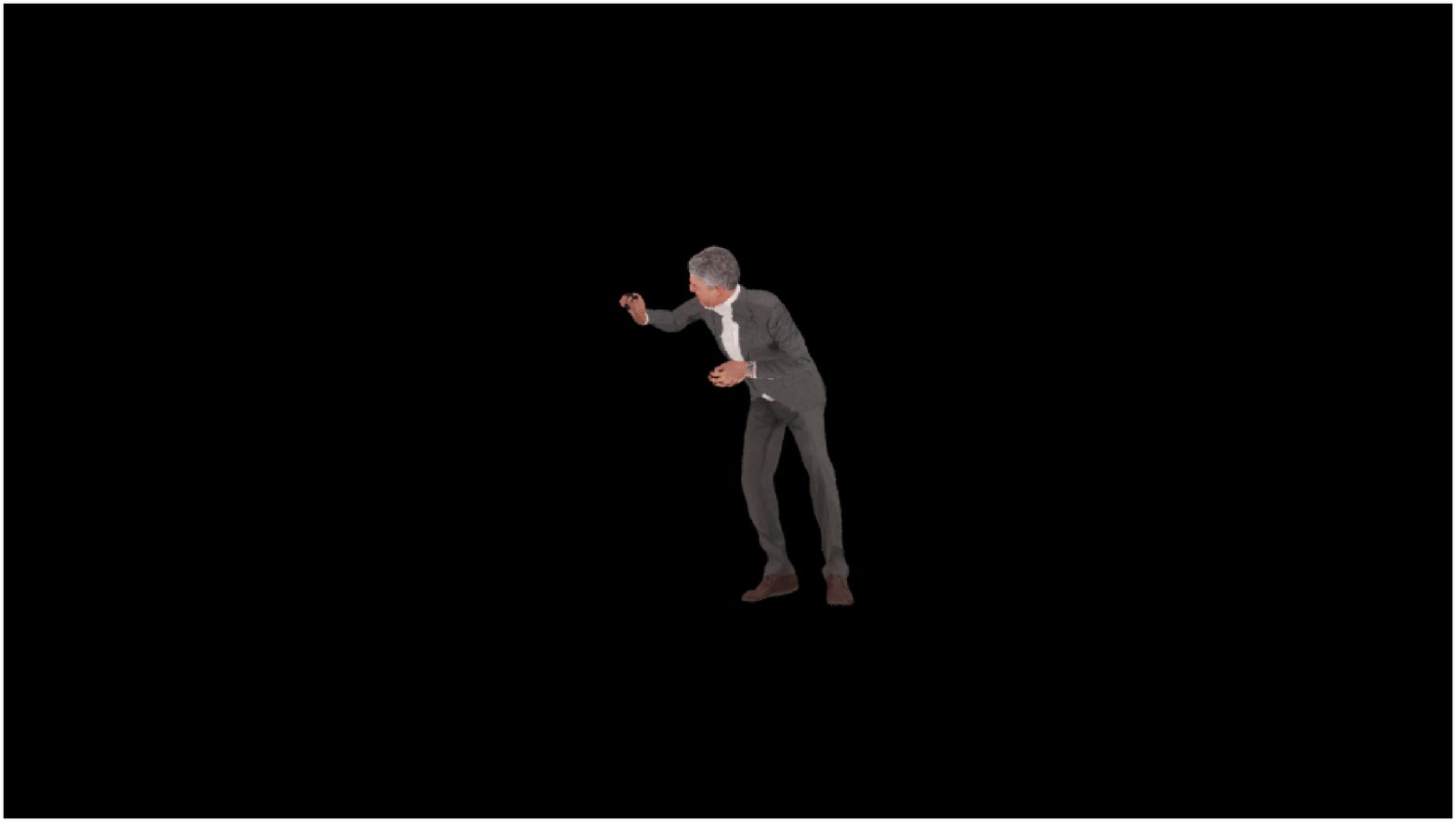}}
\subfloat[]{
\includegraphics[height=45pt]{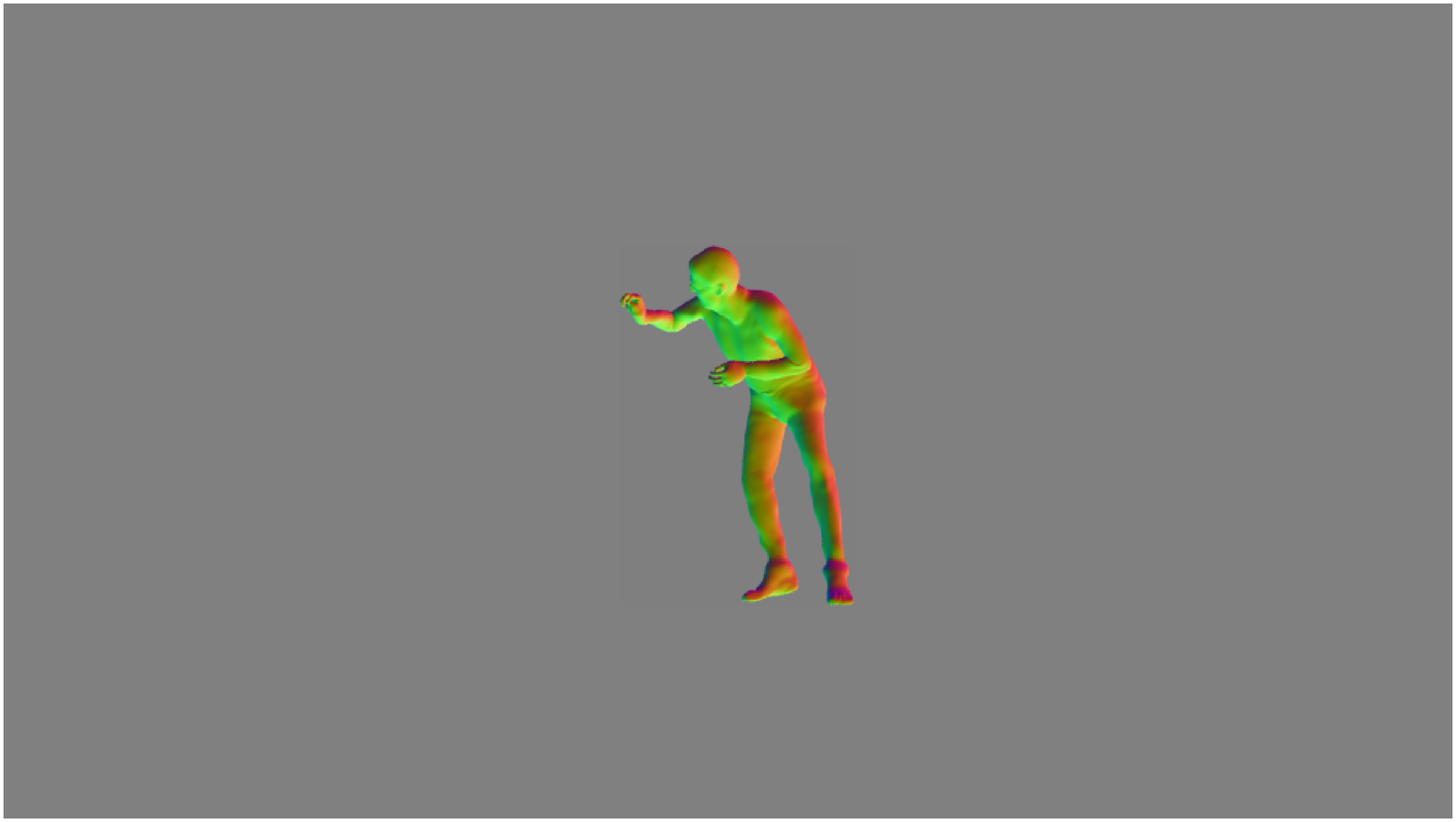}}
\subfloat[]{
\includegraphics[height=45pt]{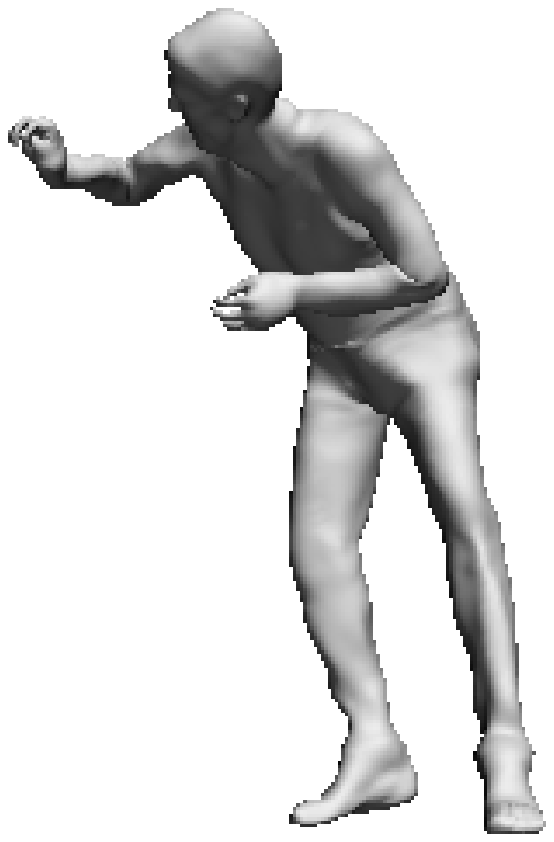}}
\vspace{-15pt}
\subfloat[]{
\includegraphics[height=45pt]{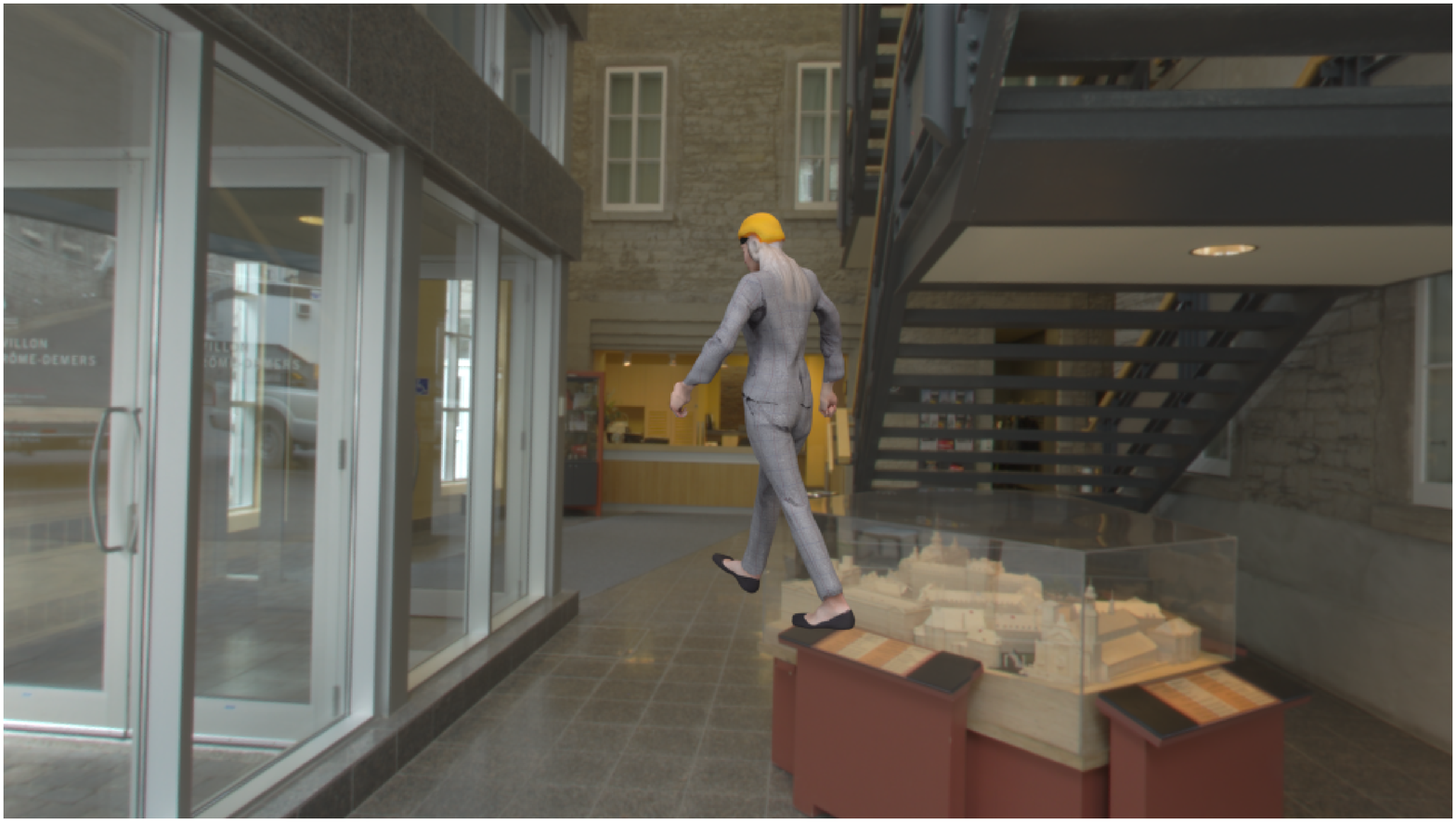}}
\subfloat[]{
\includegraphics[height=45pt]{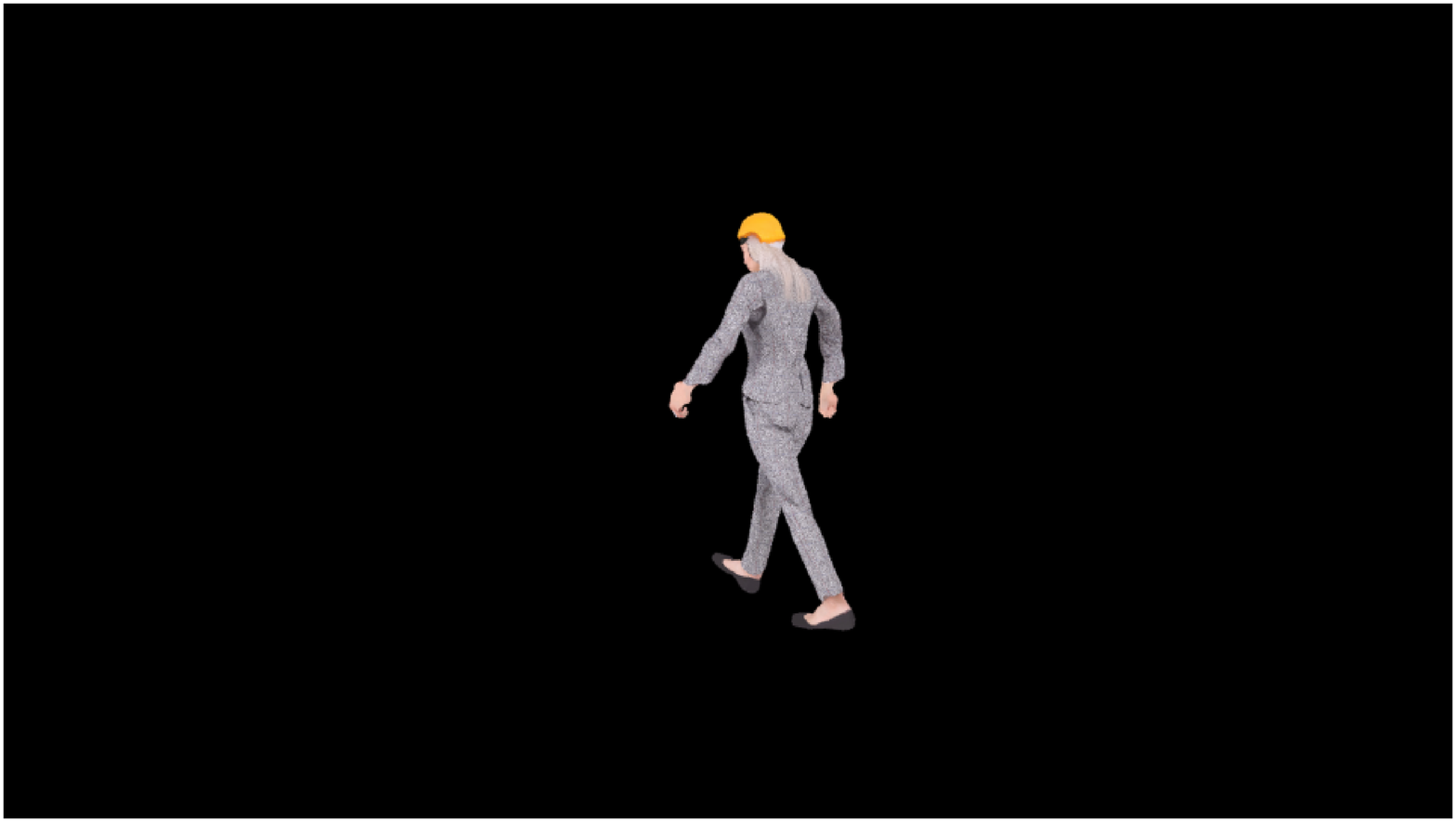}}
\subfloat[]{
\includegraphics[height=45pt]{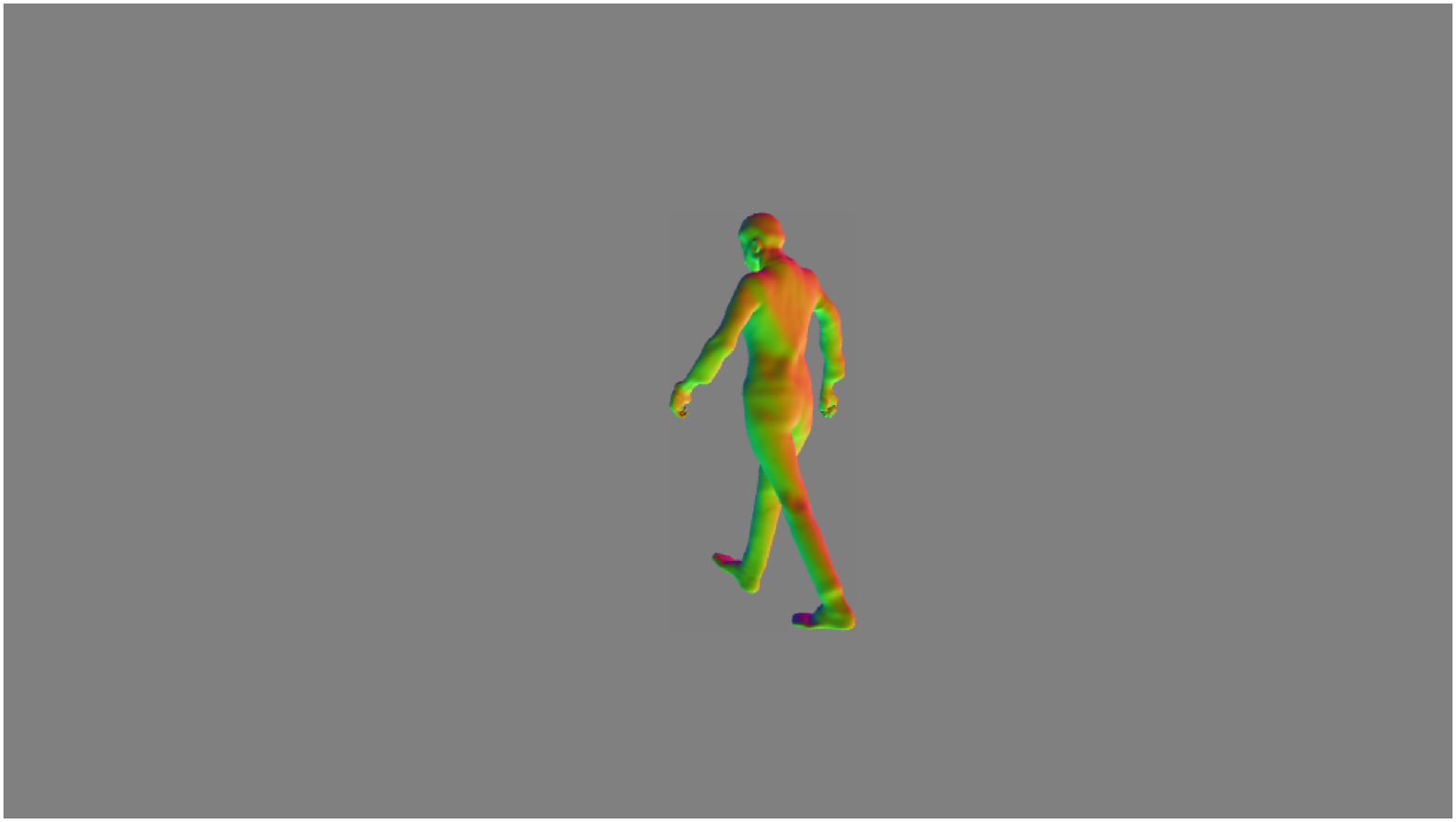}}
\subfloat[]{
\includegraphics[height=45pt]{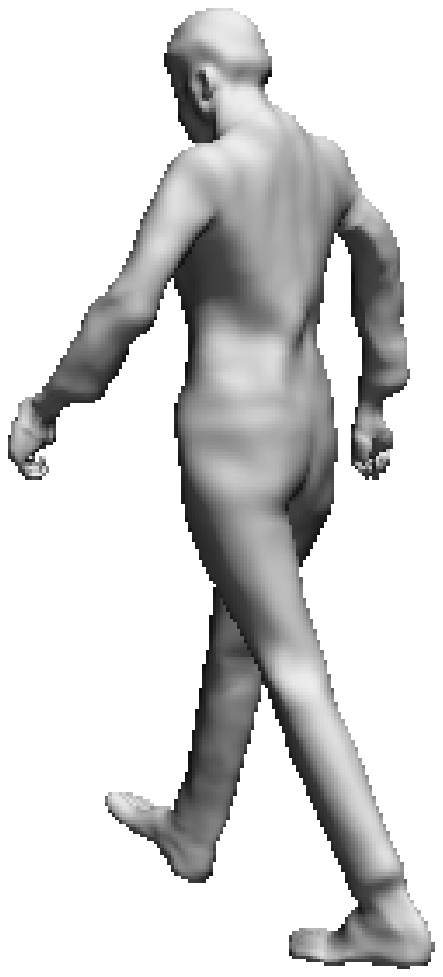}}
\vspace{-15pt}
\subfloat[(a) RGB]{
\includegraphics[height=45pt]{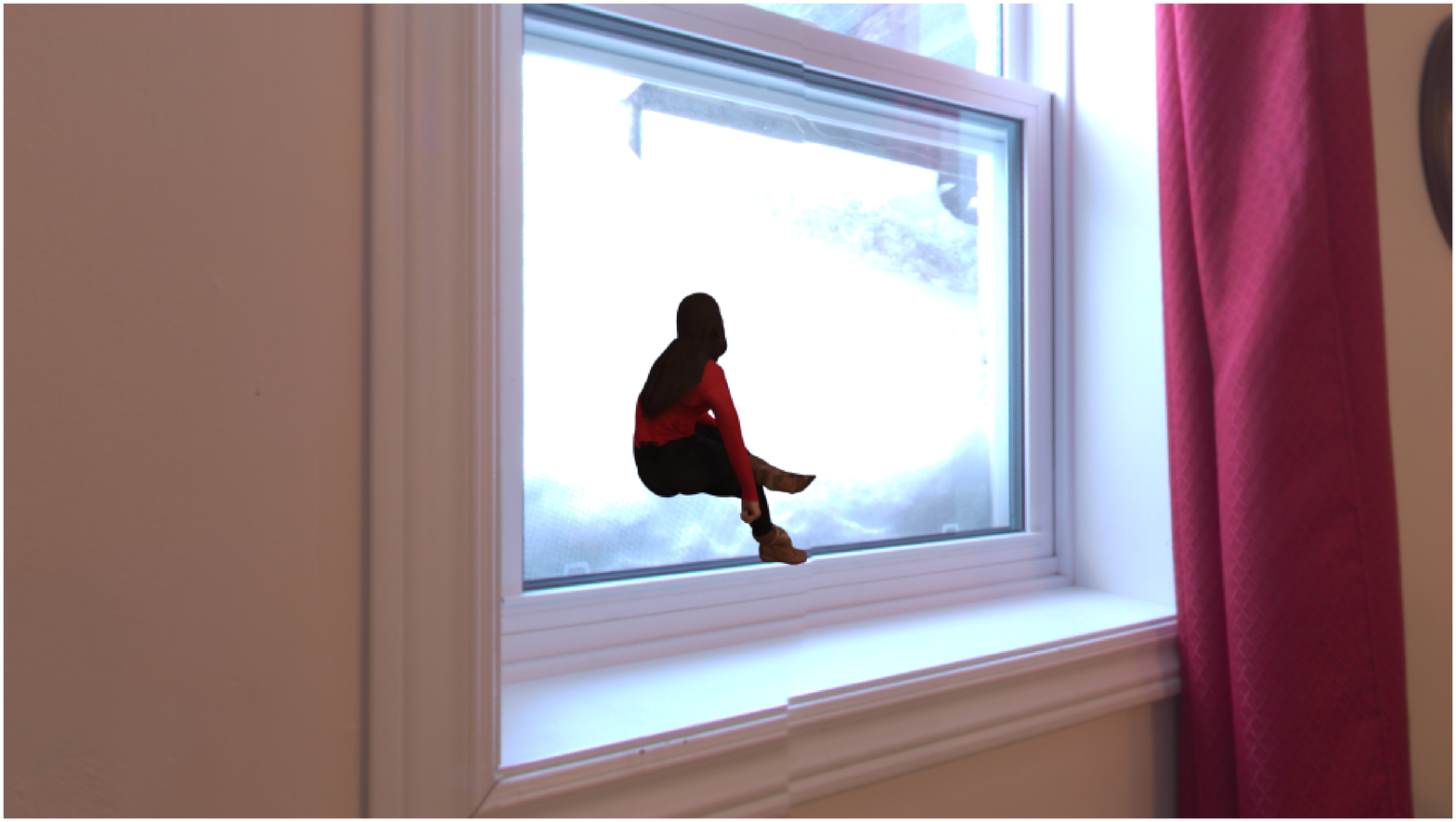}}
\subfloat[(b) Albedo]{
\includegraphics[height=45pt]{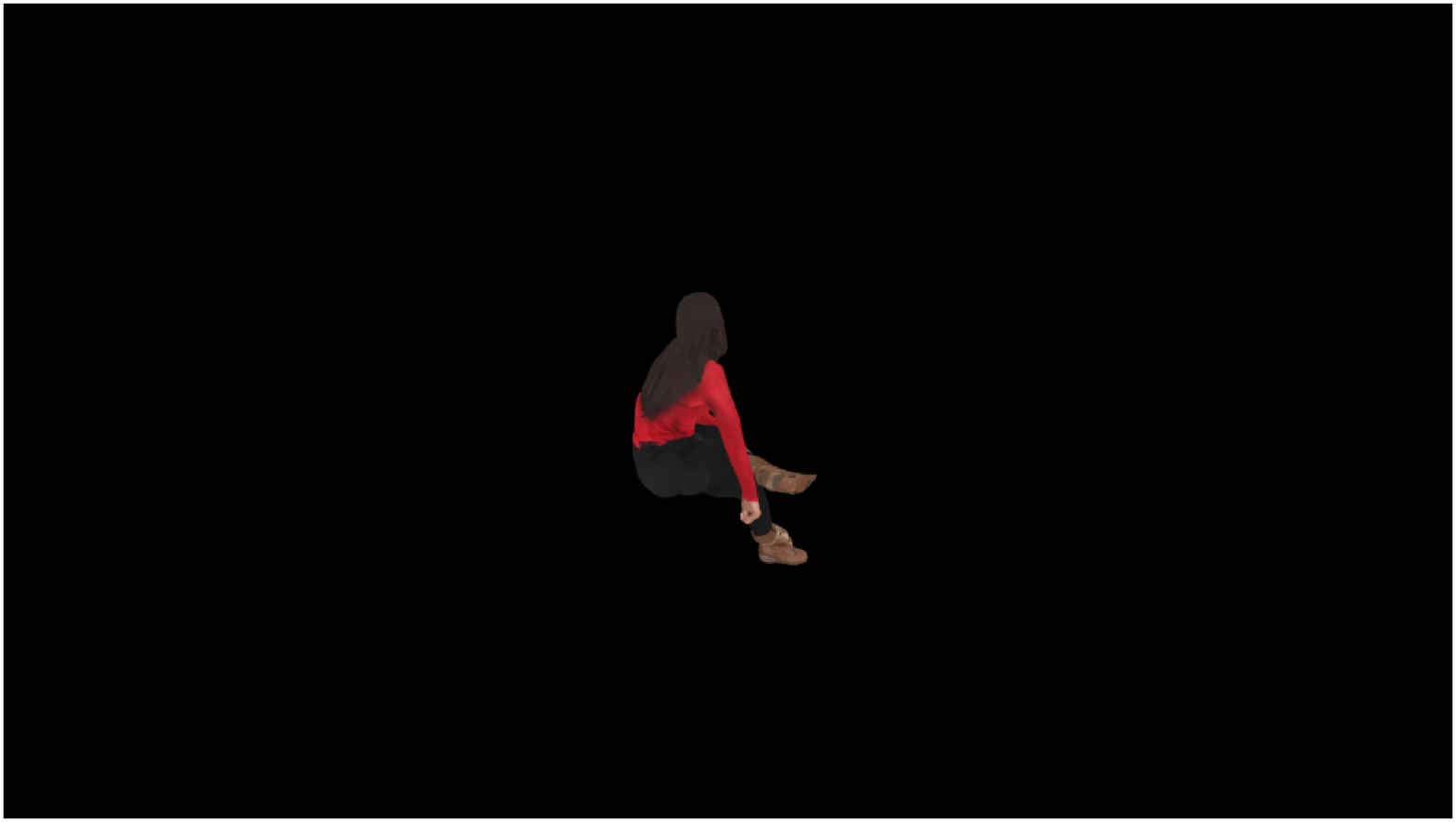}}
\subfloat[(c) Normal]{
\includegraphics[height=45pt]{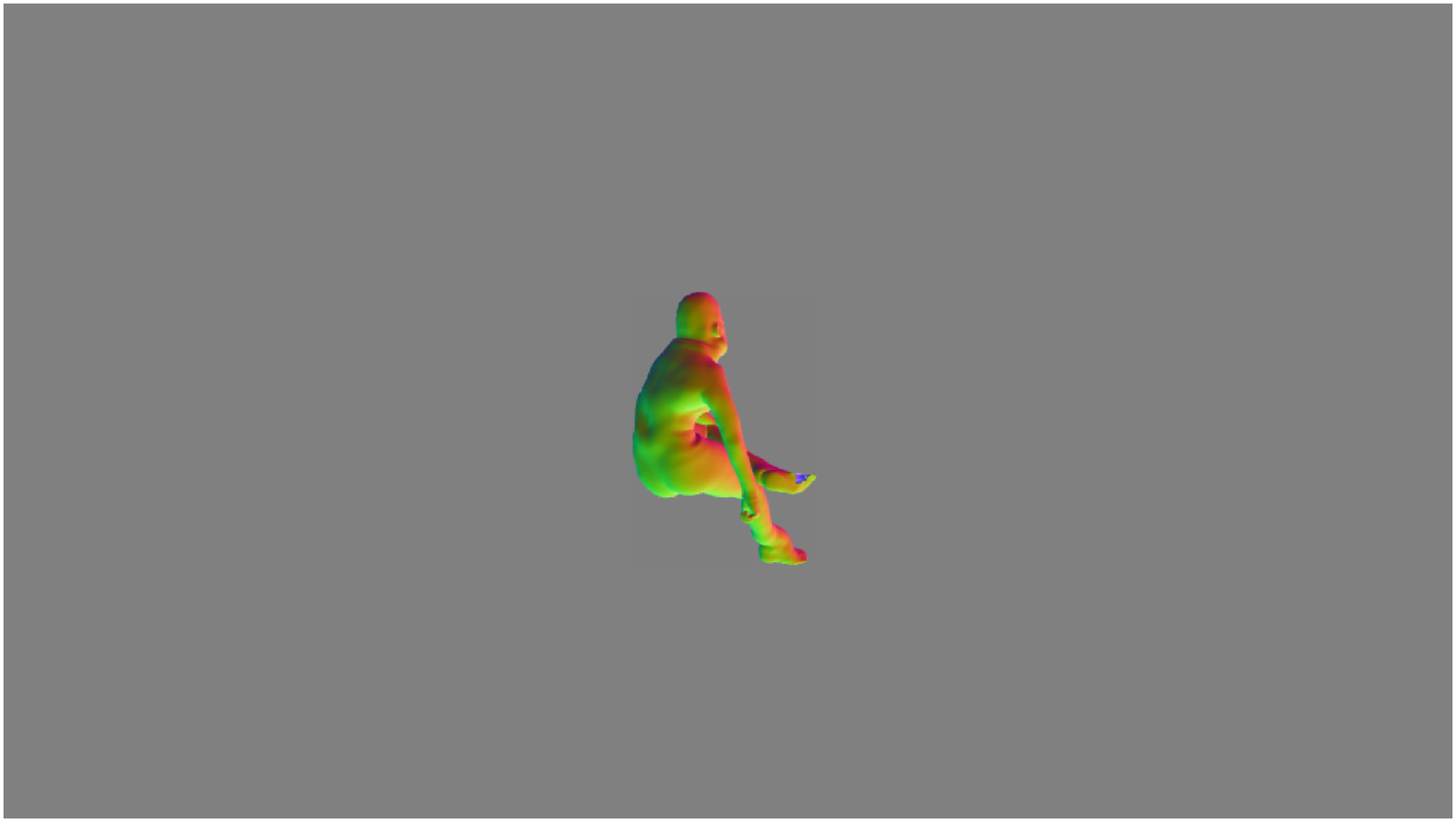}}
\subfloat[(d) Mesh]{
\includegraphics[height=45pt]{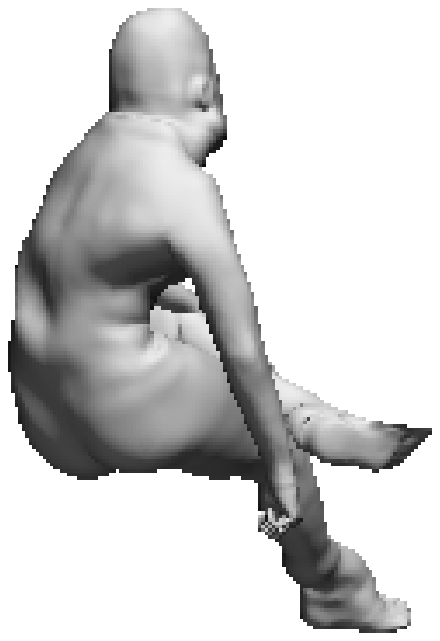}}
\caption{Example synthetic training data.}
\label{fig:synthetic}
\end{figure}

\begin{figure}
\centering
\subfloat[]{
\includegraphics[height=45pt]{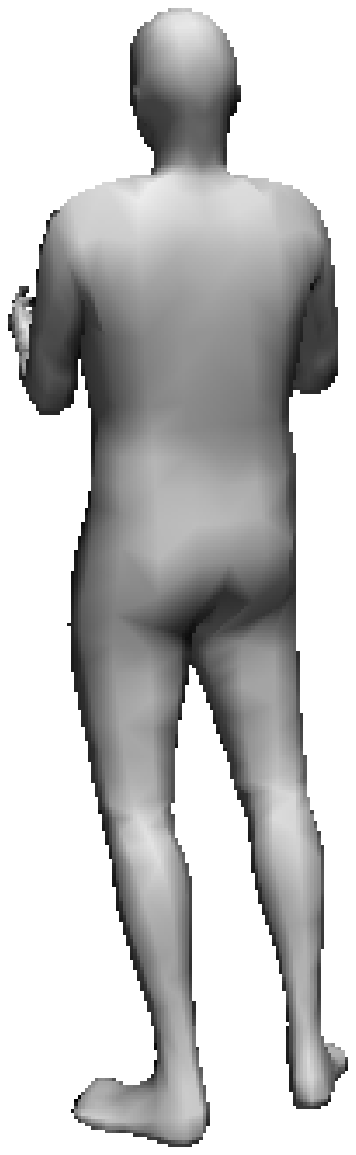}}
\subfloat[]{
\includegraphics[height=45pt]{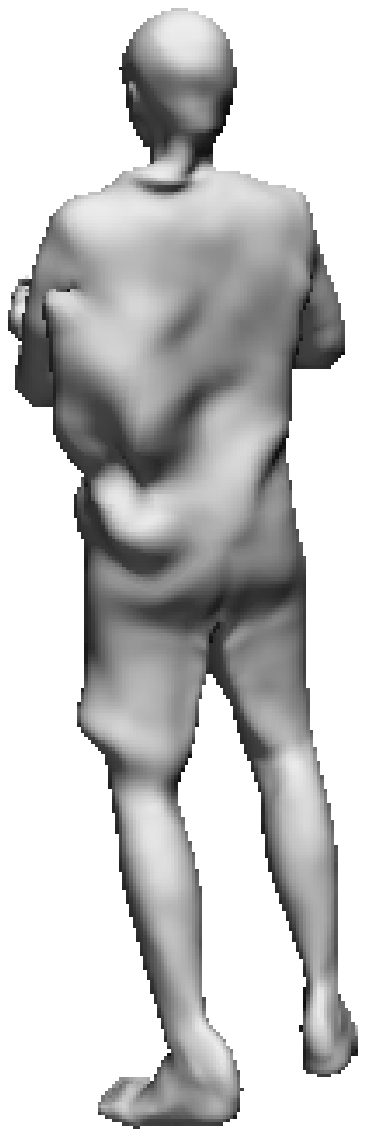}}
\subfloat[]{
\includegraphics[height=45pt]{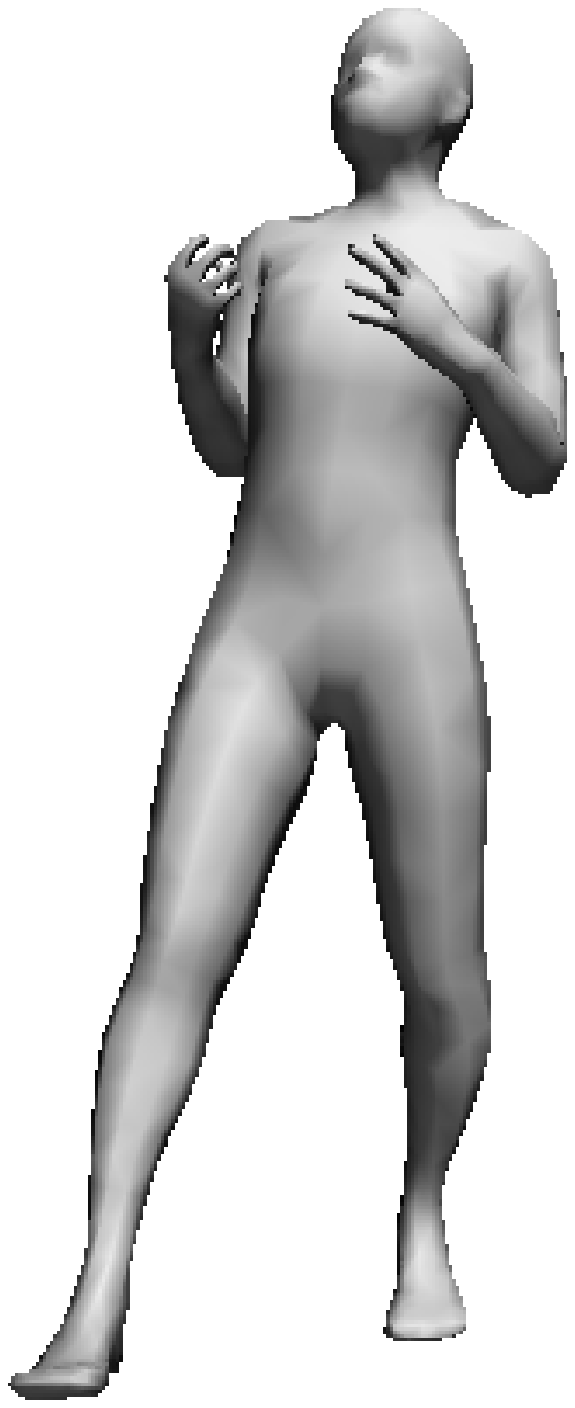}}
\subfloat[]{
\includegraphics[height=45pt]{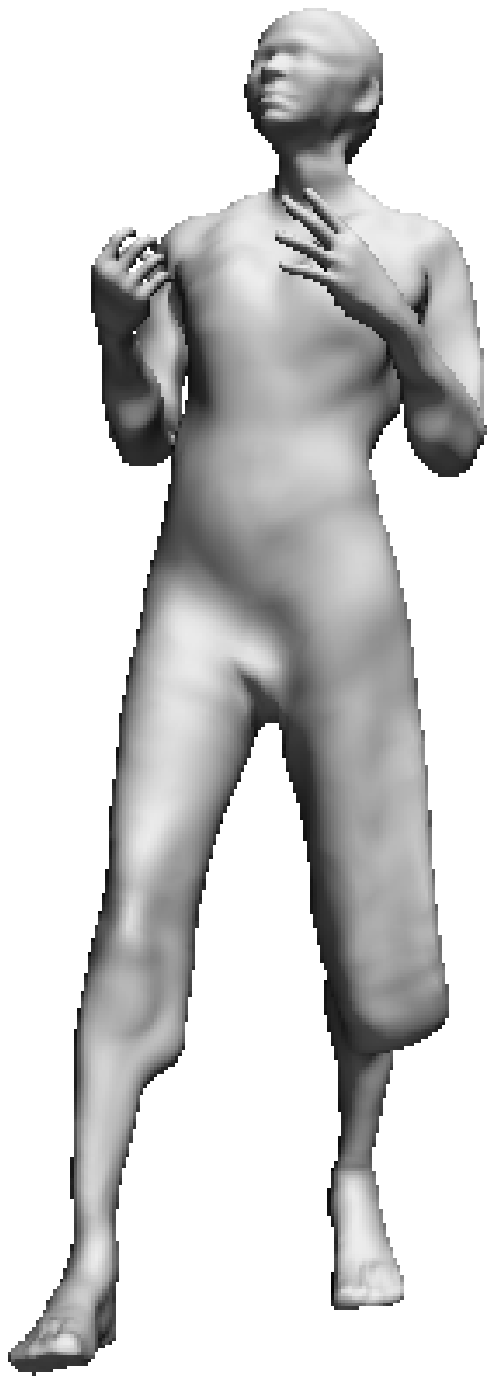}}
\vspace{-15pt}
\subfloat[Coarse]{
\includegraphics[height=45pt]{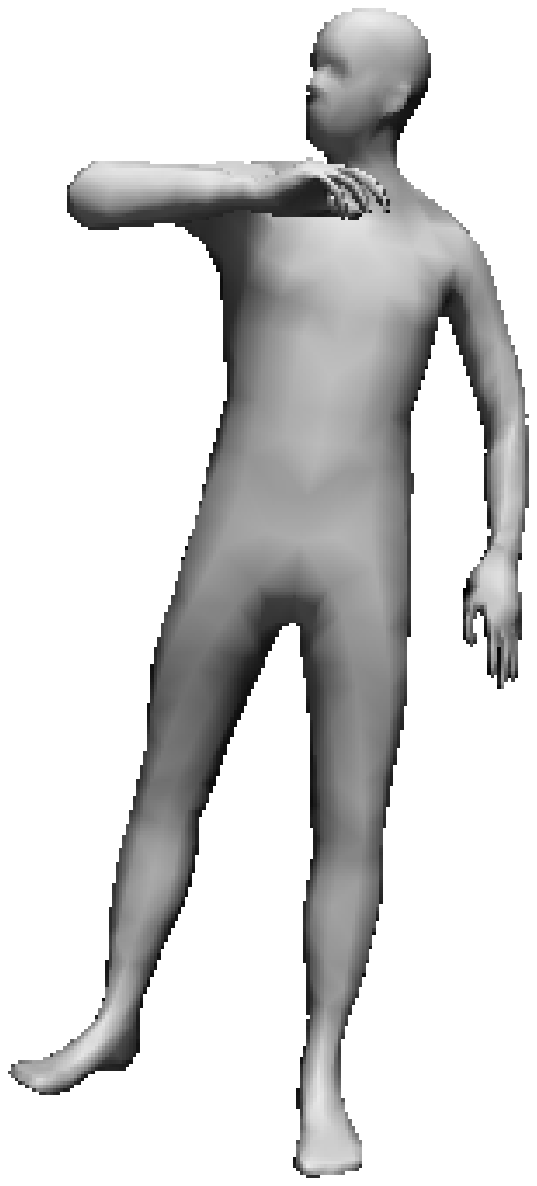}}
\subfloat[Fine]{
\includegraphics[height=45pt]{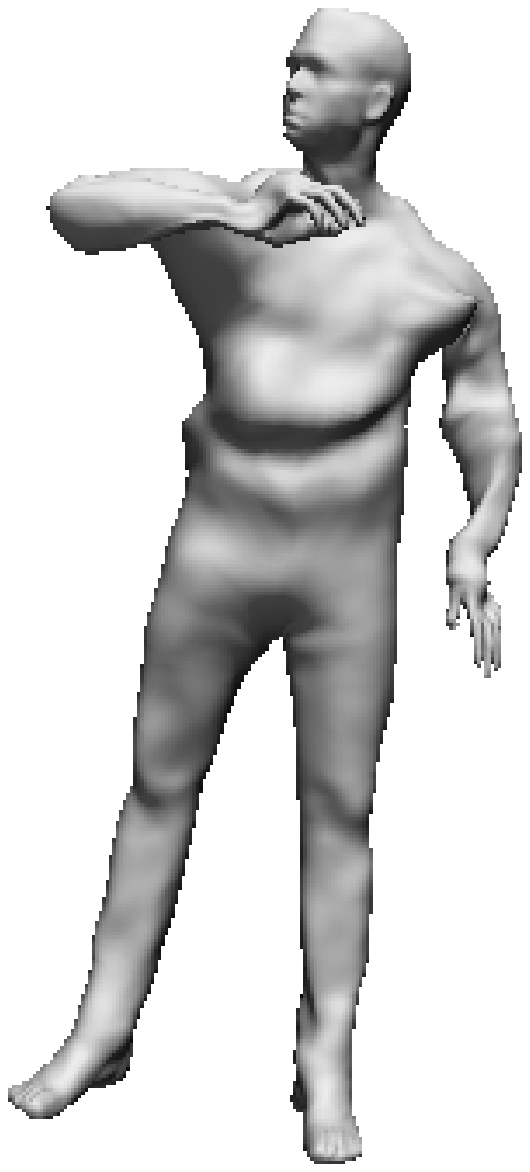}}
\subfloat[Coarse]{
\includegraphics[height=45pt]{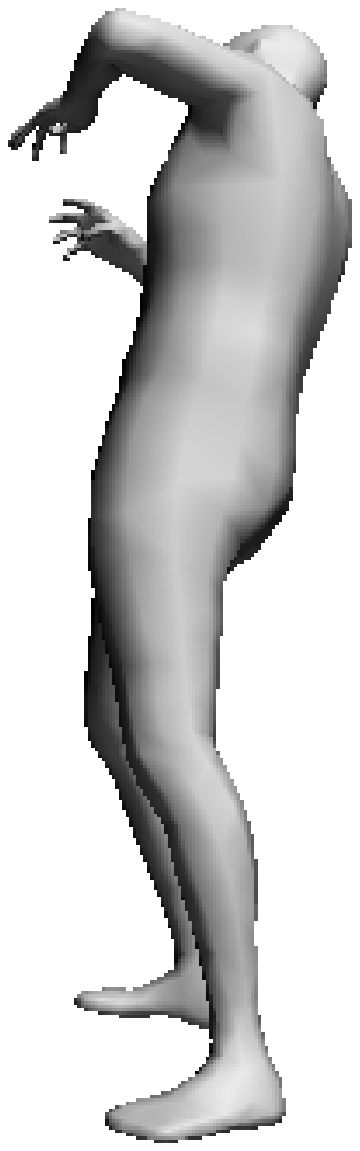}}
\subfloat[Fine]{
\includegraphics[height=45pt]{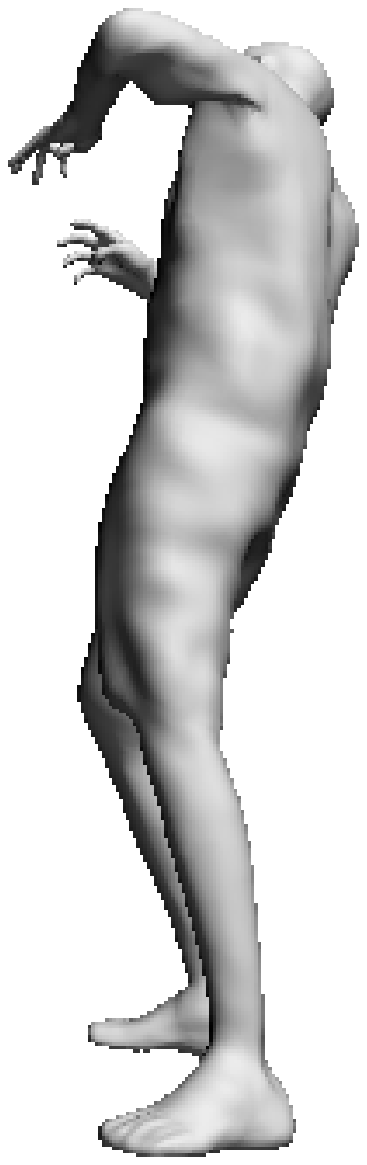}}
\caption{Example coarse and fine mesh pairs for pre-training MeshRef Module.}
\label{fig:coarse_fine}
\end{figure}

See Fig. \ref{fig:synthetic} for example synthetic training data. See Fig. \ref{fig:coarse_fine} for example coarse and fine mesh pairs for pre-training MeshRef module.

\clearpage
%
%
\bibliographystyle{splncs04}
\bibliography{egbib}
\end{document}